\documentclass{article}

\usepackage{microtype}
\usepackage{graphicx}
\usepackage{booktabs} % for professional tables

\usepackage{hyperref}

\usepackage[accepted]{icml2024}

\usepackage{amsmath}
\usepackage{verbatim}
\usepackage{natbib}
\usepackage{amssymb}
\usepackage{arydshln}
\usepackage{paralist}
\usepackage{wrapfig}
\setcitestyle{square,numbers}
\usepackage{mathtools}
\usepackage{circledsteps}
\usepackage{xspace}
\usepackage{subcaption}
\usepackage{bbm}
\usepackage{multirow}
\usepackage{enumitem}
\newcommand{\mat}[1]{\boldsymbol{#1}}
\renewcommand{\vec}[1]{{\bf{#1}}}

\newcommand{\prob}{\mathbb{P}}

\newcommand{\RR}{\mathbb{R}}
\newcommand{\EE}{\mathbb{E}}

\providecommand{\mA}{\ensuremath{\mat{A}}}
\providecommand{\mB}{\ensuremath{\mat{B}}}
\providecommand{\mC}{\ensuremath{\mat{C}}}
\providecommand{\mF}{\ensuremath{\mat{F}}}
\providecommand{\mH}{\ensuremath{\mat{H}}}
\providecommand{\mI}{\ensuremath{\mat{I}}}
\providecommand{\mM}{\ensuremath{\mat{M}}}
\providecommand{\mP}{\ensuremath{\mat{P}}}
\providecommand{\mQ}{\ensuremath{\mat{Q}}}
\providecommand{\mR}{\ensuremath{\mat{R}}}
\providecommand{\mS}{\ensuremath{\mat{S}}}
\providecommand{\mT}{\ensuremath{\mat{T}}}
\providecommand{\mU}{\ensuremath{\mat{U}}}
\providecommand{\mV}{\ensuremath{\mat{V}}}
\providecommand{\mW}{\ensuremath{\mat{W}}}
\providecommand{\mX}{\ensuremath{\mat{X}}}
\providecommand{\vf}{\ensuremath{\vec{f}}}
\providecommand{\vp}{\ensuremath{\vec{p}}}
\providecommand{\vq}{\ensuremath{\vec{q}}}
\providecommand{\vw}{\ensuremath{\vec{w}}}
\providecommand{\vx}{\ensuremath{\vec{x}}}

\providecommand{\shE}{\ensuremath{\set{\widehat{E}}}}

\providecommand{\tf}{\ensuremath{\widetilde{f}}}

\providecommand{\hG}{\ensuremath{\widehat{G}}}

\providecommand{\mtH}{\ensuremath{\mat{\widetilde{H}}}}

\providecommand{\mtM}{\ensuremath{\mat{\widetilde{M}}}}

\providecommand{\saQ}{\ensuremath{\set{\acute{Q}}}}
\providecommand{\sgQ}{\ensuremath{\set{\grave{Q}}}}

\providecommand{\mtR}{\ensuremath{\mat{\widetilde{R}}}}

\providecommand{\ds}{\ensuremath{\ddot{s}}}

\providecommand{\mtU}{\ensuremath{\mat{\widetilde{U}}}}
\providecommand{\mbU}{\ensuremath{\mat{\overline{U}}}}
\providecommand{\mdU}{\ensuremath{\mat{\ddot{U}}}}
\providecommand{\mbdU}{\ensuremath{\mat{\ddot{\overline{U}}}}}
\providecommand{\mbtU}{\ensuremath{\mat{\widetilde{\overline{U}}}}}

\providecommand{\mtV}{\ensuremath{\mat{\widetilde{V}}}}
\providecommand{\mdV}{\ensuremath{\mat{\ddot{V}}}}

\providecommand{\vdw}{\ensuremath{\vec{\ddot{w}}}}
\providecommand{\vtw}{\ensuremath{\vec{\widetilde{w}}}}
\providecommand{\mtW}{\ensuremath{\mat{\widetilde{W}}}}
\providecommand{\mbW}{\ensuremath{\mat{\overline{W}}}}
\providecommand{\mdW}{\ensuremath{\mat{\ddot{W}}}}
\providecommand{\mbdW}{\ensuremath{\mat{\ddot{\overline{W}}}}}
\providecommand{\mbtW}{\ensuremath{\mat{\widetilde{\overline{W}}}}}

\providecommand{\bone}{\mathbbm{1}}

\newcommand{\argmax}{\mathop{\mathrm{argmax}}}

\newcommand{\set}[1]{\mathcal{#1}}
\providecommand{\sC}{\ensuremath{\set{C}}}

\providecommand{\sE}{\ensuremath{\set{E}}}
\providecommand{\sH}{\ensuremath{\set{H}}}
\providecommand{\sL}{\ensuremath{\set{L}}}
\providecommand{\sN}{\ensuremath{\set{N}}}
\providecommand{\sO}{\ensuremath{\set{O}}}
\providecommand{\sP}{\ensuremath{\set{P}}}
\providecommand{\sQ}{\ensuremath{\set{Q}}}
\providecommand{\sR}{\ensuremath{\set{R}}}
\providecommand{\sV}{\ensuremath{\set{V}}}
\providecommand{\sZ}{\ensuremath{\set{Z}}}

\def\l2#1{\ensuremath\|#1\|_2}
\def\bl2#1{\ensuremath\big\|#1\big\|_2}
\def\bbl2#1{\ensuremath\Big\|#1\Big\|_2}
\def\bbbl2#1{\ensuremath\bigg\|#1\bigg\|_2}
\def\bbbbl2#1{\ensuremath\bigg\|#1\Bigg\|_2}

\def\lfr#1{\ensuremath\|#1\|_F}

\def\bblfr#1{\ensuremath\Big\|#1\Big\|_F}

\def\linf#1{\ensuremath\|#1\|_\infty}

\def\size#1{\ensuremath|#1|}

\def\bbsize#1{\ensuremath\Big|#1\Big|}
\def\bbbsize#1{\ensuremath\bigg|#1\bigg|}

\providecommand{\xv}{\mX_{\textup{ent}}}
\providecommand{\xr}{\mX_{\textup{rel}}}
\providecommand{\nv}{\ensuremath{|\sV|}}
\providecommand{\nr}{\ensuremath{|\sR|}}
\providecommand{\nE}{\ensuremath{|\sE|}}
\providecommand{\nhE}{\ensuremath{|\shE|}}

\providecommand{\mzero}{\ensuremath{\mat{0}}}
\providecommand{\vzero}{\ensuremath{\vec{0}}}

\def\diag#1{\text{diag}{\left(#1\right)}}
\DeclareMathOperator{\dg}{\texttt{deg}}

\providecommand{\fb}{FB15K237\xspace}
\providecommand{\cde}{CoDEx-M\xspace}
\providecommand{\umls}{UMLS-43\xspace}

\usepackage{amsmath}
\usepackage{amssymb}
\usepackage{mathtools}
\usepackage{amsthm}

\theoremstyle{plain}
\newtheorem{theorem}{Theorem}[section]

\newtheorem{lemma}[theorem]{Lemma}
\newtheorem{corollary}[theorem]{Corollary}
\theoremstyle{definition}
\newtheorem{definition}[theorem]{Definition}

\theoremstyle{remark}

\theoremstyle{plain}
\newtheorem*{thm:transbound}{Theorem~\ref*{thm:transbound}}
\newtheorem*{thm:boundtd}{Theorem~\ref*{thm:boundtd}}
\newtheorem*{thm:boundsm}{Theorem~\ref*{thm:boundsm}}

\icmltitlerunning{PAC-Bayesian Generalization Bounds for Knowledge Graph Representation Learning}

\begin{document}

\twocolumn[
\icmltitle{PAC-Bayesian Generalization Bounds for Knowledge Graph \\Representation Learning}




\begin{icmlauthorlist}
\icmlauthor{Jaejun Lee}{kaist}
\icmlauthor{Minsung Hwang}{kaist}
\icmlauthor{Joyce Jiyoung Whang}{kaist}
%\icmlauthor{Firstname4 Lastname4}{sch}
%\icmlauthor{Firstname5 Lastname5}{yyy}
%\icmlauthor{Firstname6 Lastname6}{sch,yyy,comp}
%\icmlauthor{Firstname7 Lastname7}{comp}
%\icmlauthor{Firstname8 Lastname8}{sch}
%\icmlauthor{Firstname8 Lastname8}{yyy,comp}
\end{icmlauthorlist}

\icmlaffiliation{kaist}{School of Computing, KAIST, Daejeon, South Korea}
%\icmlaffiliation{comp}{Company Name, Location, Country}
%\icmlaffiliation{sch}{School of ZZZ, Institute of WWW, Location, Country}

\icmlcorrespondingauthor{Joyce Jiyoung Whang}{jjwhang@kaist.ac.kr}
%\icmlcorrespondingauthor{Firstname2 Lastname2}{first2.last2@www.uk}

\icmlkeywords{Generalization Bound, Knowledge Graph, Representation Learning, PAC-Bayesian, Graph Neural Network, Knowledge Graph Completion, Knowledge Graph Embedding}

%TLDR: We present the first PAC-Bayesian generalization bounds for a broad class of knowledge graph representation learning methods ranging from GNN-based to shallow-architecture methods.

\vskip 0.3in
]



\printAffiliationsAndNotice{}  % leave blank if no need to mention equal contribution

\begin{abstract}
While a number of knowledge graph representation learning (KGRL) methods have been proposed over the past decade, very few theoretical analyses have been conducted on them. In this paper, we present the first PAC-Bayesian generalization bounds for KGRL methods. To analyze a broad class of KGRL models, we propose a generic framework named ReED (Relation-aware Encoder-Decoder), which consists of a relation-aware message passing encoder and a triplet classification decoder. Our ReED framework can express at least 15 different existing KGRL models, including not only graph neural network-based models such as R-GCN and CompGCN but also shallow-architecture models such as RotatE and ANALOGY. Our generalization bounds for the ReED framework provide theoretical grounds for the commonly used tricks in KGRL, e.g., parameter-sharing and weight normalization schemes, and guide desirable design choices for practical KGRL methods. We empirically show that the critical factors in our generalization bounds can explain actual generalization errors on three real-world knowledge graphs.
\end{abstract}

%While a number of representation learning methods have been proposed for knowledge graph completion over the past decade, very few theoretical analyses have been conducted on them. We present the first theoretical generalization bounds for a broad class of knowledge graph representation learning (KGRL) methods. We propose a generic framework named ReED (Relation-aware Encoder-Decoder), which consists of a relation-aware message passing encoder and a triplet classification decoder. Our ReED framework can express at least 15 different existing KGRL models, including not only graph neural network-based models such as R-GCN and CompGCN but also simpler models such as RotatE and ANALOGY. Our PAC-Bayesian generalization bounds for the ReED framework provide theoretical grounds for the commonly used tricks in KGRL, e.g., parameter-sharing and weight normalization schemes, and guide desirable design choices for practical KGRL methods. We empirically show that the critical factors in our generalization bounds can explain actual generalization errors on three real-world knowledge graphs.

%e.g., the advantage of a mean aggregator over a sum aggregator
%e.g., parameter-sharing and weight normalization schemes,

\section{Introduction}
Knowledge graphs consist of entities and relations where a known fact is represented as a triplet of a head entity, a relation, and a tail entity, e.g., (\texttt{Washington\_DC}, \texttt{Capital\_Of}, \texttt{USA}). Since real-world knowledge graphs do not exhaustively represent all known facts, knowledge graph completion aims to predict missing facts in knowledge graphs. That is, given an incomplete knowledge graph, the goal is to add missing triplets by finding plausible combinations of the entities and relations. This task can be cast into a triplet classification problem where a model determines whether a given triplet is plausible or not~\cite{survey}.

Knowledge graph representation learning (KGRL) has been considered one of the most effective approaches for knowledge graph completion~\citep{transe,transr,dist,quate,hynt,vista}. By learning representations of entities and relations in a knowledge graph, KGRL methods compute a score for each triplet and determine whether the given triplet is likely true or false based on the score~\cite{transh}. While shallow-architecture models such as RotatE~\cite{rotate} and ANALOGY~\cite{analogy} also work well in practice, many recently proposed methods~\cite{rgcn,comp,wgcn,ingram} utilize graph neural networks (GNNs)~\cite{gnnsurv} or message passing neural networks (MPNNs)~\cite{mpnn} to improve the performance by adding a neural message passing encoder.

While KGRL techniques are widely used in many real-life applications~\cite{survey}, very few theoretical analyses have been conducted on them. Recently,~\citeauthor{rwl}~\cite{rwl} and~\citeauthor{lpwl}~\cite{lpwl} have extended the Weisfeiler-Lehman (WL) test~\cite{wl} to multi-relational graphs to investigate the expressive power of GNNs for knowledge graphs; the expressive power indicates how well a model can distinguish graphs with different structures~\cite{gin}. On the other hand, the generalization bound indicates how successfully a model solves a given task on the entire dataset compared to its performance on a training set~\cite{gcrade,gendl}. Though a number of KGRL methods have been proposed over the past decade, the generalization bounds of KGRL have rarely been studied.

%However, the generalization bound has not been studied for learning representations on knowledge graphs, 
%of KGRL has not been studied yet, where the generalization bound indicates how successfully a model solves a given task on the entire dataset compared to its performance on a training set~\cite{gcrade,gendl}.

%whereas the generalization bound indicates how successfully a model solves a given task on the entire dataset compared to its performance on a training set~\cite{gcrade,gendl}. Our focus in this paper is on the generalization bound of KGRL.

PAC (Probably Approximately Correct) learning theory provides fundamental tools for analyzing the generalization bounds~\cite{paclearn}; the generalization bounds have been explored using different PAC-based approaches, such as the VC dimension-based~\cite{vc}, the Rademacher complexity-based~\cite{rade}, and the PAC-Bayesian approaches~\cite{pac,pacmargin}. There have been some  studies about the generalization bounds for deep neural networks~\cite{vcnn,radenn,nnpac} or for GNNs on standard graphs~\cite{difpac,subpac,ncrade,radetrans,oodlp} but not for knowledge graphs. Regarding the generalization bounds of GNNs,~\citeauthor{gcpac}~\cite{gcpac} have shown that the PAC-Bayesian approach can make a tighter bound than the other approaches~\cite{gcvc,gcrade}. 

%However, there has been no analysis of the generalization bound for KGRL methods.

In this paper, we present the first PAC-Bayesian generalization bounds for KGRL methods. To comprehensively represent and analyze various KGRL models and their possible variants, we design the \textbf{Re}lation-aware \textbf{E}ncoder-\textbf{D}ecoder (ReED) framework consisting of the \textbf{R}elation-\textbf{A}ware \textbf{M}essage-\textbf{P}assing (RAMP) encoder and a triplet classification decoder; ReED is a generic framework representing at least 15 different KGRL methods, including both GNN-based and shallow-architecture models. We derive concrete generalization bounds for ReED by proposing a transductive PAC-Bayesian approach for a deterministic triplet classifier, which extends the analyses of~\citeauthor{pactrans}~\cite{pactrans} and~\citeauthor{nnpac}~\cite{nnpac}. We also empirically show the increasing and decreasing trends of the actual generalization errors regarding the critical factors in the generalization bounds using three real-world knowledge graphs. Our theoretical generalization bounds and the empirical observations provide useful guidelines for designing practical KGRL methods. Our contributions can be summarized as:
%\footnote{Our code and data are available at \url{https://anonymous.4open.science/r/kgbound-M2S5/}}

\begin{compactitem}
\item We propose a novel ReED framework representing various KGRL methods. Our RAMP encoder in ReED is a comprehensive neural encoder for KGRL that can express models such as CompGCN~\cite{comp} and R-GCN~\cite{rgcn}.
\item We formulate two types of triplet classification decoders in ReED to cover a wide range of KGRL methods; ReED can represent TransR~\cite{transr}, RotatE~\cite{rotate}, DistMult~\cite{dist}, ANALOGY~\cite{analogy}, and so forth.
\item We prove the generalization bounds for the ReED framework by unrolling two-step recursions for adequately modeling the interactions between relation and entity representations. Our work is the first study about PAC-Bayesian generalization bounds for KGRL.
\item We analyze our theoretical findings from a practical model design perspective and empirically show the relationships between the critical factors in the theoretical bounds and the actual generalization errors.\footnote{Our code and data are available at \url{https://github.com/bdi-lab/ReED}}
\end{compactitem}

\section{Knowledge Graph Completion via Triplet Classification}
\label{sec:prob}
Given a knowledge graph, the goal of knowledge graph completion is to add plausible triplets by finding appropriate combinations of the entities and relations. We consider a standard transductive knowledge graph completion~\cite{transe,rotate} where it is assumed that all entities and relations are observed during training, and a model predicts the plausibility of new combinations of the entities and relations. This can be viewed as a triplet classification problem, where a model determines whether a given triplet is true or false~\cite{survey}. For example, assume a knowledge graph contains entities, \texttt{Washington\_DC}, \texttt{USA}, and \texttt{Vienna}, and relations \texttt{Capital\_Of} and \texttt{Contains}. Consider two triplets (\texttt{USA}, \texttt{Contains}, \texttt{Washington\_DC}) and (\texttt{Vienna}, \texttt{Capital\_Of}, \texttt{USA}) that are missing in the given knowledge graph; the former triplet is true and the latter triplet is false. Like this, for new combinations of entities and relations, a triplet classification method predicts whether a given triplet is true or false and then adds only ones considered to be true for knowledge graph completion.

%consisting of entities and relations, knowledge graph completion aims to predict missing triplets by finding plausible combinations of the entities and relations. This knowledge graph completion task can be cast into a triplet classification problem that determines whether a given triplet is true or false, i.e., plausible or not~\cite{survey}. 

Let us consider a fully observed knowledge graph where the labels of all triplets are known and represent it as $G = (\sV, \xv, \sR, \xr, \sE)$ where $\sV$ is a set of entities, $\xv\in \RR^{\nv\times d_0}$ is a matrix of entity features, $d_0$ is the dimension of an entity feature vector, $\sR$ is a set of relations, $\xr\in \RR^{\nr\times d_0'}$ is a matrix of relation features, $d_0'$ is the dimension of a relation feature vector, $\sE\subseteq \sV\times\sR\times\sV$ such that $\sE\coloneqq\sE^+ \cup \sE^-$ where $\sE^+$ is a set of true triplets and $\sE^-$ is a set of false triplets. In practice, if the entity and relation features are unavailable, we can use one-hot encoding for $\xv$ and $\xr$. Each triplet $(h,r,t) \in \sE$ is associated with its label $y_{hrt}\in\{0,1\}$ such that $y_{hrt}=1$ if $(h,r,t)\in\sE^+$ and $y_{hrt}=0$ if $(h,r,t)\in\sE^-$. By sampling triplets from $\sE$ without replacement, we create a set of training triplets denoted by $\shE$. As a result, we create a training knowledge graph $\hG= (\sV, \xv, \sR, \xr, \shE)$. Note that $\sE$ is the full triplet set and $\shE$ is the training triplet set. A triplet classification method is trained with $\hG$ at training time and predicts $y_{hrt}$ of a triplet $(h,r,t)\in\sE$ at inference time.

%Let us consider a triplet classifier $f_{\vw} :\sV\times\sR\times\sV\rightarrow\RR^2$, where given a triplet $(h,r,t)\in\sE$, $f_{\vw}(h,r,t)[1]$ indicates the score for being a negative triplet and $f_{\vw}(h,r,t)[2]$ indicates the score for being a positive triplet.

\paragraph*{Notation} For an entity $v$ and a matrix $\mM$, let $\mM[v,:]$ denote the row of $\mM$ corresponding to $v$. Also, for a relation $r$, $\mM[r,:]$ indicates the row of $\mM$ corresponding to $r$. Given a triplet $(h,r,t)$, $\mM[t,h]$ indicates the element of $\mM$ at the row corresponding to $t$ and the column corresponding to $h$. Let a matrix with a superscript $(l)$ indicate the matrix at the $l$-th layer and $L$ be the total number of layers in the RAMP encoder. Let $d_l$ and $d'_l$ denote the dimension of entity and relation representations at the $l$-th layer, respectively. Given a vector $\vx$, \(\diag{\vx}\) is a diagonal matrix whose diagonal is $\vx$. Also, \(\mzero_{n_1\times n_2}\) is an all-zero matrix of size $n_1\times n_2$. All vectors are row vectors. More details are in Appendix~\ref{appn:not}.

%We denote a horizontal concatenation as \(\left[\begin{matrix}\cdot & \cdot \end{matrix}\right]\) and a vertical concatenation as \(\left[\begin{matrix}\cdot\\ \cdot\end{matrix}\right]\) for any vectors or matrices. 

%Also, \(\mzero_{n_1\times n_2}\) is an all-zero matrix of size $n_1\times n_2$.

\section{ReED Framework for Knowledge Graph Representation Learning}
\label{sec:reed}
Our ReED framework consists of the RAMP encoder and a triplet classification decoder, where we introduce two different types of decoders: the translational distance decoder and the semantic matching decoder.

%: the translational distance decoder and the semantic matching decoder. 

%These decoders can be either combined with the RAMP encoder or used standalone.

\subsection{Relation-Aware Message-Passing (RAMP) Encoder}
\label{sec:ramp}

\begin{table*}
  \centering
  \caption{Our RAMP encoder can represent R-GCN~\cite{rgcn}, WGCN~\cite{wgcn}, and CompGCN~\cite{comp} by appropriately setting the functions and matrices in Definition~\ref{def:ramp}.}\label{tb:encoder}
%  \vspace{-0.7cm}
  \begin{tabular*}{\linewidth}{@{\extracolsep{\fill}} rccccc }\\
  \toprule
   & $\phi$ & $\rho$, $\psi$ & $\mW_{r}^{(l)}$ & $\mU_{r}^{(l)}$& $\mS_r^{(l)}[v,:]$\\
  \midrule
  R-GCN~\cite{rgcn} & ReLU & identity & $\mW_{r}^{(l)}$ &$\mzero_{d_{l-1}'\times d_{l}}$& $\frac{1}{c_{v,r}}\mA_r[v,:]$\\
  WGCN~\cite{wgcn} & Tanh & identity  & $\alpha_r^{(l)}\mW_{0}^{(l)}$ &$\mzero_{d_{l-1}'\times d_{l}}$& $\mA_r[v,:]$ \\
  CompGCN (Sub)~\cite{comp} & Tanh & identity  & $\mW_{\lambda(r)}^{(l)}$ &$-\mW_{\lambda(r)}^{(l)}$& $\mA_r[v,:]$\\
  CompGCN (Mult)~\cite{comp} & Tanh & identity  & \(\diag{\mR^{(l-1)}[r,:]}\mW_{\lambda(r)}^{(l)}\) &$\mzero_{d_{l-1}'\times d_{l}}$& $\mA_r[v,:]$\\
  \bottomrule
  \end{tabular*}
\end{table*}

Many recently proposed KGRL methods employ GNNs~\cite{comp} or MPNNs~\cite{rgcn} to learn representations of entities by aggregating representations of the neighboring entities and relations. We formulate our RAMP encoder in Definition~\ref{def:ramp} to represent the aggregation process in a general form that can subsume many existing KGRL encoders.

\begin{definition}[RAMP Encoder for KGRL]\label{def:ramp}
Given a training knowledge graph $\hG= (\sV, \xv, \sR, \xr, \shE)$, for $l \in \{1,2,\ldots,L\}$, the RAMP encoder is defined by
{\small
  \begin{align}
    &\mH^{(0)} = \xv, \mR^{(0)} = \xr\notag\\
    &\mM_{r}^{(l)}[v,:]=\left[\begin{matrix}\mH^{(l-1)}[v,:] & \mR^{(l-1)}[r,:] \end{matrix}\right],\qquad v\in\sV, r\in \sR\notag\\
    &\mH^{(l)} = \phi\left(\mH^{(l-1)}\mW_{0}^{(l)}+\rho\left(\sum_{r\in\sR}\mS_r^{(l)}\psi(\mM_{r}^{(l)})\left[\begin{matrix}
      \mW_{r}^{(l)}\\ \mU_{r}^{(l)}
    \end{matrix}\right]\right)\right)\notag\\
    &\mR^{(l)} = \mR^{(l-1)}\mU_{0}^{(l)}\notag
  \end{align}
  }where $\mH^{(l)} \in \RR^{\nv\times d_l}$ is a matrix of entity representations, $\mR^{(l)} \in \RR^{\nr\times d'_l}$ is a matrix of relation representations, $\mM_{r}^{(l)} \in \RR^{\nv\times (d_{l-1}+d_{l-1}')}$ is a matrix for concatenating the entity and relation representations for each relation $r$, $\mW_{0}^{(l)}\in \RR^{d_{l-1}\times d_l}$, $\mW_{r}^{(l)}\in \RR^{d_{l-1}\times d_l}, \mU_{r}^{(l)}\in\RR^{d_{l-1}'\times d_l}$, and
  $\mU_{0}^{(l)} \in \RR^{d_{l-1}'\times d_l'}$ are learnable projection matrices, $\phi, \rho, \psi$ are Lipschitz-continuous activation functions with Lipschitz constants $C_\phi, C_\rho, C_\psi \geq 0$ and $\phi(\vzero)=\vzero, \rho(\vzero)=\vzero, \psi(\vzero)=\vzero$ where $\vzero$ is a zero vector, and $\mS_r^{(l)}\in\RR^{\nv\times\nv}$ is a graph diffusion matrix of $\hG$ for relation $r\in\sR$.
\end{definition}

In Definition~\ref{def:ramp}, an entity's representation is updated based on the entity and relation representations of its neighbors which are defined per relation using a relation-specific graph diffusion matrix $\mS_r^{(l)}$ for $r\in\sR$. The graph diffusion matrices are constructed by decoupling the training triplets based on the relations so that $\mS_r^{(l)}$ represents the connections between entities with relation $r$. A simple way to define $\mS_r^{(l)}$ is to consider an adjacency matrix $\mA_r\in\{0,1\}^{\nv\times\nv}$ for each relation $r$ such that $\mA_r[t,h]=1$ if $(h,r,t) \in \shE^+$, $\mA_r[t,h]=0$ otherwise. When $\mS_r^{(l)}[v,:]\coloneqq\mA_r[v,:]$, it implies that a model uses a sum aggregator in aggregating neighbors' representations. On the other hand, $\mS_r^{(l)}$ can also be set to a degree-normalized adjacency matrix, i.e., $\mS_r^{(l)}[v,:]\coloneqq \mA_r[v,:]/\dg(v)$ where $\dg(v)$ is the degree of $v$; this implies a model uses a mean aggregator.

%\footnote{Since a reverse triplet, $(t,r^{-1},h)$, is also considered during training~\cite{rgcn,comp}, the direction of the aggregation (from $h$ to $t$ vs. from $t$ to $h$) does not make a meaningful difference.}

%\footnote{In GNNs or MPNNs, a reverse triplet, $(t,r^{-1},h)$, is also considered during training~\cite{rgcn,comp}. Therefore, the direction of the aggregation (from $h$ to $t$ vs. from $t$ to $h$) does not make a meaningful difference.}

%In this context, the degree $\dg(v)$ can be defined by either the total number of neighbors or the number of neighbors having relation $r$. 

Several well-known GNN-based KGRL encoders can be considered as special cases of the RAMP encoder. For example, Table~\ref{tb:encoder} shows that our RAMP encoder can represent R-GCN~\cite{rgcn}, WGCN~\cite{wgcn}, and CompGCN~\cite{comp} with two different composition operators, subtraction (Sub) and multiplication (Mult). The key is to appropriately set the activation functions $\phi, \rho, \psi$, the projection matrices $\mW_{r}^{(l)}$ and $\mU_{r}^{(l)}$, and the graph diffusion matrices $\mS_r^{(l)}$. In Table~\ref{tb:encoder}, $c_{v,r}$ for R-GCN is a problem-specific normalization constant defined in~\cite{rgcn} and $c_{v,r}$ is usually set to be the number of neighbors of $v$ connected by $r$. Also, for WGCN, $\alpha_r^{(l)}$ is a relation-specific parameter~\cite{wgcn}. For CompGCN, $\lambda(r)$ is a function that categorizes the relation $r$ into one of the normal, inverse, and self connections defined in~\cite{comp}. Also, in CompGCN, the dimensions of the entity and relation representations should be the same for all layers. The RAMP encoder can also represent CompGCN with a circular correlation operator, which is omitted for brevity in Table~\ref{tb:encoder}. More details are described in Appendix~\ref{appn:ramp}.

\subsection{Triplet Classification Decoder}
\label{sec:tcd}

\begin{table*}
  \centering
  \caption{Translational Distance Decoder in Definition~\ref{def:td} can represent TransR~\cite{transr} and RotatE~\cite{rotate} and Semantic Matching Decoder in Definition~\ref{def:sm} can represent DistMult~\cite{dist} and ANALOGY~\cite{analogy} by appropriately setting the projection matrices.}\label{tb:decoder}
%  \vspace{-0.7cm}
    \begin{tabular*}{\linewidth}{clllllll}\\
  \toprule
  Decoder & Model & \multicolumn{6}{l}{Projection Matrices Setup}\\
  \midrule
  \multirow{2}{*}[-1ex]{\begin{tabular}{@{}c@{}}Translational\\Distance\end{tabular}} & TransR~\cite{transr} & \multicolumn{2}{l}{$\mbW_{r}^{\langle j \rangle}\coloneqq\mT_{\text{ent}}^{\langle j \rangle}\mF_{r}^{\langle j \rangle}$}&
  \multicolumn{2}{l}{$\mV_{r}^{\langle j \rangle}\coloneqq\mT_{\text{ent}}^{\langle j \rangle}\mF_{r}^{\langle j \rangle}$}&
  \multicolumn{2}{l}{$\mbU_{r}^{\langle j \rangle}\coloneqq\mT_{\text{rel}}^{\langle j \rangle}$}\\
  &RotatE~\cite{rotate} & \multicolumn{2}{l}{$\mbW_{r}^{\langle j \rangle}\coloneqq \mT_{\text{ent}}^{\langle j \rangle}\left[\begin{matrix}\mP_r^{\langle j \rangle} & \mQ_r^{\langle j \rangle}\\-\mQ_r^{\langle j \rangle} & \mP_r^{\langle j \rangle}\end{matrix}\right]$}&
  \multicolumn{2}{l}{$\mV_r^{\langle j \rangle}\coloneqq \mT_{\text{ent}}^{\langle j \rangle}$}&
  \multicolumn{2}{l}{$\mbU_r^{\langle j \rangle} \coloneqq  \mzero_{d'_{L}\times d_{L+1}}$}\\
  % \noalign{\vskip 0.5ex}
  % \hline\noalign{\vskip 0.5ex}
  \midrule
  \multirow{2}{*}{\begin{tabular}{@{}c@{}}Semantic\\Matching\end{tabular}} & DistMult~\cite{dist} & \multicolumn{6}{l}{$\mbU_r^{\langle j \rangle} \coloneqq \mT_{\text{ent}}^{\langle j \rangle} \left(\diag{\mR^{(L)}[r,:]\mT_{\text{rel}}^{\langle j \rangle}}\right){\mT_{\text{ent}}^{\langle j \rangle}}^\top$} \\
  & ANALOGY~\cite{analogy} & \multicolumn{2}{l}{$\mbU_r^{\langle j \rangle} \coloneqq \mT_{\text{ent}}^{\langle j \rangle}\mB_{r}^{\langle j \rangle}{\mT_{\text{ent}}^{\langle j \rangle}}^\top$ where}&
  \multicolumn{4}{l}{$\mB_{r_1}^{\langle j \rangle}{\mB_{r_2}^{\langle j \rangle}} = {\mB_{r_2}^{\langle j \rangle}}\mB_{r_1}^{\langle j \rangle}, \forall r_1, r_2 \in \sR$} \\
  \bottomrule
  \end{tabular*}
\end{table*}

%A triplet classification decoder computes the scores of each triplet to determine whether a given triplet is true or false. We design two different types of decoders depending on how the interactions between entity and relation representations are modeled. 
%These decoders can be either combined with the RAMP encoder or used standalone; ReED can represent not only GNN-based KGRL models but also shallow-architecture models.

Using the entity and relation representations returned by our RAMP encoder (i.e., $\mH^{(L)}$ and $\mR^{(L)}$), we design a triplet classification decoder to compute the scores of each triplet for determining whether a given triplet is true or false. While $\mH^{(L)}$ and $\mR^{(L)}$ are assumed to come from our RAMP encoder in general, i.e., $L>0$, we can also skip the RAMP encoder and directly apply our triplet classification decoder. When the RAMP encoder is bypassed, i.e., $L=0$, we assume $\mH^{(0)} \coloneqq \xv, \mR^{(0)} \coloneqq \xr\notag$.

Let $f_{\vw}:\sV\times\sR\times\sV\rightarrow\RR^2$ denote a triplet classifier with the parameters $\vw$. Given a triplet $(h,r,t)$, the classifier assigns two different scores, each of which is stored in $f_{\vw}(h,r,t)[0]$ and $f_{\vw}(h,r,t)[1]$, where the former is proportional to the likelihood of $(h,r,t)$ being false, and the latter is proportional to the likelihood of $(h,r,t)$ being true. 

Depending on how the interactions between entity and relation representations are modeled for computing the scores, we design two decoders: the translational distance (TD) decoder and the semantic matching (SM) decoder. The terms, `translational distance' and `semantic matching', have been also used in~\cite{survey}. In the TD decoder, the scores of $(h,r,t)$ are computed by the distance between $h$ and $t$ after a relation-specific translation is carried out. On the other hand, in the SM decoder, the score of $(h,r,t)$ is computed by the similarity between the individual components of the triplet. Definition~\ref{def:td} defines the TD decoder.

%Indeed, our two decoders generalize many existing knowledge graph embedding methods.

\begin{definition}[Translational Distance Decoder]\label{def:td}
For a triplet $(h,r,t)$, the TD decoder computes $f_{\vw}(h,r,t)[j]$ for $j\in \{0,1\}$ using the following formulation:
\begin{align}
  &f_{\vw}(h,r,t)[j] = \notag\\ &-\l2{\mH^{(L)}[h,:]\mbW_{r}^{\langle j \rangle}+\mR^{(L)}[r,:]\mbU_{r}^{\langle j \rangle}-\mH^{(L)}[t,:]\mV_{r}^{\langle j \rangle}}\notag
\end{align}
where $\mbW_{r}^{\langle j \rangle}, \mV_{r}^{\langle j \rangle}\in \RR^{d_{L}\times d_{L+1}}$ and $\mbU_{r}^{\langle j \rangle} \in \RR^{d_{L}'\times d_{L+1}}$ are learnable projection matrices and $d_{L+1}$ is the dimension of the final entity and relation representations.
\end{definition}

Note that $\mbW_{r}^{\langle j \rangle}$ and $\mV_{r}^{\langle j \rangle}$ carry out the relation-specific translation for the head and tail entity, respectively, whereas $\mbU_{r}^{\langle j \rangle}$ is a projection matrix for relations. When $L=0$, $\mH^{(0)}$ and $\mR^{(0)}$ are fixed to non-learnable matrices, $\xv$ and $\xr\notag$, respectively. By appropriately setting $\mbW_{r}^{\langle j \rangle}$, $\mV_{r}^{\langle j \rangle}$, and $\mbU_{r}^{\langle j \rangle}$ in Definition~\ref{def:td}, we can represent existing knowledge graph embedding methods as special cases of our TD decoder. Table~\ref{tb:decoder} shows how we should set the three projection matrices in Definition~\ref{def:td} to present TransR~\cite{transr} and RotatE~\cite{rotate}. To allow our decoder to express existing simple knowledge graph embedding methods having no encoder, we introduce two learnable matrices $\mT_{\text{\text{ent}}}^{\langle j \rangle}\in\RR^{d_{L}\times \bar{d}}$ and $\mT_{\text{\text{rel}}}^{\langle j \rangle}\in\RR^{d_{L}'\times \bar{d'}}$ which are only used for specializing our decoder to a particular existing model. When we do not need to simulate an existing model, we can simply drop $\mT_{\text{\text{ent}}}^{\langle j \rangle}$ and $\mT_{\text{\text{rel}}}^{\langle j \rangle}$. For TransR, the entity projection matrices are set to $\mT_{\text{ent}}^{\langle j \rangle}\mF_{r}^{\langle j \rangle}$ where $\mF_{r}^{\langle j \rangle}\in\RR^{\bar{d}\times d_{L+1}}$, and we set $\bar{d}'=d_{L+1}$. Also, the relation projection matrix is not defined per relation but shared across all relations. In RotatE, the embedding vectors are originally defined in a complex space~\cite{rotate}. To represent RotatE in our framework, we separately handle the real part and the imaginary part of an embedding vector and concatenate them to represent the whole embedding vector. Let us define $\mR^{(L)}[r,:]\mT_{\text{rel}}^{\langle j \rangle} \coloneqq \left[\begin{matrix}\vp_r^{\langle j \rangle} & \vq_r^{\langle j \rangle}\end{matrix}\right]$ where $\vp_r^{\langle j \rangle}$ indicates the real part and $\vq_r^{\langle j \rangle}$ indicates the imaginary part. Note that $(\vp_r^{\langle j \rangle}[i])^2+(\vq_r^{\langle j \rangle}[i])^2=1$ for $i\in\{0,1,\cdots,\text{$\frac{\bar{d'}}{2}$$-$$1$}\}$. Also, we define $\mP_r^{\langle j \rangle} \coloneqq \diag{\vp_r^{\langle j \rangle}}$ and $\mQ_r^{\langle j \rangle} \coloneqq \diag{\vq_r^{\langle j \rangle}}$, and set $\bar{d}=\bar{d'}=d_{L+1}$ to represent RotatE as a special case of our TD decoder. Definition~\ref{def:td} can also represent TransE~\cite{transe}, TransH~\cite{transh}, and PairRE~\cite{pairre}, which are described in Appendix~\ref{appn:td}. 

We also define the SM decoder in Definition~\ref{def:sm}.
\begin{definition}[Semantic Matching Decoder]\label{def:sm}
For a triplet $(h,r,t)$, the SM decoder computes $f_{\vw}(h,r,t)[j]$ for $j\in \{0,1\}$ using the following formulation:
\begin{align}
f_{\vw}(h,r,t)[j] = \mH^{(L)}[h,:]\mbU_{r}^{\langle j \rangle}{(\mH^{(L)}[t,:])}^\top\notag
\end{align}
where $\mbU_{r}^{\langle j \rangle}\in \RR^{d_{L}\times d_{L}}$ is a relation-specific learnable projection matrix.
\end{definition}

When $L>0$, the entity representation matrix $\mH^{(L)}$ is the output of our RAMP encoder and is multiplied to the relation-specific learnable projection matrix to score a triplet. Table~\ref{tb:decoder} shows that DistMult~\cite{dist} and ANALOGY~\cite{analogy} are special cases of Definition~\ref{def:sm}. For DistMult, we set $\bar{d}=\bar{d'}$ in $\mT_{\text{\text{ent}}}^{\langle j \rangle}$ and $\mT_{\text{\text{rel}}}^{\langle j \rangle}$. For ANALOGY, we define $\mB_r^{\langle j \rangle}\in\RR^{\bar{d}\times \bar{d}}$ which is a normal matrix. Definition~\ref{def:sm} can also represent RESCAL~\cite{rescal}, HolE~\cite{hole}, ComplEx~\cite{complex}, SimplE~\cite{simple}, and QuatE~\cite{quate}, which are described in Appendix~\ref{appn:sm}. 

\begin{figure}
  \centering
  \includegraphics[width=\columnwidth]{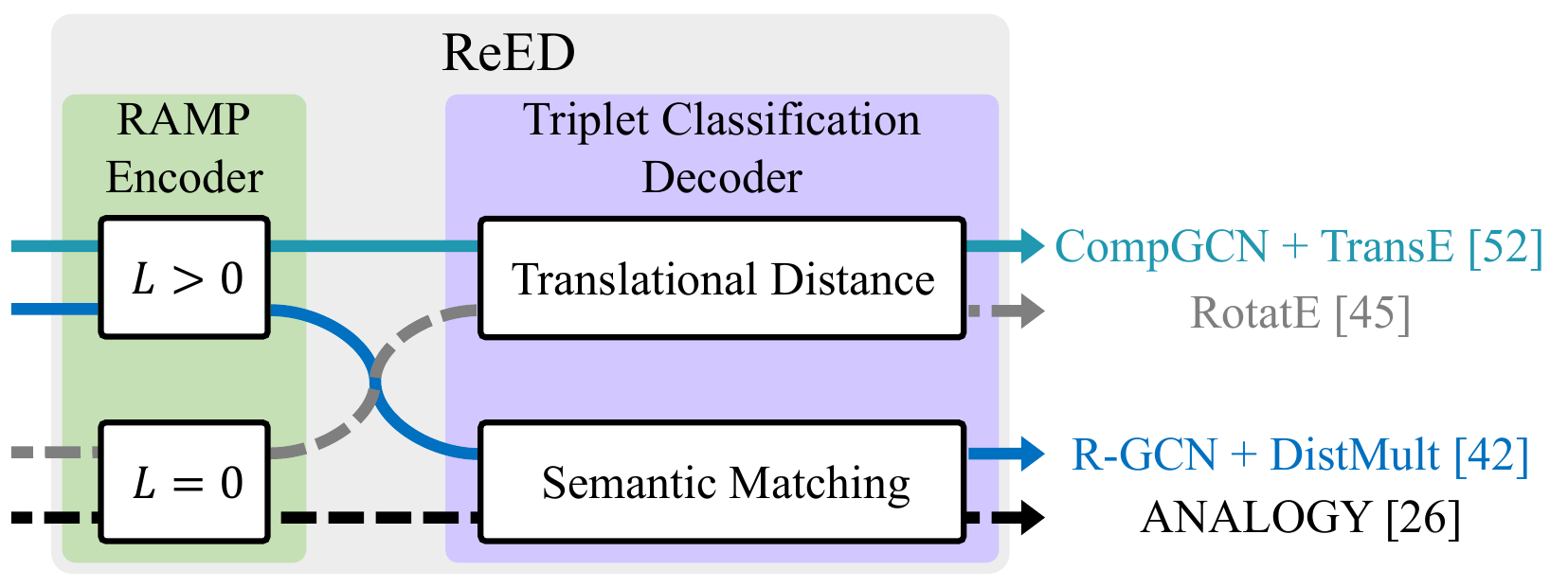}
%  \vspace{-0.5cm}
  \caption{Using different instantiations and combinations of the RAMP encoder and the triplet classification decoder, ReED can express many existing KGRL methods.}
  \label{fig:reed}
%  \vspace{-0.1cm}
\end{figure}

\subsection{Expressing Existing KGRL Methods Using ReED}
ReED can represent many existing KGRL methods ranging from simple shallow-architecture models~\cite{analogy,rotate} to neural encoder-based models~\cite{rgcn,comp}. In ReED, a triplet classification decoder can be either combined with the RAMP encoder (i.e., $L>0$) or used standalone (i.e., $L=0$). Also, the triplet classification decoder can be either the TD or SM decoder. Using different combinations of the RAMP encoder and the decoder, ReED can express various KGRL models, as illustrated in Figure~\ref{fig:reed}. While only four examples are presented in Figure~\ref{fig:reed}, ReED can express more diverse KGRL methods using different instantiations and configurations of the RAMP encoder (Section~\ref{sec:ramp}) and the triplet classification decoder (Section~\ref{sec:tcd}).

%Let us consider the case of $L>0$. Since \citeauthor{comp}~\cite{comp} use CompGCN as an encoder and TransE as a decoder, the model is a special case of our ReED framework. Also,~\citeauthor{rgcn}~\cite{rgcn} use R-GCN as an encoder and DistMult as a decoder, and thus, the model is also a special case of our ReED framework.

%When we consider the case of $L>0$, \citeauthor{comp}~\cite{comp} use CompGCN as an encoder and TransE as a decoder. Thus, the model proposed by~\citeauthor{comp}~\cite{comp} is a special case of our ReED framework. Also, the model by~\citeauthor{rgcn}~\cite{rgcn} uses R-GCN as an encoder and DistMult as a decoder; thus, this model can be considered a special case of our ReED framework. 

%Note that CompGCN is a special case of our RAMP encoder shown in Table~\ref{tb:encoder}.

\section{Generalization Bounds for ReED}
By proposing a PAC-Bayesian approach for deterministic triplet classifiers for knowledge graph completion, we present PAC-Bayesian generalization bounds of ReED. 

\subsection{Transductive PAC-Bayesian Approach for Deterministic Triplet Classifiers}\label{sec:transbound}
The PAC-Bayesian generalization bound relies on the KL divergence of a posterior distribution $\sQ$ on a hypothesis space from a prior distribution $\sP$ independent of the training set, where $\sP$ indicates prior knowledge about a given problem and $\sQ$ is learned by a learning algorithm~\cite{pac,pacmargin}. While the PAC-Bayesian approach was originally designed for analyzing stochastic models~\cite{pacmargin}, our triplet classifiers are deterministic models, i.e., the model parameters are fixed after training. Let us consider a deterministic triplet classifier $f_{\vw}(h,r,t)$ which assigns scores of the labels $0$ and $1$ for $(h,r,t)\in\sZ$, where $\sZ$ is a finite set of triplets. To gauge the risk of the classifier, a $\gamma$-margin loss is defined in Definition~\ref{def:margin}, where the loss is taken into account when the score of the ground-truth label $y_{hrt}$ of a triplet $(h,r,t)$ is less than or equal to that of the other label with the margin of $\gamma$. Note that the margin loss is one of the most commonly used loss functions in KGRL~\cite{survey}.

\begin{definition}[$\gamma$-Margin Loss of Triplet Classifier]\label{def:margin}
Given a finite triplet set $\sZ\subseteq\sV\times\sR\times\sV$, for any $\gamma > 0$ and a triplet classifier $f_{\vw}:\sV\times\sR\times\sV\rightarrow\RR^2$ with parameters $\vw$~that assigns scores for the label $0$ and $1$ for $(h,r,t)\in\sZ$, a $\gamma$-margin loss is defined as
{\small
\begin{align}
  & \sL_{\gamma,\sZ}(f_{\vw}) = \notag\\ &\frac{1}{|\sZ|}\sum_{(h,r,t)\in\sZ}\bone[f_{\vw}(h,r,t)[y_{hrt}]\leq\gamma+f_{\vw}(h,r,t)[1-y_{hrt}]] \notag
\end{align}
}where $f_{\vw}(h,r,t)[0]$ is the score for label $0$, $f_{\vw}(h,r,t)[1]$ is the score for label $1$, $y_{hrt}$ is the ground-truth label of $(h,r,t)$, and $\bone[\cdot]$ is an indicator function.
\end{definition}

When $\sZ$ is set to the training triplet set $\shE$, then $\sL_{\gamma, \shE}(f_\vw)$ is referred to as an empirical loss. On the other hand, a classification loss of a triplet classifier is defined as Definition~\ref{def:cls}.

\begin{definition}[Classification Loss of Triplet Classifier]
\label{def:cls}
Given a finite triplet set $\sZ\subseteq\sV\times\sR\times\sV$, for a triplet classifier $f_{\vw}:\sV\times\sR\times\sV\rightarrow\RR^2$ with parameters $\vw$~that assigns scores for the label $0$ and $1$ for $(h,r,t)\in\sZ$, the classification loss is defined as
{\small
\begin{align}
 & \sL_{0,\sZ}(f_{\vw}) = \notag\\ &\frac{1}{|\sZ|}\sum_{(h,r,t)\in\sZ}\bone[f_{\vw}(h,r,t)[y_{hrt}]\leq f_{\vw}(h,r,t)[1-y_{hrt}]] \notag
\end{align}
}
\end{definition}

When $\sZ$ is set to the full triplet set $\sE$, then $\sL_{0,\sE}(f_\vw)$ is referred to as an expected loss. The generalization bound is defined as the upper bound of the difference between the expected loss $\sL_{0,\sE}(f_{\vw})$ and the empirical loss $\sL_{\gamma,\shE}(f_{\vw})$. The generalization bound hints at the level of discrepancy in the model performance between the full and training sets.

As described in Section~\ref{sec:prob}, the full triplet set $\sE$ is finite, and the training triplets in $\shE$ are sampled from $\sE$ without replacement; thus, we need a transductive PAC-Bayesian analysis, which assumes the full set is finite. We derive Theorem~\ref{thm:transbound} from Corollary~7 in~\citeauthor{pactrans}~\cite{pactrans} by extending it to a deterministic triplet classifier. A main idea of the proof of Theorem~\ref{thm:transbound} is to add random perturbations~\cite{nnpac} to the fixed parameters $\vw$. The proof is available in Appendix~\ref{appn:transbound}.

%Assume that the prior distribution $\sP$ is independent of the training triplets.

\begin{theorem}[Transductive PAC-Bayesian Generalization Bound for a Deterministic Triplet Classifier]\label{thm:transbound}
Let \(f_{\vw}:\sV\times\sR\times\sV\rightarrow\RR^2\) be a deterministic triplet classifier with parameters $\vw$, and \(\sP\) be any prior distribution on $\vw$. Let us consider the finite full triplet set $\sE\subseteq\sV\times\sR\times\sV$. We construct a posterior distribution \(\sQ_{\vw+\ddot{\vw}}\) by adding any random perturbation \(\ddot{\vw}\) to \(\vw\) such that $\prob(\max_{(h,r,t)\in\sE}\linf{f_{\vw+\ddot{\vw}}(h,r,t)-f_{\vw}(h,r,t)} < \frac{\gamma}{4})>\frac{1}{2}$. Then, for any \(\gamma,\delta > 0\), with probability \(1-\delta\) over the choice of a training triplet set \(\shE\) drawn from the full triplet set \(\sE\) (such that $20\leq \nhE \leq \nE-20$ and $\nE \geq 40$) without replacement, for any \(\vw\), we have 
{\small
\begin{align}
&\sL_{0,\sE}(f_{\vw}) \leq \notag\\&\sL_{\gamma,\shE}(f_{\vw}) +\sqrt{\frac{1-\frac{\nhE}{\nE}}{2\nhE}\left[2D_{KL}(\sQ_{\vw+\ddot{\vw}}\|\sP)+\ln\frac{4\theta(\nhE,\nE)}{\delta}\right]}\notag
\end{align}
}where $\sL_{\gamma,\shE}(f_{\vw})$ is defined in Definition~\ref{def:margin}, $\sL_{0,\sE}(f_{\vw})$ is defined in Definition~\ref{def:cls}, \(D_{KL}(\sQ_{\vw+\ddot{\vw}}\|\sP)\) is the KL-divergence of $\sQ_{\vw+\ddot{\vw}}$ from $\sP$, and \(\theta(\nhE,\nE) = 3\sqrt{\nhE(1-\frac{\nhE}{\nE})}\ln{\nhE}\).
\end{theorem}

\subsection{PAC-Bayesian Generalization Bounds for ReED}
To present the PAC-Bayesian generalization bounds for ReED, we make the following assumptions:
\begin{enumerate}[label=A.\arabic*, nosep]
  \item\label{assump:a1} All activation functions $\phi,\rho,\psi$ in Definition~\ref{def:ramp} are Lipschitz-continuous with respect to the Euclidean norm of their input and output vectors, i.e., there exists a Lipschitz constant $C_g$ such that $\l2{g(\vx_1)-g(\vx_2)}\leq C_g\l2{\vx_1-\vx_2}$ for an activation function $g(\cdot)$ and any vectors $\vx_1$ and $\vx_2$. For example, ReLU, LeakyReLU, Tanh, SoftPlus, Sigmoid, ArcTan, and Softsign are Lipschitz continuous functions with Lipschitz constant $1$; these are 1-Lipschitz activation functions~\cite{lips}.
  \item\label{assump:a2} The training triplets in $\shE$ are sampled from the finite full triplet set $\sE$ without replacement.
  \item\label{assump:a3} Regarding the sizes of $\sE$ and $\shE$, we assume $\nE \geq 40$ and $20\leq \nhE \leq \nE -20$.
\end{enumerate}

Let us consider a triplet classification model that uses the RAMP encoder in Definition~\ref{def:ramp} and the TD decoder in Definition~\ref{def:td}. Using~Theorem~\ref{thm:transbound}, we compute the generalization bound of this model in Theorem~\ref{thm:boundtd}. Recall the graph diffusion matrix $\mS_r^{(l)}$ in Definition~\ref{def:ramp}. Given a fully observed knowledge graph $G$, the full triplet set $\sE$ is finite. Therefore, for every possible training knowledge graph $\hG$ and the corresponding training triplet set $\shE$ sampled from $\sE$, the infinity norms of the graph diffusion matrices for $\hG$ exist per relation. Let $k_r$ denote the maximum value of the infinity norms for all possible graph diffusion matrices among all layers for relation $r$. Note that $k_r$ is independent of the choice of the training triplet set and dependent on the full triplet set. 

\begin{theorem}[Generalization Bound for ReED with Translational Distance Decoder]\label{thm:boundtd}
  For any \(L\geq 0\), let \(f_{\vw} : \sV\times\sR\times\sV\rightarrow\RR^2\) be a triplet classifier designed by the combination of the RAMP encoder with $L$-layers in Definition~\ref{def:ramp} and the TD decoder in Definition~\ref{def:td}. Let $k_r$ be the maximum of the infinity norms for all possible $\mS_r^{(l)}$ in the RAMP encoder. Then, for any \(\delta, \gamma>0\), with probability at least $1-\delta$ over a training triplet set \(\shE\) (such that $20\leq \nhE \leq \nE -20$) sampled without replacement from the full triplet set \(\sE\), for any \(\vw\), we have
  {\small
  \begin{align*}
    &\sL_{0,\sE}(f_{\vw})\leq \sL_{\gamma,\shE}(f_{\vw})\notag\\ 
    &+\sO\!\left(\!\!\!\sqrt{\frac{1-\frac{\nhE}{\nE}}{\nhE}\left[\frac{N_{\vw} L^2\zeta_L^2 s^{2L}d\ln{(N_{\vw} d)}}{\gamma^2}+\ln\frac{\theta(\nhE,\nE)}{\delta}\right]}\right)\notag
  \end{align*}
  }where \(\theta(\nhE,\nE) = 3\sqrt{\nhE(1-\frac{\nhE}{\nE})}\ln{\nhE}, N_{\vw} =2\nr L+\)\\
  $6\nr+2L$, $\zeta_L = 2\tau^L\l2{\xv}+2\kappa\l2{\xr}\left(\sum_{i=0}^{L-1}\tau^i\right) + $\\
  $\l2{\xr}$, $\tau= C_\phi+\kappa, \kappa=C_\phi C_\rho C_\psi\sum_{r\in\sR}k_r, d=$\\
  $\max\left(\max_{0\leq l\leq L+1}(d_l),\max_{0\leq l\leq L+1}(d_l')\right), s_{L+1}=$\\
  $\max_{r,j}(\max(\lfr{\mbW_{r}^{\langle j \rangle}},\lfr{\mbU_{r}^{\langle j \rangle}},\lfr{\mV_{r}^{\langle j \rangle}}))$, $s_l =$\\
  $\max(\lfr{\mW_{0}^{(l)}}, \lfr{\mU_{0}^{(l)}},\max_r\lfr{\mW_{r}^{(l)}},\max_r\lfr{\mU_{r}^{(l)}})$ for $l \in \{1,2,\ldots,L\}$, and $s=\max_{1\leq l\leq L+1}(s_l)$.
\end{theorem}

In Theorem~\ref{thm:boundsm}, we also compute the generalization bound of a model that uses the RAMP encoder in Definition~\ref{def:ramp} and the SM decoder in Definition~\ref{def:sm}.

\begin{theorem}[Generalization Bound for ReED with Semantic Matching Decoder]\label{thm:boundsm}
  For any \(L\geq 0\), let \(f_{\vw} : \sV\times\sR\times\sV\rightarrow\RR^2\) be a triplet classifier designed by the combination of the RAMP encoder with $L$-layers in Definition~\ref{def:ramp} and the SM decoder in Definition~\ref{def:sm}. Let $k_r$ be the maximum of the infinity norms for all possible $\mS_r^{(l)}$ in the RAMP encoder. Then, for any \(\delta, \gamma>0\), with probability at least $1-\delta$ over a training triplet set \(\shE\) (such that $20\leq \nhE \leq \nE -20$) sampled without replacement from the full triplet set \(\sE\), for any \(\vw\), we have
  {\small
\begin{align*}
      & \sL_{0,\sE}(f_{\vw}) \leq \sL_{\gamma,\shE}(f_{\vw})\\
      &+\sO\!\left(\!\!\!\sqrt{\frac{1-\frac{\nhE}{\nE}}{\nhE}\left[\frac{N_{\vw} L^2\eta_{L}^4{s}^{4L}d\ln{(N_{\vw} d)}}{\gamma^2}+\ln\frac{\theta(\nhE,\nE)}{\delta}\right]}\right)\notag
    \end{align*}
    }where \(\theta(\nhE,\nE) = 3\sqrt{\nhE(1-\frac{\nhE}{\nE})}\ln{\nhE}, N_{\vw} = 2\nr L + \) \\
    $2\nr+2L$, $\eta_L = \tau^L\l2{\xv}+\kappa\l2{\xr}\sum_{i=0}^{L-1}\tau^i, d=$\\
    $\max(\max_{0\leq l\leq L}(d_l),\max_{0\leq l\leq L}(d_l')), \tau =C_\phi+\kappa,$\\
    $\kappa=C_\phi C_\rho C_\psi\sum_{r\in\sR}k_r$, $s_{L+1}=\max_{r,j}\lfr{\mbU_{r}^{\langle j \rangle}}, s_l = $\\
    $\max(\lfr{\mW_{0}^{(l)}}, \lfr{\mU_{0}^{(l)}},\max_r\lfr{\mW_{r}^{(l)}},\max_r\lfr{\mU_{r}^{(l)}})$\\
    for $l \in \{1,2,\ldots,L\}$, and $s=\max_{1\leq l\leq L+1}(s_l)$.
\end{theorem}

When we derive Theorem~\ref{thm:boundtd} and Theorem~\ref{thm:boundsm} from Theorem~\ref{thm:transbound}, we assume the prior distribution to be the Gaussian distribution with the zero mean and the standard deviation $\sigma$. We also assume that the perturbation follows the same Gaussian distribution as the prior distribution since the perturbation can follow any distribution, as indicated in Theorem~\ref{thm:transbound}. To derive the generalization bounds of ReED using Theorem~\ref{thm:transbound}, we need to find $\sigma$ such that $\prob(\max_{(h,r,t)\in\sE}\linf{f_{\vw+\ddot{\vw}}(h,r,t)-f_{\vw}(h,r,t)} < \frac{\gamma}{4})>\frac{1}{2}$ is satisfied, where $\ddot{\vw}$ follows the Gaussian distribution with the zero mean and the standard deviation of $\sigma$. In this process, we express $\linf{f_{\vw+\ddot{\vw}}(h,r,t)-f_{\vw}(h,r,t)}$ using our encoder (Definition~\ref{def:ramp}) and decoder (Definition~\ref{def:td} or Definition~\ref{def:sm}). Note that $\sigma$ should be independent of the learned parameters $\vw$ since $\sigma$ is the standard deviation of the prior distribution that should be independent of the training data. Thus, we use an approximation of the norm of $\vw$ instead of its actual norm when computing $\sigma$; the actual norm of $\vw$ is considered to be within a certain range from our approximation. Finally, we express the generalization bound in terms of the actual norm of $\vw$ using the covering number arguments~\cite{radenn}. The full proofs of Theorem~\ref{thm:boundtd} and Theorem~\ref{thm:boundsm} are in Appendix~\ref{appn:mainbound}.

%When we derive Theorem~\ref{thm:boundtd} and Theorem~\ref{thm:boundsm} from Theorem~\ref{thm:transbound}, we assume the same Gaussian distribution for both the prior and perturbation distributions. Let $\sigma$ denote the standard deviation of the Gaussian distribution.  
%Even though we follow some proof techniques of~\citeauthor{gcpac}~\cite{gcpac}, our proofs require much more complex derivations than~\citeauthor{gcpac}~\cite{gcpac}, mainly due to the relation representations, which should be taken into account for every step and makes the recursion step more tricky.  

To prove Theorem~\ref{thm:boundtd} and Theorem~\ref{thm:boundsm}, we follow some proof techniques of~\citeauthor{gcpac}~\cite{gcpac} considering standard graphs with a single relation. However, our proofs require more complex derivations than~\citeauthor{gcpac}~\cite{gcpac} because knowledge graphs have multiple relations. In our proofs, the interactions between the entities and the relations should be considered in the recurrence relation of an entity representation, leading to two-step unrolling of each recursion step. As a result, the generalization bounds are computed by considering (i) the norm of the difference between an unperturbed entity representation and a perturbed entity representation, (ii) the norm of the difference between an unperturbed relation representation and a perturbed relation representation, (iii) the norm of an entity representation, and (iv) the norm of a relation representation.

When we use one-hot encoding for $\xv$ and $\xr$, the spectral norms of $\xv$ and $\xr$ are one, and the maximum dimension becomes $d=|\sV|$. Let us assume that $\phi,\rho,\psi$ are 1-Lipschitz activation functions. In this case, we can simplify $\zeta_L=4\tau^L-1$ in Theorem~\ref{thm:boundtd} and $\eta_L=2\tau^L-1$ in Theorem~\ref{thm:boundsm}. In Corollary~\ref{cor:simp}, we present a simplified form of our generalization bounds by leaving model-dependent terms and regarding the rest as a constant.

%to deliver some high-level implications.

\begin{corollary}[Simplified Form of the Generalization Bounds for ReED]\label{cor:simp}
  For any \(L\geq 0\), let \(f_{\vw} : \sV\times\sR\times\sV\rightarrow\RR^2\) be a triplet classifier with the combination of the RAMP encoder with $L$-layers in Definition~\ref{def:ramp} and a decoder defined in Definition~\ref{def:td} or Definition~\ref{def:sm}. Let $k_r$ be the maximum of the infinity norms for all possible $\mS_r^{(l)}$ in the RAMP encoder. Then for any \(\delta, \gamma>0\), with probability at least $1-\delta$ over a training triplet set \(\shE\) (such that $20\leq \nhE \leq \nE -20$) 
  drawn without replacement from the full triplet set \(\sE\), for any \(\vw\), we have
 \begin{align}
  &\sL_{0,\sE} (f_{\vw})\leq\sL_{\gamma,\shE}(f_{\vw})+ \notag\\
  &\begin{cases}
      \displaystyle
      \sO\left(L{\tau}^{L}s^{L}\sqrt{N_{\vw}\ln{N_{\vw}}}\right) & (\textup{Translational Distance})\\
      \\\displaystyle
      \sO\left(L{\tau}^{2L}s^{2L}\sqrt{N_{\vw}\ln{N_{\vw}}}\right) & (\textup{Semantic Matching})\\
    \end{cases}\notag  
  \end{align}
  where $s$ is the maximum of the Frobenius norms of all learnable matrices and $\tau=1+\sum_{r\in\sR}k_r$, and $N_{\vw}$ is the total number of learnable matrices.
\end{corollary}

In Corollary~\ref{cor:simp}, the bounds are largely affected by $k_r$ which is the maximum infinity norm of all possible graph diffusion matrices across all layers for relation $r$. Recall two different ways of defining a graph diffusion matrix $\mS_r^{(l)}$ discussed in Section~\ref{sec:ramp}: an adjacency matrix (corresponding to a sum aggregator) or a degree-normalized adjacency matrix (corresponding to a mean aggregator). Note that $k_r$ becomes the maximum degree of an entity per relation when the sum aggregator is used, whereas $k_r$ becomes at most one when the mean aggregator is used since each row is normalized. Thus, a mean aggregator can be a better option than a sum aggregator in reducing the generalization bounds. The total number of learnable matrices $N_{\vw}$ is another critical factor: the generalization bounds decrease when the number of parameters is reduced. This can explain the effectiveness of the parameter-sharing strategies in \citeauthor{comp}~\cite{comp} and the basis or block decomposition ideas in \citeauthor{rgcn}~\cite{rgcn}. On the other hand, the maximum of the Frobenius norms of learnable matrices $s$ also critically affects the generalization bounds. Therefore, the generalization bounds decrease when the weight matrices or the entity/relation representations are normalized. This observation can provide theoretical justification for weight normalization adapted in~\citeauthor{wnorm}~\cite{wnorm} and normalization of entity representations used in~\citeauthor{transe}~\cite{transe}. In Corollary~\ref{cor:simp}, $\tau s$ is usually greater than one because $s$ is typically not less than one~\cite{difpac}. Thus, the generalization bounds sharply increase when the number of encoder layers $L$ increases.

%the maximum value of the infinity norms for all possible graph diffusion matrices among all layers for relation $r$. 
%This provides a theoretical explanation of the need for normalizing the graph diffusion matrix. Let us consider two different graph diffusion matrices $\mS_r$, an adjacency matrix (corresponding to a sum aggregator) or a degree-normalized adjacency matrix (corresponding to a mean aggregator), discussed in Section~\ref{sec:ramp}. 

\section{Experiments}
\label{sec:exp}
%\subparagraph*{Datasets and Experimental Setup} 

We conduct experiments on three real-world knowledge graphs: \fb~\cite{fb}, \cde~\cite{codex}, and \umls~\cite{umls,umls43}. Note that \fb and \cde are well-known knowledge graph benchmarks extracted from commonly used knowledge bases, Freebase and Wikidata, respectively, and \umls is another benchmark extracted from a popular biomedical knowledge base, UMLS. On all datasets, we create $\shE$ randomly sampled from $\sE$ without replacement with the sampling probability of $0.8$. In the ReED framework, we use the RAMP encoder in Definition~\ref{def:ramp} with $L$ layers and use either the TD or SM decoder (Definition~\ref{def:td} and Definition~\ref{def:sm}). In the RAMP encoder, we use $\rho=\psi=\text{identity}$ and $\phi=\text{LeakyReLU}$. We use one-hot encoding for $\xv$ and $\xr$. More details about the datasets and the settings are in Appendix~\ref{appn:exp}.

\begin{figure*}
\centering
\includegraphics[height=9.8cm]{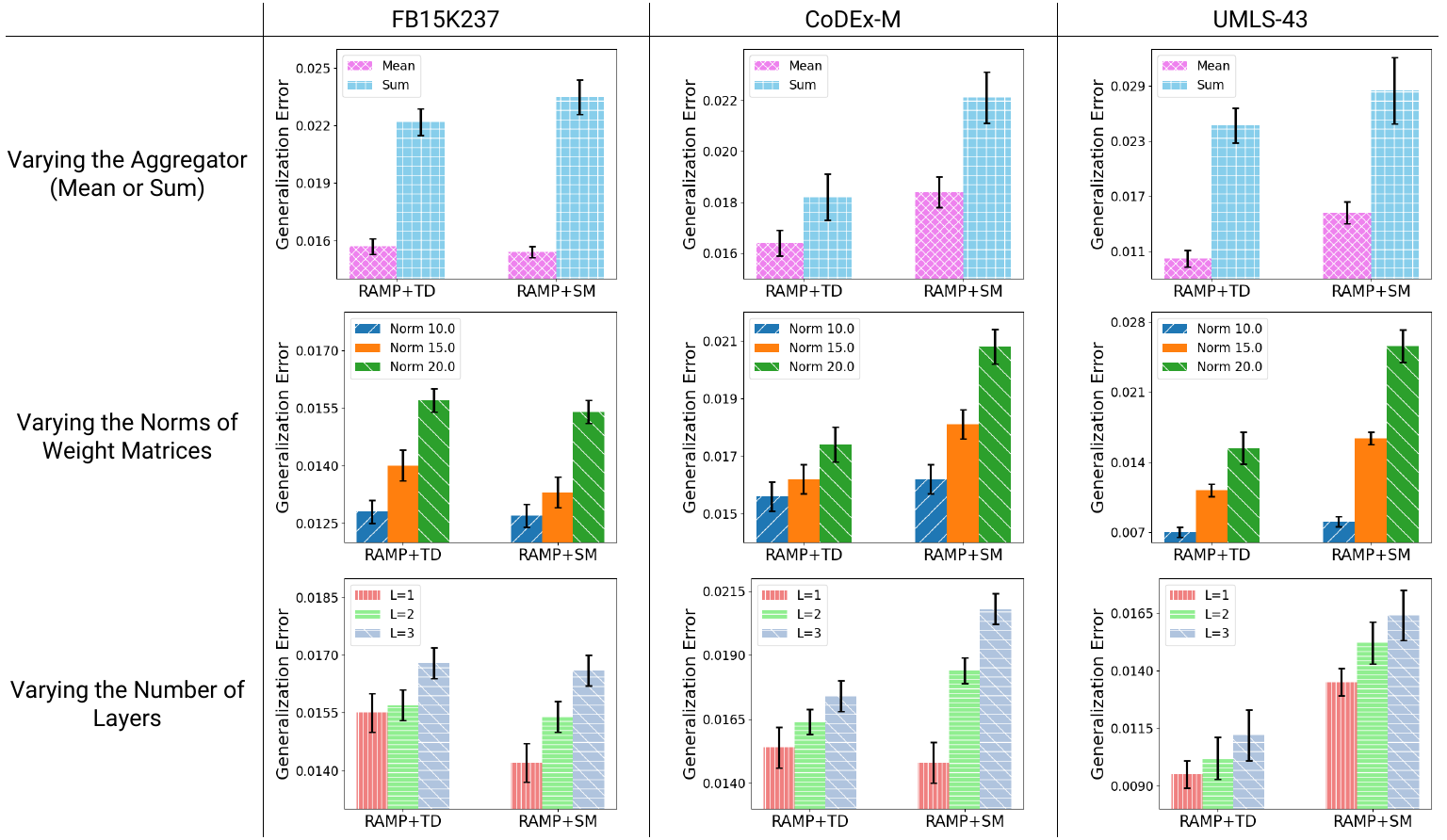}
\caption{Generalization Errors of ReED according to different aggregators, norms of the weight matrices, and numbers of layers in the RAMP encoder. In ReED, two different triplet classification decoders, TD or SM, are used. The changing trends in generalization errors according to the three different factors align with the theoretical findings in Corollary~\ref{cor:simp}.}
\label{fig:exp}
\end{figure*}

%Varying the Aggregator (Mean or Sum)
%Varying the Norms of Weight Matrices
%Varying the Number of Layers

%The mean aggregator shows lower generalization errors than the sum aggregator. Also, the generalization errors increase as the norms of the weight matrices and the number of encoder layers increase. Corollary~\ref{cor:simp}

%\subparagraph*{Experimental Results} 

We measure the generalization errors on real-world datasets, where a generalization error is the actual difference between the expected and empirical losses empirically observed in a particular experiment; the generalization bound is the theoretical upper bound of these generalization errors. In Corollary~\ref{cor:simp}, among the factors that affect the generalization bounds, we empirically measure the effects of the following three factors: (i) whether a model uses a mean aggregator or a sum aggregator, which affects $k_r$, (ii) the Frobenius norms of the learnable matrices $s$, and (iii) the number of layers $L$ in the RAMP encoder\footnote{Note that $N_\vw$ is proportional to $L$ since $N_\vw = \sO(\nr L)$ as indicated in Theorem~\ref{thm:boundtd} and Theorem~\ref{thm:boundsm}.}. We compare the generalization errors by varying one of these three factors while the other two factors are controlled. Note that $s$ is defined as the maximum of Frobenius norms of all learnable matrices. Indeed, to control the effect of the norms of these matrices more precisely, the Frobenius norm of each weight matrix should be fixed to be $s$ as described in the proof of our generalization bounds in Appendix~\ref{appn:mainbound}. Thus, we normalize each weight matrix after each backpropagation step. We use $s\in\{10.0,15.0,20.0\}$ and $L\in\{1,2,3\}$ for all datasets. Figure~\ref{fig:exp} shows the generalization errors of ReED depending on the decoder: RAMP+TD and RAMP+SM. We repeat all experiments 10 times and visualize the mean and the standard deviation. Across all datasets and all models, the mean aggregator shows lower generalization errors than the sum aggregator. Also, the generalization errors increase as $s$ and $L$ increase. These empirical observations are aligned with our theoretical findings in Corollary~\ref{cor:simp}. Even though the theoretical generalization bound indicates the upper limit of the possible generalization errors, Figure~\ref{fig:exp} shows that the critical factors explaining the generalization bounds also affect an actual generalization error.

\begin{figure}[t]
\centering
\includegraphics[height = 4.5cm]{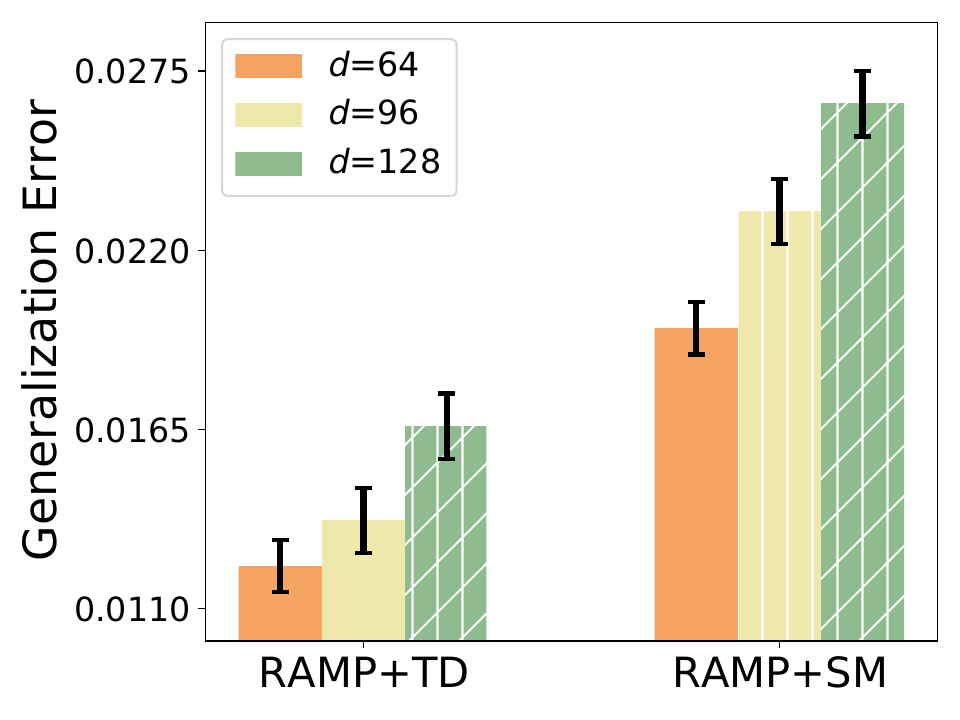}
\caption{Generalization Errors of ReED on FB15K237 according to different maximum dimensions $d$.}
\label{fig:expd}
\end{figure}

We conduct additional experiments using the initial features of entities and relations instead of using one-hot encoding for $\xv$ and $\xr$ on \fb to observe the generalization errors by varying $d$. We extract the initial features by feeding the textual descriptions of entities and relations to BERT~\cite{bert} and reduce the dimension of the extracted features to 32 using PCA; we use the resulting features as $\mathbf{X}_\text{ent}$ and $\mathbf{X}_\text{rel}$. Then, we calculate the generalization errors of ReED according to different maximum dimensions $d$ while fixing the other factors (e.g., the aggregator, the norms of weight matrices, and the number of layers). Figure~\ref{fig:expd} shows the generalization errors of ReED with different $d( = d_1 =d_2 =\cdots = d_L = d_{L+1} = d'_1=d'_2=\cdots=d'_L=d'_{L+1}) \in \{64,96,128\}$. We observe that the generalization errors increase as $d$ increases, which aligns with the expected tendency.

%First, we compare the generalization errors depending on a mean or sum aggregator by varying the number of layers $L$, where $s$ is fixed. Second, we compare the generalization errors according to different $s$ and varying $L$, where the aggregator is fixed to a mean aggregator.   
%To measure the empirical loss, we use the margin of $\gamma=0.5$. Also, to sufficiently train the models, we run all models for 2,000 epochs. 

 %This implies that our theoretical generalization bounds can provide useful guidelines for practical design choices in KGRL methods.

%Therefore, we believe that our analysis on the theoretical generalization bounds is useful for designing practically effective triplet classifiers for knowledge graphs.

%\begin{table}[t]
%  \centering
%  \small
%  \setlength{\tabcolsep}{0.55em}
%  \caption{Generalization Trends of ReED on FB15K237 by varying $d$}\label{tb:txt}
%    \begin{tabular}{cccc}\\
%  \toprule
% & d = 64 & d = 96 & d = 128\\
% ReED+TD & $0.0123{\scriptscriptstyle\pm 0.0008}$& $0.0137{\scriptscriptstyle\pm 0.0010}$& $0.0166{\scriptscriptstyle\pm 0.0006}$\\
% ReED+SM & $0.0196{\scriptscriptstyle\pm 0.0010}$& $0.0232{\scriptscriptstyle\pm 0.0009}$& $0.0265{\scriptscriptstyle\pm 0.0010}$\\
%  \bottomrule
%  \end{tabular}
%\end{table}

\section{Related Work and Discussion}
Regarding the generalization ability of knowledge graph embedding,~\citeauthor{kgmle}~\cite{kgmle} have computed the expected number of incorrect predictions made by knowledge graph embedding methods, which differs from a standard generalization bound defined by the difference between the expected and the empirical errors; their work is neither applicable to GNN-based models nor a margin loss. 

While there have been some studies about the generalization bounds for GNNs~\cite{gcpac,difpac,randbound}, they have considered graph classification tasks on standard graphs with a single relation. For example,~\citeauthor{gcpac}~\cite{gcpac} assume that graphs are i.i.d. samples drawn from some unknown infinite distribution. Also,~\citeauthor{difpac}~\cite{difpac} consider a twice-differentiable loss function to compute Hessian-based bounds, and~\citeauthor{randbound}~\cite{randbound} apply MPNNs on the underlying continuous space from which graphs are sampled. Our work and these previous works~\cite{gcpac,difpac,randbound} are significantly distinct in that (i) we deal with knowledge graphs having multiple relations, (ii) our target task is a triplet classification, (iii) we assume a finite full set since the triplets are finite given a fixed knowledge graph while the previous studies~\cite{gcpac,difpac,randbound} assume that graphs are sampled from an infinite space.

Different PAC-Bayes approaches have been explored in various perspectives. For example,~\citeauthor{pacprimer}~\cite{pacprimer} provides a survey about the PAC-Bayes framework, including the extention of the KL divergence to $f$-divergence for expressing a more general divergence class in computing PAC-Bayes bounds. Also,~\citeauthor{pacptuto}~\cite{pacptuto} provides a recent survey about various tight PAC-Bayes bounds in varied settings. Though our study focuses on the traditional KL divergence and considers the transductive PAC-Bayesian approach, we expect our work to be extended to a broader class of divergences or information-theoretic approaches.

%~\citeauthor{gcpac}~\cite{gcpac} have recently proposed the generalization bounds for graph neural networks, their analysis has been conducted on graph classification tasks on standard graphs under an inductive learning setting. On the other hand, our analysis is conducted on triplet classification tasks on knowledge graphs under a transductive learning setting. \textcolor{red}{\cite{randbound}}

\section{Conclusion and Future Work}
To comprehensively analyze the generalization bounds for KGRL, we propose a generic framework, ReED, that can subsume many existing KGRL methods. We prove the PAC-Bayesian generalization bounds for ReED having two different triplet classification decoders. Our analysis provides theoretical evidence for the benefits of the parameter-sharing and weight normalization schemes and the advantage of a mean aggregator over a sum aggregator within a neural encoder in reducing the generalization bounds in KGRL.

We note that the ReED framework cannot exhaustively cover all existing KGRL methods. Specifically, the graph attention networks~\cite{kbat,gat,gatv2} are hard to consider in ReED with the current form. Extending ReED to the attention mechanisms is one of our future works. Also, we plan to investigate the relationships between the generalization ability and the expressivity~\cite{wlvc} in KGRL based on our findings in the generalization bounds of KGRL.

%\paragraph*{Limitations and Future Work}
%The proposed ReED framework cannot exhaustively cover the knowledge graph representation learning methods. Specifically, the graph attention networks~\cite{gat,gatv2} are hard to be considered in ReED with the current form. Also, we will extend our work to inductive link prediction settings~\cite{grail} to deal with real-world knowledge graphs more efficiently. Finally, we also plan to compute the lower bound of the generalization error~\cite{difpac} of knowledge graph representation learning methods to approximate the actual model performance more tightly.

\section*{Impact Statement}
Most of our contributions in this paper are theoretical, and our work aims to advance the field of Machine Learning at a fundamental level. Considering that knowledge graphs are widely utilized in information retrieval (e.g., Google Knowledge Graph), a societal consequence of our work is to improve the retrieval performance by providing theoretical insights for KGRL methods. Our findings and their practical implications can guide the desirable designs of future KGRL methods. Generally speaking, our generalization bounds indicate that reducing the number of learnable parameters, the norms of weight matrices, and the maximum infinity norm of the graph diffusion matrices is beneficial to decreasing the generalization bounds.

%This paper presents work whose goal is to advance the field of Machine Learning. There are many potential societal consequences of our work, none which we feel must be specifically highlighted here.

\section*{Acknowledgements}
This research was supported by an NRF grant funded by
MSIT 2022R1A2C4001594 (Extendable Graph Representation Learning) and an IITP grant funded by MSIT 2022-0-
00369 (Development of AI Technology to support Expert
Decision-making that can Explain the Reasons/Grounds for
Judgment Results based on Expert Knowledge).

\bibliography{gbound_bib}

\begin{thebibliography}{62}
\providecommand{\natexlab}[1]{#1}
\providecommand{\url}[1]{\texttt{#1}}
\expandafter\ifx\csname urlstyle\endcsname\relax
  \providecommand{\doi}[1]{doi: #1}\else
  \providecommand{\doi}{doi: \begingroup \urlstyle{rm}\Url}\fi

\bibitem[Alquier(2024)]{pacptuto}
Alquier, P.
\newblock User-friendly introduction to pac-bayes bounds.
\newblock \emph{Foundations and Trends® in Machine Learning}, 17\penalty0
  (2):\penalty0 174--303, 2024.

\bibitem[Barcelo et~al.(2022)Barcelo, Galkin, Morris, and Orth]{rwl}
Barcelo, P., Galkin, M., Morris, C., and Orth, M.~R.
\newblock {Weisfeiler} and {Leman} go relational.
\newblock In \emph{Proceedings of the 1st Learning on Graphs Conference}, pp.\
  46:1--46:26, 2022.

\bibitem[Bartlett \& Mendelson(2002)Bartlett and Mendelson]{rade}
Bartlett, P.~L. and Mendelson, S.
\newblock Rademacher and {G}aussian complexities: Risk bounds and structural
  results.
\newblock \emph{Journal of Machine Learning Research}, 3:\penalty0 463--482,
  2002.

\bibitem[Bartlett et~al.(2017)Bartlett, Foster, and Telgarsky]{radenn}
Bartlett, P.~L., Foster, D.~J., and Telgarsky, M.~J.
\newblock Spectrally-normalized margin bounds for neural networks.
\newblock In \emph{Proceedings of the 31st Conference on Neural Information
  Processing Systems}, pp.\  6240--6249, 2017.

\bibitem[Bartlett et~al.(2019)Bartlett, Harvey, Liaw, and Mehrabian]{vcnn}
Bartlett, P.~L., Harvey, N., Liaw, C., and Mehrabian, A.
\newblock Nearly-tight {VC}-dimension and pseudodimension bounds for piecewise
  linear neural networks.
\newblock \emph{Journal of Machine Learning Research}, 20\penalty0
  (63):\penalty0 1--17, 2019.

\bibitem[B{\'e}gin et~al.(2014)B{\'e}gin, Germain, Laviolette, and
  Roy]{pactrans}
B{\'e}gin, L., Germain, P., Laviolette, F., and Roy, J.-F.
\newblock {PAC}-{B}ayesian theory for transductive learning.
\newblock In \emph{Proceedings of the 17th International Conference on
  Artificial Intelligence and Statistics}, pp.\  105--113, 2014.

\bibitem[Bodenreider(2004)]{umls}
Bodenreider, O.
\newblock The unified medical language system ({UMLS}): integrating biomedical
  terminology.
\newblock \emph{Nucleic Acids Research}, 32\penalty0 (suppl\_1):\penalty0
  D267--D270, 2004.

\bibitem[Bordes et~al.(2013)Bordes, Usunier, Garcia-Duran, Weston, and
  Yakhnenko]{transe}
Bordes, A., Usunier, N., Garcia-Duran, A., Weston, J., and Yakhnenko, O.
\newblock Translating embeddings for modeling multi-relational data.
\newblock In \emph{Proceedings of the 27th Conference on Neural Information
  Processing Systems}, pp.\  2787--2795, 2013.

\bibitem[Brody et~al.(2022)Brody, Alon, and Yahav]{gatv2}
Brody, S., Alon, U., and Yahav, E.
\newblock How attentive are graph attention networks?
\newblock In \emph{Proceedings of the 10th International Conference on Learning
  Representations}, 2022.

\bibitem[Chao et~al.(2021)Chao, He, Wang, and Chu]{pairre}
Chao, L., He, J., Wang, T., and Chu, W.
\newblock {PairRE}: Knowledge graph embeddings via paired relation vectors.
\newblock In \emph{Proceedings of the 59th Annual Meeting of the Association
  for Computational Linguistics and the 11th International Joint Conference on
  Natural Language Processing (Volume 1: Long Papers)}, pp.\  4360--4369, 2021.

\bibitem[Chung et~al.(2023)Chung, Lee, and Whang]{hynt}
Chung, C., Lee, J., and Whang, J.~J.
\newblock Representation learning on hyper-relational and numeric knowledge
  graphs with transformers.
\newblock In \emph{Proceedings of the 29th ACM SIGKDD Conference on Knowledge
  Discovery and Data Mining}, pp.\  310--322, 2023.

\bibitem[Devlin et~al.(2019)Devlin, Chang, Lee, and Toutanova]{bert}
Devlin, J., Chang, M.-W., Lee, K., and Toutanova, K.
\newblock {BERT}: Pre-training of deep bidirectional transformers for language
  understanding.
\newblock In \emph{Proceedings of the 2019 Conference of the North American
  Chapter of the Association for Computational Linguistics: Human Language
  Technologies, Volume 1 (Long and Short Papers)}, pp.\  4171--4186, 2019.

\bibitem[Esser et~al.(2021)Esser, Vankadara, and Ghoshdastidar]{ncrade}
Esser, P.~M., Vankadara, L.~C., and Ghoshdastidar, D.
\newblock Learning theory can (sometimes) explain generalisation in graph
  neural networks.
\newblock In \emph{Proceedings of the 35th Conference on Neural Information
  Processing Systems}, pp.\  27043--27056, 2021.

\bibitem[Garg et~al.(2020)Garg, Jegelka, and Jaakkola]{gcrade}
Garg, V.~K., Jegelka, S., and Jaakkola, T.
\newblock Generalization and representational limits of graph neural networks.
\newblock In \emph{Proceedings of the 37th International Conference on Machine
  Learning}, pp.\  3419--3430, 2020.

\bibitem[Gilmer et~al.(2017)Gilmer, Schoenholz, Riley, Vinyals, and Dahl]{mpnn}
Gilmer, J., Schoenholz, S.~S., Riley, P.~F., Vinyals, O., and Dahl, G.~E.
\newblock Neural message passing for quantum chemistry.
\newblock In \emph{Proceedings of the 34th International Conference on Machine
  Learning}, pp.\  1263--1272, 2017.

\bibitem[Guedj(2019)]{pacprimer}
Guedj, B.
\newblock A primer on pac-bayesian learning.
\newblock In \emph{arXiv preprint}, 2019.

\bibitem[Huang et~al.(2023)Huang, Orth, Ceylan, and Barceló]{lpwl}
Huang, X., Orth, M.~R., Ceylan, {\.I}.~{\.I}., and Barceló, P.
\newblock A theory of link prediction via relational {Weisfeiler-Leman} on
  knowledge graphs.
\newblock In \emph{Proceedings of the 37th Conference on Neural Information
  Processing Systems}, 2023.

\bibitem[Ju et~al.(2023)Ju, Li, Sharma, and Zhang]{difpac}
Ju, H., Li, D., Sharma, A., and Zhang, H.~R.
\newblock Generalization in graph neural networks: Improved {PAC}-{B}ayesian
  bounds on graph diffusion.
\newblock In \emph{Proceedings of the 26th International Conference on
  Artificial Intelligence and Statistics}, pp.\  6314--6341, 2023.

\bibitem[Kazemi \& Poole(2018)Kazemi and Poole]{simple}
Kazemi, S.~M. and Poole, D.
\newblock {SimplE} embedding for link prediction in knowledge graphs.
\newblock In \emph{Proceedings of the 32nd Conference on Neural Information
  Processing Systems}, pp.\  4284--4295, 2018.

\bibitem[Kingma \& Ba(2015)Kingma and Ba]{adam}
Kingma, D.~P. and Ba, J.~L.
\newblock Adam: A method for stochastic optimization.
\newblock In \emph{Proceedings of the 3rd International Conference on Learning
  Representations}, 2015.

\bibitem[Kuželka \& Wang(2019)Kuželka and Wang]{kgmle}
Kuželka, O. and Wang, Y.
\newblock Generalization bounds for knowledge graph embedding (trained by
  maximum likelihood).
\newblock NeurIPS 2019 Workshop on Machine Learning with Guarantees, 2019.

\bibitem[Lee et~al.(2023{\natexlab{a}})Lee, Chung, Lee, Jo, and Whang]{vista}
Lee, J., Chung, C., Lee, H., Jo, S., and Whang, J.~J.
\newblock {VISTA}: Visual-textual knowledge graph representation learning.
\newblock In \emph{Findings of the Association for Computational Linguistics:
  EMNLP 2023}, pp.\  7314--7328, 2023{\natexlab{a}}.

\bibitem[Lee et~al.(2023{\natexlab{b}})Lee, Chung, and Whang]{ingram}
Lee, J., Chung, C., and Whang, J.~J.
\newblock {InGram}: Inductive knowledge graph embedding via relation graphs.
\newblock In \emph{Proceedings of the 40th International Conference on Machine
  Learning}, pp.\  18796--18809, 2023{\natexlab{b}}.

\bibitem[Liao et~al.(2021)Liao, Urtasun, and Zemel]{gcpac}
Liao, R., Urtasun, R., and Zemel, R.
\newblock A {PAC}-{B}ayesian approach to generalization bounds for graph neural
  networks.
\newblock In \emph{Proceedings of the 9th International Conference on Learning
  Representations}, 2021.

\bibitem[Lin et~al.(2015)Lin, Liu, Sun, Liu, and Zhu]{transr}
Lin, Y., Liu, Z., Sun, M., Liu, Y., and Zhu, X.
\newblock Learning entity and relation embeddings for knowledge graph
  completion.
\newblock In \emph{Proceedings of the 29th AAAI Conference on Artificial
  Intelligence}, pp.\  2181--2187, 2015.

\bibitem[Liu et~al.(2017)Liu, Wu, and Yang]{analogy}
Liu, H., Wu, Y., and Yang, Y.
\newblock Analogical inference for multi-relational embeddings.
\newblock In \emph{Proceedings of the 34th International Conference on Machine
  Learning}, pp.\  2168--2178, 2017.

\bibitem[Lloyd et~al.(2023)Lloyd, Liu, and Gaunt]{umls43}
Lloyd, O., Liu, Y., and Gaunt, T.~R.
\newblock Assessing the effects of hyperparameters on knowledge graph embedding
  quality.
\newblock \emph{Journal of Big Data}, 10\penalty0 (1):\penalty0 59, 2023.

\bibitem[Ma et~al.(2021)Ma, Deng, and Mei]{subpac}
Ma, J., Deng, J., and Mei, Q.
\newblock Subgroup generalization and fairness of graph neural networks.
\newblock In \emph{Proceedings of the 35th Conference on Neural Information
  Processing Systems}, pp.\  1048--1061, 2021.

\bibitem[Maskey et~al.(2022)Maskey, Levie, Lee, and Kutyniok]{randbound}
Maskey, S., Levie, R., Lee, Y., and Kutyniok, G.
\newblock Generalization analysis of message passing neural networks on large
  random graphs.
\newblock In \emph{Proceedings of the 36th Conference on Neural Information
  Processing Systems}, pp.\  4805--4817, 2022.

\bibitem[McAllester(2003)]{pacmargin}
McAllester, D.
\newblock Simplified {PAC}-{B}ayesian margin bounds.
\newblock In \emph{Proceedings of the 16th Annual Conference on Computational
  Learning Theory and 7th Kernel Workshop}, pp.\  203--215, 2003.

\bibitem[McAllester(1998)]{pac}
McAllester, D.~A.
\newblock Some {PAC}-{B}ayesian theorems.
\newblock In \emph{Proceedings of the 11th Annual Conference on Computational
  Learning Theory}, pp.\  230--234, 1998.

\bibitem[Morris et~al.(2023)Morris, Geerts, Tönshoff, and Grohe]{wlvc}
Morris, C., Geerts, F., Tönshoff, J., and Grohe, M.
\newblock {WL} meet {VC}.
\newblock In \emph{Proceedings of the 40th International Conference on Machine
  Learning}, pp.\  25275--25302, 2023.

\bibitem[Nathani et~al.(2019)Nathani, Chauhan, Sharma, and Kaul]{kbat}
Nathani, D., Chauhan, J., Sharma, C., and Kaul, M.
\newblock Learning attention-based embeddings for relation prediction in
  knowledge graphs.
\newblock In \emph{Proceedings of the 57th Annual Meeting of the Association
  for Computational Linguistics}, pp.\  4710--4723, 2019.

\bibitem[Neyshabur et~al.(2017)Neyshabur, Bhojanapalli, McAllester, and
  Srebro]{gendl}
Neyshabur, B., Bhojanapalli, S., McAllester, D., and Srebro, N.
\newblock Exploring generalization in deep learning.
\newblock In \emph{Proceedings of the 31st Conference on Neural Information
  Processing Systems}, pp.\  5947--5956, 2017.

\bibitem[Neyshabur et~al.(2018)Neyshabur, Bhojanapalli, and Srebro]{nnpac}
Neyshabur, B., Bhojanapalli, S., and Srebro, N.
\newblock A {PAC}-{B}ayesian approach to spectrally-normalized margin bounds
  for neural networks.
\newblock In \emph{Proceedings of the 6th International Conference on Learning
  Representations}, 2018.

\bibitem[Nickel et~al.(2011)Nickel, Tresp, and Kriegel]{rescal}
Nickel, M., Tresp, V., and Kriegel, H.-P.
\newblock A three-way model for collective learning on multi-relational data.
\newblock In \emph{Proceedings of the 28th International Conference on Machine
  Learning}, pp.\  809--816, 2011.

\bibitem[Nickel et~al.(2016)Nickel, Rosasco, and Poggio]{hole}
Nickel, M., Rosasco, L., and Poggio, T.
\newblock Holographic embeddings of knowledge graphs.
\newblock In \emph{Proceedings of the 30th AAAI Conference on Artificial
  Intelligence}, pp.\  1955--1961, 2016.

\bibitem[Oono \& Suzuki(2020{\natexlab{a}})Oono and Suzuki]{radetrans}
Oono, K. and Suzuki, T.
\newblock Optimization and generalization analysis of transduction through
  gradient boosting and application to multi-scale graph neural networks.
\newblock In \emph{Proceedings of the 34th Conference on Neural Information
  Processing Systems}, pp.\  18917--18930, 2020{\natexlab{a}}.

\bibitem[Oono \& Suzuki(2020{\natexlab{b}})Oono and Suzuki]{wnorm}
Oono, K. and Suzuki, T.
\newblock Graph neural networks exponentially lose expressive power for node
  classification.
\newblock In \emph{Proceedings of the 8th International Conference on Learning
  Representations}, 2020{\natexlab{b}}.

\bibitem[Safavi \& Koutra(2020)Safavi and Koutra]{codex}
Safavi, T. and Koutra, D.
\newblock {C}o{DE}x: A comprehensive knowledge graph completion benchmark.
\newblock In \emph{Proceedings of the 2020 Conference on Empirical Methods in
  Natural Language Processing}, pp.\  8328--8350, 2020.

\bibitem[Scarselli et~al.(2018)Scarselli, Tsoi, and Hagenbuchner]{gcvc}
Scarselli, F., Tsoi, A.~C., and Hagenbuchner, M.
\newblock The {Vapnik-Chervonenkis} dimension of graph and recursive neural
  networks.
\newblock \emph{Neural Networks}, 108:\penalty0 248--259, 2018.

\bibitem[Schlichtkrull et~al.(2018)Schlichtkrull, Kipf, Bloem, van~den Berg,
  Titov, and Welling]{rgcn}
Schlichtkrull, M., Kipf, T.~N., Bloem, P., van~den Berg, R., Titov, I., and
  Welling, M.
\newblock Modeling relational data with graph convolutional networks.
\newblock In \emph{Proceedings of the 15th Extended Semantic Web Conference},
  pp.\  593--607, 2018.

\bibitem[Shang et~al.(2019)Shang, Tang, Huang, Bi, He, and Zhou]{wgcn}
Shang, C., Tang, Y., Huang, J., Bi, J., He, X., and Zhou, B.
\newblock End-to-end structure-aware convolutional networks for knowledge base
  completion.
\newblock In \emph{Proceedings of the 33rd AAAI Conference on Artificial
  Intelligence}, pp.\  3060--3067, 2019.

\bibitem[Socher et~al.(2013)Socher, Chen, Manning, and Ng]{ntn}
Socher, R., Chen, D., Manning, C.~D., and Ng, A.~Y.
\newblock Reasoning with neural tensor networks for knowledge base completion.
\newblock In \emph{Proceedings of the 27th Conference on Neural Information
  Processing Systems}, pp.\  926--934, 2013.

\bibitem[Sun et~al.(2019)Sun, Deng, Nie, and Tang]{rotate}
Sun, Z., Deng, Z.-H., Nie, J.-Y., and Tang, J.
\newblock {RotatE}: Knowledge graph embedding by relational rotation in complex
  space.
\newblock In \emph{Proceedings of the 7th International Conference on Learning
  Representations}, 2019.

\bibitem[Teru et~al.(2020)Teru, Denis, and Hamilton]{grail}
Teru, K.~K., Denis, E.~G., and Hamilton, W.~L.
\newblock Inductive relation prediction by subgraph reasoning.
\newblock In \emph{Proceedings of the 37th International Conference on Machine
  Learning}, pp.\  9448--9457, 2020.

\bibitem[Toutanova \& Chen(2015)Toutanova and Chen]{fb}
Toutanova, K. and Chen, D.
\newblock Observed versus latent features for knowledge base and text
  inference.
\newblock In \emph{Proceedings of the 3rd Workshop on Continuous Vector Space
  Models and their Compositionality}, pp.\  57--66, 2015.

\bibitem[Tropp(2012)]{rmat}
Tropp, J.~A.
\newblock User-friendly tail bounds for sums of random matrices.
\newblock \emph{Foundations of Computational Mathematics}, 12\penalty0
  (4):\penalty0 389--434, 2012.

\bibitem[Trouillon et~al.(2016)Trouillon, Welbl, Riedel, Gaussier, and
  Bouchard]{complex}
Trouillon, T., Welbl, J., Riedel, S., Gaussier, {\'E}., and Bouchard, G.
\newblock Complex embeddings for simple link prediction.
\newblock In \emph{Proceedings of the 33rd International Conference on Machine
  Learning}, pp.\  2071--2080, 2016.

\bibitem[Valiant(1984)]{paclearn}
Valiant, L.~G.
\newblock A theory of the learnable.
\newblock \emph{Communications of the ACM}, 27\penalty0 (11):\penalty0
  1134--1142, 1984.

\bibitem[Vapnik \& Chervonenkis(1971)Vapnik and Chervonenkis]{vc}
Vapnik, V.~N. and Chervonenkis, A.~Y.
\newblock On the uniform convergence of relative frequencies of events to their
  probabilities.
\newblock \emph{Theory of Probability and Its Applications}, 16\penalty0
  (2):\penalty0 264--280, 1971.

\bibitem[Vashishth et~al.(2020)Vashishth, Sanyal, Nitin, and Talukdar]{comp}
Vashishth, S., Sanyal, S., Nitin, V., and Talukdar, P.
\newblock Composition-based multi-relational graph convolutional networks.
\newblock In \emph{Proceedings of the 8th International Conference on Learning
  Representations}, 2020.

\bibitem[Veli{\v{c}}kovi{\'c} et~al.(2018)Veli{\v{c}}kovi{\'c}, Cucurull,
  Casanova, Romero, Li{\`o}, and Bengio]{gat}
Veli{\v{c}}kovi{\'c}, P., Cucurull, G., Casanova, A., Romero, A., Li{\`o}, P.,
  and Bengio, Y.
\newblock Graph attention networks.
\newblock In \emph{Proceedings of the 6th International Conference on Learning
  Representations}, 2018.

\bibitem[Virmaux \& Scaman(2018)Virmaux and Scaman]{lips}
Virmaux, A. and Scaman, K.
\newblock Lipschitz regularity of deep neural networks: analysis and efficient
  estimation.
\newblock In \emph{Proceedings of the 32nd Conference on Neural Information
  Processing Systems}, pp.\  3835--3844, 2018.

\bibitem[Wang et~al.(2017)Wang, Mao, Wang, and Guo]{survey}
Wang, Q., Mao, Z., Wang, B., and Guo, L.
\newblock Knowledge graph embedding: A survey of approaches and applications.
\newblock \emph{IEEE Transactions on Knowledge and Data Engineering},
  29\penalty0 (12):\penalty0 2724--2743, 2017.

\bibitem[Wang et~al.(2014)Wang, Zhang, Feng, and Chen]{transh}
Wang, Z., Zhang, J., Feng, J., and Chen, Z.
\newblock Knowledge graph embedding by translating on hyperplanes.
\newblock In \emph{Proceedings of the 28th AAAI Conference on Artificial
  Intelligence}, pp.\  1112--1119, 2014.

\bibitem[Weisfeiler \& Lehman(1968)Weisfeiler and Lehman]{wl}
Weisfeiler, B. and Lehman, A.~A.
\newblock A reduction of a graph to a canonical form and an algebra arising
  during this reduction.
\newblock \emph{Nauchno-Technicheskaya Informatsia}, 2\penalty0 (9):\penalty0
  12--16, 1968.

\bibitem[Wu et~al.(2021)Wu, Pan, Chen, Long, Zhang, and Yu]{gnnsurv}
Wu, Z., Pan, S., Chen, F., Long, G., Zhang, C., and Yu, P.~S.
\newblock A comprehensive survey on graph neural networks.
\newblock \emph{IEEE Transactions on Neural Networks and Learning Systems},
  32\penalty0 (1):\penalty0 4--24, 2021.

\bibitem[Xu et~al.(2019)Xu, Hu, Leskovec, and Jegelka]{gin}
Xu, K., Hu, W., Leskovec, J., and Jegelka, S.
\newblock How powerful are graph neural networks?
\newblock In \emph{Proceedings of the 7th International Conference on Learning
  Representations}, 2019.

\bibitem[Yang et~al.(2015)Yang, Yih, He, Gao, and Deng]{dist}
Yang, B., Yih, W.-t., He, X., Gao, J., and Deng, L.
\newblock Embedding entities and relations for learning and inference in
  knowledge bases.
\newblock In \emph{Proceedings of the 3rd International Conference on Learning
  Representations}, 2015.

\bibitem[Zhang et~al.(2019)Zhang, Tay, Yao, and Liu]{quate}
Zhang, S., Tay, Y., Yao, L., and Liu, Q.
\newblock Quaternion knowledge graph embeddings.
\newblock In \emph{Proceedings of the 33rd Conference on Neural Information
  Processing Systems}, pp.\  2735--2745, 2019.

\bibitem[Zhou et~al.(2022)Zhou, Kutyniok, and Ribeiro]{oodlp}
Zhou, Y., Kutyniok, G., and Ribeiro, B.
\newblock {OOD} link prediction generalization capabilities of message-passing
  {GNNs} in larger test graphs.
\newblock In \emph{Proceedings of the 36th Conference on Neural Information
  Processing Systems}, pp.\  20257--20272, 2022.

\end{thebibliography}
\bibliographystyle{icml2024}

%%%%%%%%%%%%%%%%%%%%%%%%%%%%%%%%%%%%%%%%%%%%%%%%%%%%%%%%%%%%%%%%%%%%%%%%%%%%%%%
%%%%%%%%%%%%%%%%%%%%%%%%%%%%%%%%%%%%%%%%%%%%%%%%%%%%%%%%%%%%%%%%%%%%%%%%%%%%%%%
% APPENDIX
%%%%%%%%%%%%%%%%%%%%%%%%%%%%%%%%%%%%%%%%%%%%%%%%%%%%%%%%%%%%%%%%%%%%%%%%%%%%%%%
%%%%%%%%%%%%%%%%%%%%%%%%%%%%%%%%%%%%%%%%%%%%%%%%%%%%%%%%%%%%%%%%%%%%%%%%%%%%%%%
\newpage
\appendix
\onecolumn

\section{Basic Notation}
\label{appn:not}
In Table~\ref{tb:notation}, we provide a concise overview of the notation used throughout the paper. Any other notation not listed in Table~\ref{tb:notation} is clarified and detailed within the context.

\begin{table}[hbt!]
  \centering
  \caption{Overview of basic notation}
  \label{tb:notation}
  \begin{tabular}{ccccccc}
    \toprule
    & Symbol & Meaning\\
    \midrule
    & $[\begin{matrix}\cdot&\cdot\end{matrix}],\left[\begin{matrix}\cdot\\\cdot\end{matrix}\right]$ & a horizontal/vertical concatenation \\
    & $|\cdot|$ & the size of a set or the absolute value of a scalar value \\
    & $\l2{\cdot},\linf{\cdot}, \lfr{\cdot}$ & the spectral(Euclidean)/infinity/Frobenius norm of a matrix(vector) \\
    & $\diag{\vx}$ & a diagonal matrix whose diagonal is defined by the vector $\vx$ \\
    & $\mI_{n\times n}$ & an identity matrix of size $n \times n$ \\
    & $\mzero_{m\times n}$ & an all-zero matrix of size $m \times n$ \\
    & $\mzero_{n}$ & an all-zero vector of size $n$ \\
    & $\bone[\cdot]$ & an indicator function \\
    & $\prob[\cdot],\EE[\cdot]$ & the probability/expectation \\
    & $\phi(\cdot),\rho(\cdot),\psi(\cdot)$ & Lipschitz-continuous activation functions \\
    & $C_{\phi},C_{\rho},C_{\psi}$ & Lipschitz constants of $\phi, \rho, \psi$ \\
    & $\mM^\top$ & the transpose of a matrix $\mM$ \\
    & $G, \hG$ & a fully observed/training knowledge graph \\
    & $\sV,\sR$ & a set of entities/relations \\
    & $\sE,\shE$ & a full triplet set/training triplet set \\
    & $\xv,\xr$ & a matrix of entity/relation features \\
    & $d_0,d'_0$ & the dimension of the initial feature vector of an entity/relation\\
    & $d_l,d'_l$ & the dimension of an entity/relation representation at the $l$-th layer\\
    & $L$ & the total number of layers in the RAMP encoder \\
    & $y_{hrt}$ & the ground-truth label of a triplet $(h,r,t) \in \sE$ \\
    & $\mH,\mR$ & a matrix of entity/relation representations \\
    & $\mW$, $\mU$, $\mV$& the learnable projection matrices \\
    & $\mS_r$ & the graph diffusion matrix of $\hG$ for relation $r\in \sR$\\
    & $D_{KL}(\sQ\|\sP)$ & KL-divergence of $\sQ$ from $\sP$ \\
    & $\sP,\sQ$ & a prior/posterior distribution on a hypothesis space $\sH$ \\
%    & $\gamma$ & the margin in the margin loss function \\
    & $\ln(\cdot)$ & the natural logarithm \\
    \bottomrule
  \end{tabular}
\end{table}

\section{Interpreting ReED as a Generalization of Existing KGRL Methods}
\label{appn:reed}
Our ReED framework consists of the RAMP encoder and a triplet classification decoder as described in Section~{\ref{sec:reed}}. We provide the details about how our RAMP encoder and two types of triplet classification decoders (i.e., translational distance decoder and semantic matching decoder) can express diverse KGRL methods.

\subsection{Representing Existing KGRL Encoders Using RAMP Encoder}
\label{appn:ramp}
In Section~{\ref{sec:ramp}}, we define the RAMP encoder in Definition~{\ref{def:ramp}} and show that several well-known GNN-based KGRL encoders can be considered as special cases of our RAMP encoder. For example, R-GCN~\cite{rgcn}, WGCN~\cite{wgcn}, and CompGCN~\cite{comp} can be represented using the RAMP encoder by appropriately setting the activation functions $\phi, \rho, \psi$, the projection matrices $\mW_{r}^{(l)}$ and $\mU_{r}^{(l)}$, and the graph diffusion matrices $\mS_r^{(l)}$ in Definition~{\ref{def:ramp}}, as shown in Table~\ref{tb:encoder}. 

Note that CompGCN has three different variations depending on the composition operator: subtraction (Sub), multiplication (Mult), and circular correlation (Corr). We detail how CompGCN (Corr) can be represented using the RAMP encoder here; the others are all described in the main paper. Give an entity representation $\mH^{(l)}[v,:]$ and a relation representation $\mR^{(l)}[r,:]$ where both have the dimension of $d$, the circular correlation $\star$ is defined by $(\mH^{(l)}[v,:]\star\mR^{(l)}[r,:])[k] \coloneqq \sum_{i=0}^{d-1}\mH^{(l)}[v,i]\mR^{(l)}[r,\ \text{$($$k$$+$$i$$)$}\!\!\mod d]$ where $k\in\{0,1,...,\text{$d$$-$$1$}\}$. Let us define $\mC_r^{(l)}$ as
\begin{align*}
  \mC_r^{(l)} \coloneqq & \left[\begin{matrix}
    \mR^{(l)}[r,0] & \mR^{(l)}[r,1] & \cdots & \mR^{(l)}[r,\text{$d_l'$$-$$1$}]\\
    \mR^{(l)}[r,1] & \mR^{(l)}[r,2]& \cdots & \mR^{(l)}[r,0]\\
    \vdots& \vdots &\ddots & \vdots\\
    \mR^{(l)}[r,\text{$d_l'$$-$$1$}] & \mR^{(l)}[r,0] & \cdots & \mR^{(l)}[r,\text{$d_l'$$-$$2$}]\\
  \end{matrix}\right].
\end{align*}

By setting $\phi$ to Tanh, $\rho, \psi$ to the identity functions, \(\mW_{r}^{(l)}\coloneqq\mC_r^{(l-1)}\mW_{\lambda(r)}^{(l)}\), $\mU_{r}^{(l)}\coloneqq\mzero_{d_{l-1}'\times d_{l}}$, and $\mS_r^{(l)}[v,:]\coloneqq\mA_r[v,:]$, we can represent CompGCN (Corr) using our RAMP encoder. Note that the dimensions of the entity and relation representations should be the same for all layers in CompGCN.

\subsection{Translational Distance Decoder and Existing KGRL Methods}
\label{appn:td}
We define our translational distance (TD) decoder in Definition~{\ref{def:td}}, which includes three learnable projection matrices: $\mbW_{r}^{\langle j \rangle}, \mV_{r}^{\langle j \rangle}$ and $\mbU_{r}^{\langle j \rangle}$. By appropriately defining these matrices, our TD decoder can express five different knowledge graph embedding methods: TransE~\cite{transe}, TransH~\cite{transh}, TransR~\cite{transr}, RotatE~\cite{rotate}, and PairRE~\cite{pairre}. Since we already described TransR and RotatE in the main paper, we describe the other three methods here. Recall that we introduce two learnable matrices, $\mT_{\text{\text{ent}}}^{\langle j \rangle}\in\RR^{d_{L}\times \bar{d}}$ and $\mT_{\text{\text{rel}}}^{\langle j \rangle}\in\RR^{d_{L}'\times \bar{d'}}$, which are only needed for specializing our decoder to simulate an existing shallow-architecture knowledge graph embedding model, as described in Section~{\ref{sec:tcd}}.

\paragraph*{TransE~\cite{transe}}
For each triplet, TransE assumes that the heady entity representation is translated by the relation representation, and the resulting vector should be placed close to the tail entity representation. 
Our Definition~{\ref{def:td}} can trivially express TransE by setting
\[\mbW_{r}^{\langle j \rangle}\coloneqq\mT_{\text{ent}}^{\langle j \rangle}\qquad 
\mV_{r}^{\langle j \rangle}\coloneqq\mT_{\text{ent}}^{\langle j \rangle} \qquad
\mbU_{r}^{\langle j \rangle} \coloneqq \mT_{\text{rel}}^{\langle j \rangle}\]
where $\bar{d}=\bar{d'}=d_{L+1}$. Note that TransE has a constraint that $\l2{\mH^{(L)}[v,:]\mT_{\text{ent}}^{\langle j \rangle}}=1$, $\forall v \in \sV$ for $j\in\{0,1\}$.

\paragraph*{TransH~\cite{transh}}
In TransH, each entity representation vector is projected onto a relation-specific hyperplane, where a projected head entity is translated by the relation representation, and the resulting vector is assumed to be placed close to the projected tail entity representation. For relation $r$, let $\vf_r^{\langle j \rangle}\in\RR^{\bar{d}}$ denote the unit normal vector of the hyperplane for $r$. Note that $\l2{\vf_r^{\langle j \rangle}}=1$ for $j\in\{0,1\}$ and $\bar{d}=\bar{d'}=d_{L+1}$. By Setting
\[\mbW_{r}^{\langle j \rangle}\coloneqq\mT_{\text{ent}}^{\langle j \rangle}(\mI_{\bar{d}\times\bar{d}}-{\vf_r^{\langle j \rangle}}^\top\vf_r^{\langle j \rangle})\qquad
  \mV_{r}^{\langle j \rangle}\coloneqq\mT_{\text{ent}}^{\langle j \rangle}(\mI_{\bar{d}\times\bar{d}}-{\vf_r^{\langle j \rangle}}^\top\vf_r^{\langle j \rangle}) \qquad
  \mbU_{r}^{\langle j \rangle} \coloneqq \mT_{\text{rel}}^{\langle j \rangle},\]
our TD decoder can represent TransH.

\paragraph*{PairRE~\cite{pairre}}
In PairRE, a relation representation comprises two parts: representations for the head and the tail. The head and tail entity representations are translated by the relation representation corresponding to the head and tail part, respectively. The translated entity representations are assumed to be close to each other. Let $\mathfrak{f}_{r}^{\langle j \rangle}\in\RR^{\bar{d'}}$ denote the representation of $r$ for translating head entities and $\mathring{\mathfrak{f}}_{r}^{\langle j \rangle}\in\RR^{\bar{d'}}$ denote the representation of $r$ for translating tail entities. Let us define $\mat{\mathfrak{F}}_{r}^{\langle j \rangle}=\diag{\mathfrak{f}_{r}^{\langle j \rangle}}$ and $\mathring{\mat{\mathfrak{F}}}_{r}^{\langle j \rangle}=\diag{\mathring{\mathfrak{f}}_{r}^{\langle j \rangle}}$. It is assumed that the entity representations are on the unit circle, i.e., $\l2{\mH^{(L)}[v,:]\mT_{\text{ent}}^{\langle j \rangle}}=1$, $\forall v \in \sV$ for $j\in\{0,1\}$. Given $\bar{d}=\bar{d'}=d_{L+1}$, we can represent PairRE by setting
\[\mbW_{r}^{\langle j \rangle}\coloneqq\mT_{\text{ent}}^{\langle j \rangle}\mat{\mathfrak{F}}_{r}^{\langle j \rangle}\qquad
 \mV_{r}^{\langle j \rangle}\coloneqq\mT_{\text{ent}}^{\langle j \rangle}\mathring{\mat{\mathfrak{F}}}_{r}^{\langle j \rangle}\qquad
 \mbU_r^{\langle j \rangle} \coloneqq \mzero_{d'_L\times d_{L+1}}.\]

%Since the relation representation is separately learned for the head and tail, we have $\bar{d'}=2\bar{d}$.

\subsection{Semantic Matching Decoder and Existing KGRL Methods}
\label{appn:sm}
Our semantic matching (SM) decoder in Definition~{\ref{def:sm}} can represent seven different knowledge graph embedding methods: RESCAL~\cite{rescal}, DistMult~\cite{dist}, HolE~\cite{hole}, ComplEx~\cite{complex}, ANALOGY~\cite{analogy}, SimplE~\cite{simple}, and QuatE~\cite{quate}. To show our SM decoder can be specialized to express these existing methods, we introduce two learnable matrices, $\mT_{\text{ent}}^{\langle j \rangle}\in \RR^{d_L\times \bar{d}}$ and $\mT_{\text{rel}}^{\langle j \rangle}\in\RR^{d_L'\times\bar{d}'}$. Note that we need to only define $\mbU_{r}^{\langle j \rangle}$ in Definition~{\ref{def:sm}} to simulate the existing methods. We omit DistMult and ANALOGY here since they are described in the main paper.

%Now, we describe how our semantic matching decoder can generalize the existing knowledge graph embedding methods mentioned in the main paper. Our semantic matching decoder linearly transforms the head entity based on the relation, and then calculates the dot product similarity between the projected head entity and the tail entity. Such simple philosophy is shared with many existing methods, such as 
%Similar to the instances of our translational distance decoder, these existing knowledge graph embedding methods require extra embedding matrices, $\mT_{\text{ent}}\in \RR^{d_L\times \bar{d}}$ and $\mT_{\text{rel}}\in\RR^{d_L'\times\bar{d}'}$ for complete formulation. Note that these embedding matrices are used to decompose the relation-specific projection matrices.

\paragraph*{RESCAL~\cite{rescal}}
By defining a relation-specific projection matrix $\mB_r^{\langle j \rangle}\in \RR^{\bar{d}\times\bar{d}}$, we can easily represent RESCAL by setting
\[\mbU_{r}^{\langle j \rangle}\coloneqq \mT_{\text{ent}}^{\langle j \rangle}\mB_{r}^{\langle j \rangle}{\mT_{\text{ent}}^{\langle j \rangle}}^\top.\]

%\paragraph*{DistMult~\cite{dist}}
%DistMult is similar to RESCAL, but it has one constraint regarding the relation-specific projection matrix: diagonality. DistMult uses relation-specific diagonal matrices with the representation of each relation as its diagonal entries. Therefore, we use $\diag{\mR^{(L)}\mT_{\text{rel}}^{\langle j \rangle}[r,:]}$ to formulate the relation-specific diagonal matrix of DistMult, which leads to
%\[\mbU_{r}^{\langle j \rangle}\coloneqq \mT_{\text{ent}}^{\langle j \rangle} \left(\diag{\mR^{(L)}\mT_{\text{rel}}^{\langle j \rangle}[r,:]}\right){\mT_{\text{ent}}^{\langle j \rangle}}^\top\]
%with $\bar{d}=\bar{d'}$.

\paragraph*{HolE~\cite{hole} and ComplEx~\cite{complex}}
It has been known that HolE and ComplEx are special cases of ANALOGY~\cite{analogy}. Since our SM decoder can represent ANALOGY as described in Section~{\ref{sec:tcd}}, our decoder can also represent HolE and ComplEx.

\paragraph*{SimplE~\cite{simple}}
In SimplE, an entity representation is divided into two parts: the first part is the representation when the entity appears as a head entity, and the second part is the representation when the entity appears as a tail entity. For each relation $r$, its inverse relation $r^{-1}$ is also considered. As a result, a relation representation is also divided into two parts: representations for $r$ and $r^{-1}$. Given a triplet $(h,r,t)$, SimplE calculates its score by averaging the scores of $(h,r,t)$ and $(t,r^{-1},h)$. Given $\bar{d}=\bar{d'}$, we can formulate SimplE by setting

\[\mbU_{r}^{\langle j \rangle}\coloneqq \frac{1}{2}\mT_{\text{ent}}^{\langle j \rangle}\left(\diag{\big(\mR^{(L)}\mT_{\text{rel}}^{\langle j \rangle}\big)[r,:]}
\left[\begin{matrix}
  \mzero_{\bar{d'}/2,\bar{d'}/2}&\mI_{\bar{d'}/2,\bar{d'}/2}\\
  \mI_{\bar{d'}/2,\bar{d'}/2}&\mzero_{\bar{d'}/2,\bar{d'}/2}
\end{matrix}\right]\right){\mT_{\text{ent}}^{\langle j \rangle}}^\top.\]

\paragraph*{QuatE~\cite{quate}}
In QuatE, each representation vector is represented in the quaternion space. Let us denote a representation vector of $r$ for $\langle j \rangle$ in the quaternion space as a quaternion of real numbers such that $ \left[\begin{matrix}\mathfrak{a}_r^{\langle j \rangle} & \mathfrak{b}_r^{\langle j \rangle} & \mathfrak{c}_r^{\langle j \rangle} & \mathfrak{d}_r^{\langle j \rangle}\end{matrix}\right]$ where $\mathfrak{a}_r^{\langle j \rangle}$ is the real part and $\mathfrak{b}_r^{\langle j \rangle}, \mathfrak{c}_r^{\langle j \rangle}$, and $\mathfrak{d}_r^{\langle j \rangle}$ correspond to the imaginary coefficients. It is assumed that $(\mathfrak{a}_r^{\langle j \rangle}[i])^2+(\mathfrak{b}_r^{\langle j \rangle}[i])^2+(\mathfrak{c}_r^{\langle j \rangle}[i])^2+(\mathfrak{d}_r^{\langle j \rangle}[i])^2=1$ for $i\in\{0,1,\cdots,\text{$\frac{\bar{d'}}{4}$$-$$1$}\}$ and $j\in\{0,1\}$. Let us define
\[\mat{\mathfrak{A}}_r^{\langle j \rangle}\coloneqq\diag{\mathfrak{a}_r^{\langle j \rangle}}\in\RR^{\bar{d'}/4\times\bar{d'}/4}, \mat{\mathfrak{B}}_r^{\langle j \rangle}\coloneqq\diag{\mathfrak{b}_r^{\langle j \rangle}}\in\RR^{\bar{d'}/4\times\bar{d'}/4}\]
\[\mat{\mathfrak{C}}_r^{\langle j \rangle}\coloneqq\diag{\mathfrak{c}_r^{\langle j \rangle}}\in\RR^{\bar{d'}/4\times\bar{d'}/4}, \mat{\mathfrak{D}}_r^{\langle j \rangle}\coloneqq\diag{\mathfrak{d}_r^{\langle j \rangle}}\in\RR^{\bar{d'}/4\times\bar{d'}/4}.\]
Given $\bar{d}=\bar{d'}$, we can express QuatE using our SM decoder by setting
\begin{align*}
  &\mbU_r^{\langle j \rangle} \coloneqq \mT_{\text{ent}}^{\langle j \rangle}\left[\begin{matrix}
    \mat{\mathfrak{A}}_r^{\langle j \rangle} & \mat{\mathfrak{B}}_r^{\langle j \rangle} & \mat{\mathfrak{C}}_r^{\langle j \rangle} & \mat{\mathfrak{D}}_r^{\langle j \rangle}\\
    -\mat{\mathfrak{B}}_r^{\langle j \rangle} & \mat{\mathfrak{A}}_r^{\langle j \rangle}  & -\mat{\mathfrak{D}}_r^{\langle j \rangle} & \mat{\mathfrak{C}}_r^{\langle j \rangle}\\
    -\mat{\mathfrak{C}}_r^{\langle j \rangle} & \mat{\mathfrak{D}}_r^{\langle j \rangle} & \mat{\mathfrak{A}}_r^{\langle j \rangle} & -\mat{\mathfrak{B}}_r^{\langle j \rangle}\\
    -\mat{\mathfrak{D}}_r^{\langle j \rangle} & -\mat{\mathfrak{C}}_r^{\langle j \rangle} & \mat{\mathfrak{B}}_r^{\langle j \rangle} & \mat{\mathfrak{A}}_r^{\langle j \rangle}
  \end{matrix}\right]{\mT_{\text{ent}}^{\langle j \rangle}}^\top.
\end{align*}

\section{Proof of Theorem~\ref{thm:transbound}}
\label{appn:transbound}
In Section~{\ref{sec:transbound}}, we present our Theorem~\ref{thm:transbound} which states the transductive PAC-Bayesian generalization bound for a deterministic triplet classifier. We derive Theorem~\ref{thm:transbound} from the following Lemma~\ref{lem:trans} which is originally presented as Corollary~7 in \citeauthor{pactrans}~\cite{pactrans} where a transductive PAC-Bayesian generalization bound is presented for a stochastic model. We paraphrase the original version to customize it to our problem setting. 

\begin{lemma}[\citeauthor{pactrans}~\cite{pactrans}, Corollary 7]\label{lem:trans} 
For any full triplet set $\sE$ having size $\nE \geq 40$, for any stochastic triplet classifier $\tf$ following a posterior distribution $\sQ$ on a hypothesis space $\sH$, for any prior distribution $\sP$ on $\sH$, for any $\gamma,\delta>0$, with probability at least $1-\delta$, over the choice of a training triplet set $\shE$ (such that $20\leq \nhE \leq \nE-20$) drawn without replacement from the full triplet set $\sE$, we have
  \[\sL_{\gamma,\sE}\left(\tf\right) \leq \sL_{\gamma,\shE}\left(\tf\right)+\sqrt{\frac{1-\frac{\nhE}{\nE}}{2\nhE}\left[D_{KL}(\sQ\|\sP)+\ln\frac{\theta(\nhE,\nE)}{\delta}\right]}\]
where $\sL_{\gamma,\sZ}\left(\tf\right)\coloneqq {\EE_{\vtw\sim\sQ}}\big[\sL_{\gamma,\sZ}(f_{\vtw})\big]$, $f_{\vtw}$ is a deterministic triplet classifier with parameters $\vtw$, \(\sL_{\gamma,\sZ}(f_{\vtw})\) is defined in Definition~{\ref{def:margin}}, \(D_{KL}(\sQ\|\sP)\) is the KL-divergence of \(\sQ\) from \(\sP\), and \(\theta(\nhE,\nE) = 3\sqrt{\nhE(1-\frac{\nhE}{\nE})}\ln{\nhE}\).
\end{lemma}

Note that the prior distribution $\sP$ is independent of the training triplets. While Lemma~\ref{lem:trans} considers a stochastic model, we need to consider a deterministic model because the ReED framework results in a deterministic triplet classifier. Given a deterministic triplet classifier $f_{\vw}$ with the fixed parameters $\vw$, we add random perturbations $\vdw$ to $\vw$ to simulate a stochastic triplet classifier so that Lemma~\ref{lem:trans} can be extended to Theorem~\ref{thm:transbound}.

%Assume that the prior distribution $\sP$ is independent of the training triplets.

\begin{thm:transbound}[Transductive PAC-Bayesian Generalization Bound for a Deterministic Triplet Classifier]
Let \(f_{\vw}:\sV\times\sR\times\sV\rightarrow\RR^2\) be a deterministic triplet classifier with parameters $\vw$, and \(\sP\) be any prior distribution on $\vw$. Let us consider the finite full triplet set $\sE\subseteq\sV\times\sR\times\sV$. We construct a posterior distribution \(\sQ_{\vw+\ddot{\vw}}\) by adding any random perturbation \(\ddot{\vw}\) to \(\vw\) such that $\prob(\max_{(h,r,t)\in\sE}\linf{f_{\vw+\ddot{\vw}}(h,r,t)-f_{\vw}(h,r,t)} < \frac{\gamma}{4})>\frac{1}{2}$. Then, for any \(\gamma,\delta > 0\), with probability \(1-\delta\) over the choice of a training triplet set \(\shE\) drawn from the full triplet set \(\sE\) (such that $20\leq \nhE \leq \nE-20$ and $\nE \geq 40$) without replacement, for any \(\vw\), we have 
\begin{align}
\sL_{0,\sE}(f_{\vw}) \leq \sL_{\gamma,\shE}(f_{\vw}) +\sqrt{\frac{1-\frac{\nhE}{\nE}}{2\nhE}\left[2D_{KL}(\sQ_{\vw+\ddot{\vw}}\|\sP)+\ln\frac{4\theta(\nhE,\nE)}{\delta}\right]}\notag
\end{align}
where $\sL_{\gamma,\shE}(f_{\vw})$ is defined in Definition~\ref{def:margin}, $\sL_{0,\sE}(f_{\vw})$ is defined in Definition~\ref{def:cls}, \(D_{KL}(\sQ_{\vw+\ddot{\vw}}\|\sP)\) is the KL-divergence of $\sQ_{\vw+\ddot{\vw}}$ from $\sP$, and \(\theta(\nhE,\nE) = 3\sqrt{\nhE(1-\frac{\nhE}{\nE})}\ln{\nhE}\).
\end{thm:transbound}

\begin{proof}
Let $\sH$ denote the hypothesis space of a triplet classifier $f_{\vw}$. Note that the posterior distribution $\sQ_{\vw+\vdw}$ is a distribution over $\sH$. Let $\sQ_{\vw+\vdw}\left(\vtw\right)$ be the probability density function indicating the probability of $\vtw$ being drawn from $\sQ_{\vw+\vdw}$. Also, let $\sC$ be a set of perturbed parameters $\vtw$ such that \(\sC \coloneqq \left\{\vtw\in\sH\Big| \max_{(h,r,t) \in \sE}\linf{f_{\vtw}(h,r,t)-f_{\vw}(h,r,t)}<\frac{\gamma}{4}\right\} \subset \sH\). If we define $p\coloneqq\prob_{\vtw\sim\sQ_{\vw+\vdw}}(\vtw\in\sC)$, then $p > \frac{1}{2}$ by our assumption.

%Note that for any $\vtw \in \sC$, $\max_{(h,r,t) \in \sE}\linf{f_{\vtw}(h,r,t)-f_{\vw}(h,r,t)}<\frac{\gamma}{4}$ holds.

We divide $\sQ_{\vw+\vdw}$ into two distributions $\sgQ$ and $\saQ$ where the former has a non-zero value for $\vtw \in \sC$ and the latter has a non-zero value for $\vtw \in \sH\setminus\sC$. Specifically, $\sgQ\left(\vtw\right)$ and $\saQ\left(\vtw\right)$ are defined as \[\sgQ(\vtw)=\begin{cases}\displaystyle \frac{1}{p}\sQ_{\vw+\vdw}(\vtw) & \vtw \in \sC \\ 0 & \vtw \in \sH\setminus\sC\end{cases},\quad\saQ(\vtw)=\begin{cases} 0 & \vtw \in \sC \\\displaystyle \frac{1}{1-p}\sQ_{\vw+\vdw}(\vtw) & \vtw \in \sH\setminus\sC\end{cases}\]
  
  For any $(h,r,t)\in \sE$ and $\vtw\sim\sgQ$, we have
   \begin{small}
  \begin{align*}
    &\size{(f_{\vtw}(h,r,t)[y_{hrt}]-f_{\vtw}(h,r,t)[1-y_{hrt}])-(f_{\vw}(h,r,t)[y_{hrt}]-f_{\vw}(h,r,t)[1-y_{hrt}])}\\
    =&\size{(f_{\vtw}(h,r,t)[y_{hrt}]-f_{\vw}(h,r,t)[y_{hrt}])-(f_{\vtw}(h,r,t)[1-y_{hrt}]-f_{\vw}(h,r,t)[1-y_{hrt}])}\\
    \leq&\size{f_{\vtw}(h,r,t)[y_{hrt}]-f_{\vw}(h,r,t)[y_{hrt}]}+\size{f_{\vtw}(h,r,t)[1-y_{hrt}]-f_{\vw}(h,r,t)[1-y_{hrt}]}\\
    \leq&\max_{(h,r,t)\in\sE}(\size{f_{\vtw}(h,r,t)[y_{hrt}]-f_{\vw}(h,r,t)[y_{hrt}]}+\size{f_{\vtw}(h,r,t)[1-y_{hrt}]-f_{\vw}(h,r,t)[1-y_{hrt}]})\\
    <& \frac{\gamma}{4}+\frac{\gamma}{4} =\frac{\gamma}{2} \quad (\text{sub-additivity}, \vtw\in \sC)\\
  \end{align*}
  \end{small}
  Then, we have
\begin{small}
\begin{align*}  
  &f_{\vw}(h,r,t)[y_{hrt}]-f_{\vw}(h,r,t)[1-y_{hrt}]\leq 0 \Rightarrow f_{\vtw}(h,r,t)[y_{hrt}]-f_{\vtw}(h,r,t)[1-y_{hrt}]\leq \frac{\gamma}{2}\\
  &f_{\vw}(h,r,t)[y_{hrt}]\leq f_{\vw}(h,r,t)[1-y_{hrt}]\Rightarrow f_{\vtw}(h,r,t)[y_{hrt}]\leq \frac{\gamma}{2} + f_{\vtw}(h,r,t)[1-y_{hrt}]
  \end{align*}
  \end{small}
  which indicates that
  \[\bone\big[f_{\vw}(h,r,t)[y_{hrt}]\leq f_{\vw}(h,r,t)[1-y_{hrt}]\big]
  \leq \bone\big[f_{\vtw}(h,r,t)[y_{hrt}]\leq\frac{\gamma}{2}+f_{\vtw}(h,r,t)[1-y_{hrt}]\big]\]
 
  This leads to $\sL_{0,\sE}(f_{\vw}) \leq \sL_{\frac{\gamma}{2},\sE}(f_{\vtw})$ for any $\vtw\sim\sgQ$, meaning that 
  \begin{equation}\label{eq:1}
    \sL_{0,\sE}(f_{\vw}) \leq \EE_{\vtw\sim\sgQ}\big[\sL_{\frac{\gamma}{2},\sE}(f_{\vtw})\big]
  \end{equation}
  
  Also, for any $(h,r,t) \in \sE$ and $\vtw\sim\sgQ$, we have
  \begin{small}
  \begin{align*}
  &f_{\vtw}(h,r,t)[y_{hrt}]-f_{\vtw}(h,r,t)[1-y_{hrt}]\leq \frac{\gamma}{2} \Rightarrow f_{\vw}(h,r,t)[y_{hrt}]-f_{\vw}(h,r,t)[1-y_{hrt}]\leq \gamma\\
  & f_{\vtw}(h,r,t)[y_{hrt}]\leq \frac{\gamma}{2} + f_{\vtw}(h,r,t)[1-y_{hrt}] \Rightarrow f_{\vw}(h,r,t)[y_{hrt}]\leq \gamma + f_{\vw}(h,r,t)[1-y_{hrt}]\\
  \end{align*}
  \end{small}
  which indicates that
  \begin{small}
  \[\bone\big[f_{\vtw}(h,r,t)[y_{hrt}]\leq \frac{\gamma}{2}+f_{\vtw}(h,r,t)[1-y_{hrt}]\big]\\
  \leq \bone\big[f_{\vw}(h,r,t)[	y_{hrt}] \leq \gamma+f_{\vw}(h,r,t)[1-y_{hrt}]\big]\]
  \end{small}

  This leads to $\sL_{\frac{\gamma}{2}, \shE}(f_{\vtw}) \leq \sL_{\gamma,\shE}(f_{\vw})$ for any $\vtw\sim\sgQ$.
  Then, we can end up with
  \begin{equation}\label{eq:2}
    \EE_{\vtw\sim\sgQ}\big[\sL_{\frac{\gamma}{2}, \shE}(f_{\vtw})\big] \leq \sL_{\gamma,\shE}(f_{\vw})
  \end{equation}
  
 % We assume that the number of entities, relations and the size of the train set of knowledge graph are sufficiently large ($20\leq m \leq N-20$).
 Given $20\leq \nhE \leq \nE-20$ and $\nE \geq 40$, with probability $1-\delta$, we have
  \begin{align*}
    \sL_{0,\sE}(f_{\vw}) \leq& \EE_{\vtw\sim\sgQ}\big[\sL_{\frac{\gamma}{2},\sE}(f_{\vtw})\big]\quad(\text{Eq.~\eqref{eq:1}})\\
    \leq& \EE_{\vtw\sim\sgQ}\big[\sL_{\frac{\gamma}{2},\shE}(f_{\vtw})\big]+\sqrt{\frac{1-\frac{\nhE}{\nE}}{2\nhE}\left[D_{KL}(\sgQ\|\sP)+\ln\frac{\theta(\nhE,\nE)}{\delta}\right]}\quad(\text{Lemma~\ref{lem:trans}})\\
    \leq& \sL_{\gamma,\shE}(f_{\vw})+\sqrt{\frac{1-\frac{\nhE}{\nE}}{2\nhE}\left[D_{KL}(\sgQ\|\sP)+\ln\frac{\theta(\nhE,\nE)}{\delta}\right]}\quad(\text{Eq.~\eqref{eq:2}})\\
  \end{align*}
  Also, we can derive the following.
  \begin{align*}
    D_{KL}(\sQ_{\vw+\vdw}\|\sP)=&\int_{\vtw\in\sC}\sQ_{\vw+\vdw}\ln\frac{\sQ_{\vw+\vdw}}{\sP}d\vtw + \int_{\vtw\in\sH\setminus\sC}\sQ_{\vw+\vdw}\ln\frac{\sQ_{\vw+\vdw}}{\sP}d\vtw\\
    =&p\int_{\vtw\in\sC}\frac{\sQ_{\vw+\vdw}}{p}\ln\frac{\sQ_{\vw+\vdw}}{p\sP}d\vtw+ (1-p)\int_{\vtw\in\sH\setminus\sC}\frac{\sQ_{\vw+\vdw}}{1-p}\ln\frac{\sQ_{\vw+\vdw}}{(1-p)\sP}d\vtw\\
    &+\int_{\vtw\in\sC}\sQ_{\vw+\vdw}\ln p\>d\vtw +\int_{\vtw\in\sH\setminus\sC}\sQ_{\vw+\vdw}\ln (1-p)\>d\vtw\\
    =&pD_{KL}(\sgQ\|\sP)+(1-p)D_{KL}(\saQ\|\sP)+p\ln p + (1-p)\ln (1-p)\\
  \end{align*}
  Since we know \(\frac{1}{2}<p<1\) from the assumption, we have \((-\ln 2)< p\ln p  + (1-p)\ln (1-p)<0\). Also, \(D_{KL}\) is non-negative. Therefore, we have
  \begin{align*}
    D_{KL}(\sgQ\|\sP) =& \frac{1}{p}(D_{KL}(\sQ_{\vw+\vdw}\|\sP)-(1-p)D_{KL}(\saQ\|\sP)-p\ln p - (1-p)\ln (1-p))\\
    \leq& \frac{1}{p}(D_{KL}(\sQ_{\vw+\vdw}\|\sP)+\ln 2) \leq 2D_{KL}(\sQ_{\vw+\vdw}\|\sP)+2\ln 2\\
  \end{align*}
  Finally, we show
  \begin{align*}
    \sL_{0,\sE}(f_{\vw}) \leq& \sL_{\gamma,\shE}(f_{\vw})+\sqrt{\frac{1-\frac{\nhE}{\nE}}{2\nhE}\left[D_{KL}(\sgQ\|\sP)+\ln\frac{\theta(\nhE,\nE)}{\delta}\right]}\\
    \leq&\sL_{\gamma,\shE}(f_{\vw})+\sqrt{\frac{1-\frac{\nhE}{\nE}}{2\nhE}\left[2D_{KL}(\sQ_{\vw+\vdw}\|\sP)+\ln\frac{4\theta(\nhE,\nE)}{\delta}\right]}
  \end{align*}
\end{proof}

\section{Proofs of Theorem~\ref{thm:boundtd} and Theorem~\ref{thm:boundsm}}
\label{appn:mainbound}
We provide the complete proofs of Theorem~\ref{thm:boundtd} and Theorem~\ref{thm:boundsm}.

\begin{thm:boundtd}[Generalization Bound for ReED with Translational Distance Decoder]
  For any \(L\geq 0\), let \(f_{\vw} : \sV\times\sR\times\sV\rightarrow\RR^2\) be a triplet classifier designed by the combination of the RAMP encoder with $L$-layers in Definition~\ref{def:ramp} and the TD decoder in Definition~\ref{def:td}. Let $k_r$ be the maximum of the infinity norms for all possible $\mS_r^{(l)}$ in the RAMP encoder. Then, for any \(\delta, \gamma>0\), with probability at least $1-\delta$ over a training triplet set \(\shE\) (such that $20\leq \nhE \leq \nE -20$) sampled without replacement from the full triplet set \(\sE\), for any \(\vw\), we have
  \begin{align*}
    \sL_{0,\sE}(f_{\vw})\leq \sL_{\gamma,\shE}(f_{\vw})
    +\sO\!\left(\!\!\!\sqrt{\frac{1-\frac{\nhE}{\nE}}{\nhE}\left[\frac{N_{\vw} L^2\zeta_L^2 s^{2L}d\ln{(N_{\vw} d)}}{\gamma^2}+\ln\frac{\theta(\nhE,\nE)}{\delta}\right]}\right)\notag
  \end{align*}
  where \(\theta(\nhE,\nE) = 3\sqrt{\nhE(1-\frac{\nhE}{\nE})}\ln{\nhE}, N_{\vw} =2\nr L+\)
  $6\nr+2L$, $\zeta_L = 2\tau^L\l2{\xv}+2\kappa\l2{\xr}\left(\sum_{i=0}^{L-1}\tau^i\right) + $
  $\l2{\xr}$, $\tau= C_\phi+\kappa, \kappa=C_\phi C_\rho C_\psi\sum_{r\in\sR}k_r, d=$
  $\max\left(\max_{0\leq l\leq L+1}(d_l),\max_{0\leq l\leq L+1}(d_l')\right), s_{L+1}=$
  $\max_{r,j}(\max(\lfr{\mbW_{r}^{\langle j \rangle}},\lfr{\mbU_{r}^{\langle j \rangle}},\lfr{\mV_{r}^{\langle j \rangle}}))$, $s_l =$
  $\max(\lfr{\mW_{0}^{(l)}}, \lfr{\mU_{0}^{(l)}},\max_r\lfr{\mW_{r}^{(l)}},\max_r\lfr{\mU_{r}^{(l)}})$ for $l \in \{1,2,\ldots,L\}$, and $s=\max_{1\leq l\leq L+1}(s_l)$.
\end{thm:boundtd}

\begin{proof}
We derive Theorem~\ref{thm:boundtd} from Theorem~\ref{thm:transbound} where we construct a posterior distribution $\sQ_{\vw+\vdw}$ by adding random perturbations $\vdw$ to $\vw$. Following~\cite{gcpac, nnpac}, we set the prior distribution $\sP$ as $\sN\left(\vzero_{n_\vw}, \sigma^2\mI_{n_\vw \times n_\vw}\right)$ and the posterior distribution $\sQ_{\vw+\vdw}$ as $ \sN\left(\vw, \sigma^2\mI_{n_\vw\times n_\vw}\right)$ where $n_\vw$ is the size of $\vw$. We first compute $\max_{(h,r,t)\in\sE}\linf{f_{\vw+\ddot{\vw}}(h,r,t)-f_{\vw}(h,r,t)}$, which we call the perturbation bound, so that we can calculate the standard deviation $\sigma$ of the prior distribution that satisfies $\prob\left(\max_{(h,r,t)\in\sE}\linf{f_{\vw+\vdw}(h,r,t)-f_{\vw}(h,r,t)} < \frac{\gamma}{4}\right)>\frac{1}{2}$. Afterwards, we calculate the KL divergence of $\sQ_{\vw+\vdw}$ from $\sP$ using $\sigma$ and substitute the KL divergence term in Theorem~\ref{thm:transbound} with our data and model-related terms, which finishes the proof.
  
  \item \paragraph{Perturbation bound of ReED with translational distance decoder}
  First, we compute the perturbation bound, $\max_{(h,r,t)\in\sE}\linf{f_{\vw+\ddot{\vw}}(h,r,t)-f_{\vw}(h,r,t)}$, and find $\sigma$ that makes $\prob(\max_{(h,r,t)\in\sE}\linf{f_{\vw+\vdw}(h,r,t)$ $-f_{\vw}(h,r,t)} < \frac{\gamma}{4})>\frac{1}{2}$ true. Let $\mdW$ denote a perturbation (also called noise) matrix added to the original weight matrix $\mW$. As a result, we have $\mtW = \mW + \mdW$ where $\mtW$ is a perturbed weight matrix. Also, let $\mtH^{(l)}$ and $\mtR^{(l)}$ denote the outputs of the perturbed model at the $l$-th layer. Each element of $\mdW$ is an i.i.d. element drawn from $\sN\left(0,\sigma^2\right)$. Assume that the maximum of the Frobenius norms of the noise matrices is $\ds$. That is,
  \begin{align}
    \ds = \max\bigg(&\max_l\bigg(\bblfr{\mdW_{0}^{(l)}}, \bblfr{\mdU_{0}^{(l)}},\max_r\bblfr{\mdW_{r}^{(l)}},\max_r\bblfr{\mdU_{r}^{(l)}}\bigg),
    &\max_{r,j}\left(\bblfr{\mbdW_{r}^{\langle j \rangle}},\bblfr{\mbdU_{r}^{\langle j \rangle}},\bblfr{\mdV_{r}^{\langle j \rangle}}\right)\bigg)\notag
  \end{align}
  Now, let us calculate the perturbation bound:
  \begin{align}
  \max_{(h,r,t)\in\sE}\linf{f_{\vw+\vdw}(h,r,t)-f_{\vw}(h,r,t)}
  = \max_{(h,r,t)\in\sE}\max_{j\in \{0,1\}}\bbbsize{&-\l2{\mtH^{(L)}[h,:]\mbtW_{r}^{\langle j \rangle}+\mtR^{(L)}[r,:]\mbtU_{r}^{\langle j \rangle}-\mtH^{(L)}[t,:]\mtV_{r}^{\langle j \rangle}}\notag\\
  &+\l2{\mH^{(L)}[h,:]\mbW_{r}^{\langle j \rangle}+\mR^{(L)}[r,:]\mbU_{r}^{\langle j \rangle}-\mH^{(L)}[t,:]\mV_{r}^{\langle j \rangle}}}\notag
  \end{align}
Let us define $\Phi_{l} = \max_v\l2{\mH^{(l)}[v,:]}$, $\Psi_{l} = \max_v\l2{\mtH^{(l)}[v,:]-\mH^{(l)}[v,:]}$, $\Lambda_{l}=\max_r \l2{\mR^{(l)}[r,:]}$, and $\Gamma_{l}=\max_r \l2{\mtR^{(l)}[r,:]-\mR^{(l)}[r,:]}$. Then, for any $(h,r,t)\in \sE$,
  {\small
  \begin{align}
    &\bbbsize{\bbbl2{\mtH^{(L)}[h,:]\mbtW_{r}^{\langle j \rangle}+\mtR^{(L)}[r,:]\mbtU_{r}^{\langle j \rangle}-\mtH^{(L)}[t,:]\mtV_{r}^{\langle j \rangle}}-\bbbl2{\mH^{(L)}[h,:]\mbW_{r}^{\langle j \rangle}+\mR^{(L)}[r,:]\mbU_{r}^{\langle j \rangle}-\mH^{(L)}[t,:]\mV_{r}^{\langle j \rangle}}} \notag\\
    \leq& \bbbl2{\mtH^{(L)}[h,:]\mbtW_{r}^{\langle j \rangle}+\mtR^{(L)}[r,:]\mbtU_{r}^{\langle j \rangle}-\mtH^{(L)}[t,:]\mtV_{r}^{\langle j \rangle} \quad(\text{reverse triangle inequality}) \notag\\
    &\hspace{0.5em}-\left(\mH^{(L)}[h,:]\mbW_{r}^{\langle j \rangle}+\mR^{(L)}[r,:]\mbU_{r}^{\langle j \rangle}-\mH^{(L)}[t,:]\mV_{r}^{\langle j \rangle}\right)} \notag\\
    =& \bbbl2{\left(\mtH^{(L)}-\mH^{(L)}\right)[h,:]\mbtW_{r}^{\langle j \rangle}+\left(\mtR^{(L)}-\mR^{(L)}\right)[r,:]\mbtU_{r}^{\langle j \rangle} \notag\\
    &\hspace{0.5em}-\left(\mtH^{(L)}-\mH^{(L)}\right)[t,:]\mtV_{r}^{\langle j \rangle}+\mH^{(L)}[h,:]\mbdW_{r}^{\langle 1 \rangle}+\mR^{(L)}[r,:]\mbdU_{r}^{\langle j \rangle} \notag -\mH^{(L)}[t,:]\mdV_{r}^{\langle j \rangle}} \notag\\
    \leq& \l2{\left(\mtH^{(L)}-\mH^{(L)}\right)[h,:]}\l2{\mbtW_{r}^{\langle j \rangle}}+\l2{\left(\mtR^{(L)}-\mR^{(L)}\right)[r,:]}\l2{\mbtU_{r}^{\langle j \rangle}} \notag\\
    &\hspace{0.5em}+\l2{\left(\mtH^{(L)}-\mH^{(L)}\right)[t,:]}\l2{\mtV_{r}^{\langle j \rangle}}+\l2{\mH^{(L)}[h,:]}\l2{\mbdW_{r}^{\langle j \rangle}}+\l2{\mR^{(L)}[r,:]}\l2{\mbdU_{r}^{\langle j \rangle}} \notag\\
    &\hspace{0.5em}+\l2{\mH^{(L)}[t,:]}\l2{\mdV_{r}^{\langle j \rangle}}\quad(\text{sub-additivity, sub-multiplicativity}) \notag\\
    \leq& (2\Psi_{L}+\Gamma_{L})(s_{L+1}+\ds ) + (2\Phi_{L}+\Lambda_{L})\ds\quad(\text{definitions of } \Psi_{L}, \Gamma_{L}, s_{L+1}, \Phi_L, \Lambda_{L},\text{and }\ds)\label{tdmax}
  \end{align}}
  
  To satisfy the condition of Theorem~\ref{thm:transbound}, we need the bound of $\Phi_{L}$, $\Psi_{L}$, $\Lambda_{L}$, and $\Gamma_{L}$. Since we need $\Lambda_L$ and $\Gamma_L$ to calculate $\Phi_L$ and $\Psi_L$, we first calculate the bound of $\Lambda_L$ and $\Gamma_L$. In general, we need to compute \raisebox{.9pt}{\textcircled{\raisebox{-.9pt} {\small 1}}} bound of $\Lambda_{l}$ and $\Gamma_{l}$, \raisebox{0.9pt}{\textcircled{\raisebox{-.9pt} {\small 2}}} bound of $\Phi_{l}$, and \raisebox{.9pt}{\textcircled{\raisebox{-.9pt} {\small 3}}} bound of $\Psi_{l}$. 
  \item \paragraph{\raisebox{.9pt}{\textcircled{\raisebox{-.9pt} {\small 1}}} Bound of $\Lambda_{l}$ and $\Gamma_{l}$}\mbox{}\\
  First, we calculate the upper bound of $\l2{\mR^{(l)}[r,:]}$ and 
  $\l2{\mtR^{(l)}[r,:]-\mR^{(l)}[r,:]}$.
  
  \begin{align}
    \l2{\mR^{(l)}[r,:]} =& \l2{\mR^{(0)}[r,:]\prod_{i=1}^l\mU_{0}^{(i)}} \leq \l2{\xr}\prod_{i=1}^{l}s_i\label{eq:rbound}\\
    \l2{\mtR^{(l)}[r,:]-\mR^{(l)}[r,:]}=&\l2{\mtR^{(l-1)}[r,:]\mtU_{0}^{(l)}-\mR^{(l-1)}[r,:]\mU_{0}^{(l)}}\notag\\
    =&\l2{\mtR^{(l-1)}[r,:]\mtU_{0}^{(l)}-\mR^{(l-1)}[r,:]\mtU_{0}^{(l)}+\mR^{(l-1)}[r,:]\mdU_{0}^{(l)}}\notag\\
    \leq&\l2{\mtR^{(l-1)}[r,:]-\mR^{(l-1)}[r,:]}\l2{\mtU_{0}^{(l)}}\notag\\
    &+\l2{\mR^{(l-1)}[r,:]}\l2{\mdU_{0}^{(l)}}\quad (\text{sub-additivity, sub-multiplicativity})\notag\\
    \leq&\l2{\mtR^{(l-1)}[r,:]-\mR^{(l-1)}[r,:]}(s_l+\ds )
    +\l2{\mR^{(l-1)}[r,:]}\ds\quad (\text{definitions of }s_l\text{ and }\ds)\notag\\
    \leq&\l2{\mtR^{(l-1)}[r,:]-\mR^{(l-1)}[r,:]}(s_l+\ds )+\ds \l2{\xr}\prod_{i=1}^{l-1}s_i \hspace{0.5em}(\text{Eq.\eqref{eq:rbound}})\notag
  \end{align}
 Note that 
 \begin{align}
 \l2{\mtR^{(1)}[r,:]-\mR^{(1)}[r,:]}=&\l2{\xr[r,:]\mtU_{0}^{(1)}-\xr[r,:]\mU_{0}^{(1)}}\notag\\
 =&\l2{\xr[r,:]\mdU_{0}^{(1)}}\notag\\
 \leq&\l2{\xr[r,:]}\l2{\mdU_{0}^{(1)}}\notag\\
 \leq&\l2{\xr}\ds\quad(\text{definitions of the spectral norm and }\ds)\label{eq:1rdif}
 \end{align}
 
 Then, we get
 \begin{small}
  \begin{align}
  \l2{\mtR^{(l)}[r,:]-\mR^{(l)}[r,:]}+\l2{\xr}\prod_{i=1}^{l}s_i\leq&\left(\l2{\mtR^{(l-1)}[r,:]-\mR^{(l-1)}[r,:]}+\l2{\xr}\prod_{i=1}^{l-1}s_i\right)(s_l+\ds )\notag\\
  \leq&\Big(\l2{\mtR^{(1)}[r,:]-\mR^{(1)}[r,:]}+\l2{\xr}s_1\Big)\prod_{i=2}^{l}(s_i+\ds )\quad (\text{recursion})\notag\\
  \leq&\left(\l2{\xr}\ds+\l2{\xr}s_1\right)\prod_{i=2}^{l}(s_i+\ds)\quad({\text{Eq. \eqref{eq:1rdif}}})\notag\\
    =& \l2{\xr}\prod_{i=1}^{l}(s_i+\ds )\notag
  \end{align}
  \end{small}
  Putting all this together, we have
  \begin{align}
    &\Lambda_{l}=\max_r \l2{\mR^{(l)}[r,:]}\leq \l2{\xr}\prod_{i=1}^{l}s_i\label{relPhi}\\
    &\Gamma_{l}=\max_r \l2{\mtR^{(l)}[r,:]-\mR^{(l)}[r,:]}\leq \l2{\xr}\left(\prod_{i=1}^{l}(s_i+\ds ) - \prod_{i=1}^{l}s_i\right)\label{relPsi}
  \end{align}
  Note that the bound of $\Lambda_{L}$ and $\Gamma_{L}$ can be obtained by setting $l=L$.  
  \item \paragraph{\raisebox{0.9pt}{\textcircled{\raisebox{-.9pt} {\small 2}}} Bound of $\Phi_{l}$}\mbox{}\\
  We can calculate the upper bound of $\Phi_{l}$ using sub-additivity, sub-multiplicativity, and the definitions of matrix norms. Note that the bound of $\Phi_{L}$ can be obtained by setting $l=L$.
  
  Let $v^*=\argmax_v\l2{\mH^{(l)}[v,:]}$. Then,
  \begin{footnotesize}
    \allowdisplaybreaks{
      \begin{align}
        \Phi_{l} =& \l2{\phi\left(\mH^{(l-1)}\mW_{0}^{(l)}+\rho\left(\sum_{r\in\sR}\mS_r^{(l)}\psi\left(\mM_{r}^{(l)}\right)\left[\begin{matrix}\mW_{r}^{(l)}\\\mU_{r}^{(l)}\end{matrix}\right]\right)\right)[v^*,:]}\notag\\
        \leq& C_{\phi}\l2{\left(\mH^{(l-1)}\mW_{0}^{(l)}\right)[v^*,:]+\rho\left(\sum_{r\in\sR}\mS_r^{(l)}\psi\left(\mM_{r}^{(l)}\right)\left[\begin{matrix}\mW_{r}^{(l)}\\\mU_{r}^{(l)}\end{matrix}\right]\right)[v^*,:]} \quad(\text{Lipschitzness of $\phi$, $\phi(\vzero)=\vzero$}) \notag\\
        \leq& C_{\phi}\l2{\mH^{(l-1)}[v^*,:]}\l2{\mW_{0}^{(l)}}+C_{\phi}\l2{\rho\left(\sum_{r\in\sR}\mS_r^{(l)}\psi\left(\mM_{r}^{(l)}\right)\left[\begin{matrix}\mW_{r}^{(l)}\\\mU_{r}^{(l)}\end{matrix}\right]\right)[v^*,:]}\quad\text{(sub-additivity, sub-multiplicativity)}\notag\\
        \leq& C_{\phi}\l2{\mH^{(l-1)}[v^*,:]}\l2{\mW_{0}^{(l)}}+C_{\phi}C_{\rho}\l2{\left(\sum_{r\in\sR}\mS_r^{(l)}\psi\left(\mM_{r}^{(l)}\right)\left[\begin{matrix}\mW_{r}^{(l)}\\\mU_{r}^{(l)}\end{matrix}\right]\right)[v^*,:]}\quad(\text{Lipschitzness of $\rho$, $\rho(\vzero)=\vzero$})\notag\\
        =& C_{\phi}\l2{\mH^{(l-1)}[v^*,:]}\l2{\mW_{0}^{(l)}}+C_{\phi}C_{\rho}\l2{\sum_{r\in\sR}\left(\sum_{v\in\sV}\mS_r^{(l)}[v^*,v]\left(\psi\left(\mM_{r}^{(l)}\right)[v,:]\right)\left[\begin{matrix}\mW_{r}^{(l)}\\\mU_{r}^{(l)}\end{matrix}\right]\right)}\notag\\
        =& C_{\phi}\l2{\mH^{(l-1)}[v^*,:]}\l2{\mW_{0}^{(l)}}\qquad(\text{definition of }\mM_{r}^{(l)})\notag\\
        &+C_{\phi}C_{\rho}\l2{\sum_{r\in\sR}\left(\sum_{v\in\sV}\mS_r^{(l)}[v^*,v]\left(\psi\left(\left[\begin{matrix}\mH^{(l-1)}[v,:]&\mR^{(l-1)}[r,:]\end{matrix}\right]\right)\right)\left[\begin{matrix}\mW_{r}^{(l)}\\\mU_{r}^{(l)}\end{matrix}\right]\right)}\notag\\
        \leq& C_{\phi}\l2{\mH^{(l-1)}[v^*,:]}\l2{\mW_{0}^{(l)}}\qquad(\text{sub-additivity, absolute homogeneity})\notag\\
        &+C_{\phi}C_{\rho}\sum_{r\in\sR}\sum_{v\in\sV}\bbsize{\mS_r^{(l)}[v^*,v]}\l2{\left(\psi\left(\mH^{(l-1)}[v,:]\right)\mW_{r}^{(l)}+\psi\left(\mR^{(l-1)}[r,:]\right)\mU_{r}^{(l)}\right)}\notag\\
        \leq& C_{\phi}\l2{\mH^{(l-1)}[v^*,:]}\l2{\mW_{0}^{(l)}}\qquad\text{(sub-additivity, sub-multiplicativity)}\notag\\
        &+C_{\phi}C_{\rho}\sum_{r\in\sR}\sum_{v\in\sV}\bbsize{\mS_r^{(l)}[v^*,v]}\left(\l2{\psi\left(\mH^{(l-1)}[v,:]\right)}\l2{\mW_{r}^{(l)}}+\l2{\psi\left(\mR^{(l-1)}[r,:]\right)}\l2{\mU_{r}^{(l)}}\right)\notag\\
        \leq& C_{\phi}\l2{\mH^{(l-1)}[v^*,:]}\l2{\mW_{0}^{(l)}}\quad(\text{Lipschitzness of $\psi$, $\psi(\vzero)=\vzero$})\notag\\
        &+C_{\phi}C_{\rho}C_{\psi}\sum_{r\in\sR}\sum_{v\in\sV}\bbsize{\mS_r^{(l)}[v^*,v]}\left(\l2{\mH^{(l-1)}[v,:]}\l2{\mW_{r}^{(l)}}+\l2{\mR^{(l-1)}[r,:]}\l2{\mU_{r}^{(l)}}\right)\notag\\
        \leq& C_{\phi}\Phi_{l-1}\l2{\mW_{0}^{(l)}}+C_{\phi}C_{\rho}C_{\psi}\sum_{r\in\sR}\left(\left(\Phi_{l-1}\l2{\mW_{r}^{(l)}}+\Lambda_{l-1}\l2{\mU_{r}^{(l)}}\right)\sum_{v\in\sV}\bbsize{\mS_r^{(l)}[v^*,v]}\right)\quad(\text{definitions of $\Phi_{l}$ and $\Lambda_{l}$})\notag\\
        \leq& C_{\phi}\Phi_{l-1}\l2{\mW_{0}^{(l)}}+C_{\phi}C_{\rho}C_{\psi}\sum_{r\in\sR}\left(\Phi_{l-1}\l2{\mW_{r}^{(l)}}+\Lambda_{l-1}\l2{\mU_{r}^{(l)}}\right)\linf{\mS_r^{(l)}}\quad(\text{definition of the infinity norm of a matrix})\notag\\
        \leq& C_{\phi}\Phi_{l-1}\l2{\mW_{0}^{(l)}}+C_{\phi}C_{\rho}C_{\psi}\sum_{r\in\sR}\left(\Phi_{l-1}\l2{\mW_{r}^{(l)}}+\Lambda_{l-1}\l2{\mU_{r}^{(l)}}\right)k_r\quad(\text{definition of } k_r)\notag\\
        \leq& C_{\phi}s_l\Phi_{l-1}+C_{\phi}C_{\rho}C_{\psi}s_l(\Phi_{l-1}+\Lambda_{l-1})\sum_{r\in\sR}k_r\quad(\text{definition of } s_l)\notag\\
        \leq& C_{\phi}s_l\Phi_{l-1}+C_{\phi}C_{\rho}C_{\psi}s_l\left(\Phi_{l-1}+\l2{\xr}\prod_{i=1}^{l-1}s_i\right)\sum_{r\in\sR}k_r\quad(\text{Eq.\eqref{relPhi}})\notag\\
        \leq& \tau s_l\Phi_{l-1}+\kappa\l2{\xr}\prod_{i=1}^{l}s_i\quad(\text{definitions of $\tau$ and $\kappa$})\label{eq:vbound}
      \end{align}
    }
    \end{footnotesize}

Note that $\Phi_1\leq\tau s_1\Phi_0 + \kappa \l2{\xr} s_1\leq\tau s_1\l2{\xv} + \kappa \l2{\xr} s_1$ by Eq. \eqref{eq:vbound} and the definition of the spectral norm. Then,
  
  \begin{small}
  \begin{align*}
  \left(\Phi_l-\kappa\l2{\xr}\prod_{i=1}^{l}s_i\right)-\kappa\l2{\xr}\left(\sum_{i=1}^{l-1}\tau^i\right)\left(\prod_{i=1}^{l}s_i\right)\leq&\tau s_l \Phi_{l-1}-\kappa\l2{\xr}\left(\sum_{i=1}^{l-1}\tau^i\right)\left(\prod_{i=1}^{l}s_i\right)\\
  \end{align*}
  \end{small}
  which leads to
  \begin{footnotesize}
  \begin{align*}
  \Phi_l-\kappa\l2{\xr}\left(\sum_{i=0}^{l-1}\tau^i\right)\left(\prod_{i=1}^{l}s_i\right)\leq&\tau s_l\left(\Phi_{l-1}-\kappa\l2{\xr}\left(\sum_{i=0}^{l-2}\tau^i\right)\left(\prod_{i=1}^{l-1}s_i\right)\right)\\
  \leq&\tau^{l-1}\left(\prod_{i=2}^l s_i\right)\left(\Phi_{1}-\kappa\l2{\xr}\left(\sum_{i=0}^{0}\tau^i\right)\left(\prod_{i=1}^{1}s_i\right)\right)\hspace{0.5em} (\text{recursion})\\
  \leq&\tau^{l-1}\left(\prod_{i=2}^l s_i\right)\left(\tau s_1\l2{\xv} + \kappa \l2{\xr} s_1-\kappa\l2{\xr}s_1\right)\\
  \leq&\tau^l\l2{\xv}\prod_{i=1}^l s_i
  \end{align*}
  \end{footnotesize}
 
  Finally, we get
  \begin{align}
    \Phi_{l}\leq& \tau^l\l2{\xv}\prod_{i=1}^l s_i + \kappa\l2{\xr}\left(\sum_{i=0}^{l-1}\tau^i\right)\left(\prod_{i=1}^{l}s_i\right)
    =\left(\tau^l\l2{\xv}+\kappa\l2{\xr}\sum_{i=0}^{l-1}\tau^i\right)\prod_{i=1}^{l}s_i = \eta_l \prod_{i=1}^{l}s_i\label{rmpnnPhi}
  \end{align}
  where \(\eta_l = \tau^l\l2{\xv}+\kappa\l2{\xr}\sum_{i=0}^{l-1}\tau^i\). 
  \item \paragraph{\raisebox{.9pt}{\textcircled{\raisebox{-.9pt} {\small 3}}} Bound of $\Psi_{l}$}\mbox{}\\
  We can calculate the upper bound of $\Psi_{l}$ using sub-additivity, sub-multiplicativity, and the definitions of matrix norms. Note that the bound of $\Psi_{L}$ can be obtained by setting $l=L$.  
  
  Let $v^{**}=\argmax_v\l2{\mtH^{(l)}[v,:]-\mH^{(l)}[v,:]}$. Then,
  {\allowdisplaybreaks
    {\small
    \begin{align}
    \Psi_{l} =& \left\|\phi\left(\mtH^{(l-1)}\mtW_{0}^{(l)}+\rho\left(\sum_{r\in\sR}\mS_r^{(l)}\psi\left(\mtM_{r}^{(l)}\right)\left[\begin{matrix}\mtW_{r}^{(l)}\\\mtU_{r}^{(l)}\end{matrix}\right]\right)\right)\left[v^{**},:\right]\right.\notag\\
    &\quad\left.-\phi\left(\mH^{(l-1)}\mW_{0}^{(l)}+\rho\left(\sum_{r\in\sR}\mS_r^{(l)}\psi\left(\mM_{r}^{(l)}\right)\left[\begin{matrix}\mW_{r}^{(l)}\\\mU_{r}^{(l)}\end{matrix}\right]\right)\right)[v^{**},:]\right\|_2\notag\\
    \leq& C_{\phi}\left\|\left(\left(\mtH^{(l-1)}\mtW_{0}^{(l)}\right)[v^{**},:]+\rho\left(\sum_{r\in\sR}\mS_r^{(l)}\psi\left(\mtM_{r}^{(l)}\right)\left[\begin{matrix}\mtW_{r}^{(l)}\\\mtU_{r}^{(l)}\end{matrix}\right]\right)[v^{**},:]\right)\quad(\text{Lipschitzness of }\phi)\right.\notag\\
    &\qquad\left. -\left(\left(\mH^{(l-1)}\mW_{0}^{(l)}\right)[v^{**},:]+\rho\left(\sum_{r\in\sR}\mS_r^{(l)}\psi\left(\mM_{r}^{(l)}\right)\left[\begin{matrix}\mW_{r}^{(l)}\\\mU_{r}^{(l)}\end{matrix}\right]\right)[v^{**},:]\right)\right\|_2\notag\\
    \leq& C_{\phi}\l2{\left(\mtH^{(l-1)}\mtW_{0}^{(l)}\right)[v^{**},:]-\left(\mH^{(l-1)}\mW_{0}^{(l)}\right)[v^{**},:]}\qquad(\text{sub-additivity})\notag\\
    &+C_{\phi}\l2{\rho\left(\sum_{r\in\sR}\mS_r^{(l)}\psi\left(\mtM_{r}^{(l)}\right)\left[\begin{matrix}\mtW_{r}^{(l)}\\\mtU_{r}^{(l)}\end{matrix}\right]\right)[v^{**},:]-\rho\left(\sum_{r\in\sR}\mS_r^{(l)}\psi\left(\mM_{r}^{(l)}\right)\left[\begin{matrix}\mW_{r}^{(l)}\\\mU_{r}^{(l)}\end{matrix}\right]\right)[v^{**},:]}\notag\\
    \leq& C_{\phi}\l2{\left(\mtH^{(l-1)}\mtW_{0}^{(l)}\right)[v^{**},:]-\left(\mH^{(l-1)}\mW_{0}^{(l)}\right)[v^{**},:]}\qquad(\text{Lipschitzness of $\rho$})\notag\\
    &+C_{\phi}C_{\rho}\l2{\left(\sum_{r\in\sR}\mS_r^{(l)}\psi\left(\mtM_{r}^{(l)}\right)\left[\begin{matrix}\mtW_{r}^{(l)}\\\mtU_{r}^{(l)}\end{matrix}\right]\right)[v^{**},:]-\left(\sum_{r\in\sR}\mS_r^{(l)}\psi\left(\mM_{r}^{(l)}\right)\left[\begin{matrix}\mW_{r}^{(l)}\\\mU_{r}^{(l)}\end{matrix}\right]\right)[v^{**},:]}\notag\\
    =& C_{\phi}\l2{\left(\mtH^{(l-1)}\mtW_{0}^{(l)}\right)[v^{**},:]-\left(\mH^{(l-1)}\mW_{0}^{(l)}\right)[v^{**},:]}\notag\\
    &+C_{\phi}C_{\rho}\left\|\left(\sum_{r\in\sR}\sum_{v\in\sV}\mS_r^{(l)}[v^{**},v]\left(\psi\left(\mtM_{r}^{(l)}\right)[v,:]\right)\left[\begin{matrix}\mtW_{r}^{(l)}\\\mtU_{r}^{(l)}\end{matrix}\right]\right)-\left(\sum_{r\in\sR}\sum_{v\in\sV}\mS_r^{(l)}[v^{**},v]\left(\psi\left(\mM_{r}^{(l)}\right)[v,:]\right)\left[\begin{matrix}\mW_{r}^{(l)}\\\mU_{r}^{(l)}\end{matrix}\right]\right)\right\|_2\notag\\
    =& C_{\phi}\l2{\left(\mtH^{(l-1)}\mtW_{0}^{(l)}\right)[v^{**},:]-\left(\mH^{(l-1)}\mW_{0}^{(l)}\right)[v^{**},:]}\notag\\
    &+C_{\phi}C_{\rho}\l2{\sum_{r\in\sR}\sum_{v\in\sV}\mS_r^{(l)}[v^{**},v]\left(\left(\psi\left(\mtM_{r}^{(l)}\right)[v,:]\right)\left[\begin{matrix}\mtW_{r}^{(l)}\\\mtU_{r}^{(l)}\end{matrix}\right]-\left(\psi\left(\mM_{r}^{(l)}\right)[v,:]\right)\left[\begin{matrix}\mW_{r}^{(l)}\\\mU_{r}^{(l)}\end{matrix}\right]\right)}\notag\\
    \leq& C_{\phi}\l2{\left(\mtH^{(l-1)}\mtW_{0}^{(l)}\right)[v^{**},:]-\left(\mH^{(l-1)}\mW_{0}^{(l)}\right)[v^{**},:]}\quad(\text{definition of $\mM_{r}^{(l)}$ and sub-additivity})\notag\\
    &+C_{\phi}C_{\rho}\sum_{r\in\sR}\sum_{v\in\sV}\bbsize{\mS_r^{(l)}[v^{**},v]}\left\|\psi\left(\left[\begin{matrix}\mtH^{(l-1)}[v,:]&\mtR^{(l-1)}[r,:]\end{matrix}\right]\right)\left[\begin{matrix}\mtW_{r}^{(l)}\\\mtU_{r}^{(l)}\end{matrix}\right]\right.\notag\\
    &\left.\hspace{12.5em}-\psi\left(\left[\begin{matrix}\mH^{(l-1)}[v,:]&\mR^{(l-1)}[r,:]\end{matrix}\right]\right)\left[\begin{matrix}\mW_{r}^{(l)}\\\mU_{r}^{(l)}\end{matrix}\right]\right\|_2\quad\text{(absolute homogeneity)}\notag\\
    =& C_{\phi}\l2{\left(\mtH^{(l-1)}\mtW_{0}^{(l)}\right)[v^{**},:]-\left(\mH^{(l-1)}\mW_{0}^{(l)}\right)[v^{**},:]}\notag\\
    &+C_{\phi}C_{\rho}\sum_{r\in\sR}\sum_{v\in\sV}\bbsize{\mS_r^{(l)}[v^{**},v]}\left\|\psi\left(\mtH^{(l-1)}[v,:]\right)\mtW_{r}^{(l)}+\psi\left(\mtR^{(l-1)}[r,:]\right)\mtU_{r}^{(l)}\right.\notag\\
    &\left.\hspace{12.5em}-\psi\left(\mH^{(l-1)}[v,:]\right)\mW_{r}^{(l)}-\psi\left(\mR^{(l-1)}[r,:]\right)\mU_{r}^{(l)}\right\|_2\notag\\
     =& C_{\phi}\l2{\left(\left(\mtH^{(l-1)}-\mH^{(l-1)}\right)\mtW_{0}^{(l)}\right)[v^{**},:]+\left(\mH^{(l-1)}\mdW_{0}^{(l)}\right)[v^{**},:]}\qquad(\text{definition of \mtW})\notag\\
    &+C_{\phi}C_{\rho}\sum_{r\in\sR}\sum_{v\in\sV}\bbsize{\mS_r^{(l)}[v^{**},v]}\left\|\left(\psi\left(\mtH^{(l-1)}[v,:]\right)-\psi\left(\mH^{(l-1)}[v,:]\right)\right)\mtW_{r}^{(l)}\right.\notag\\
    &\hspace{13em}+\left(\psi\left(\mtR^{(l-1)}[r,:]\right)-\psi\left(\mR^{(l-1)}[r,:]\right)\right)\mtU_{r}^{(l)}\notag\\
    &\left.\hspace{13em}+\psi\left(\mH^{(l-1)}[v,:]\right)\mdW_{r}^{(l)}+\psi\left(\mR^{(l-1)}[r,:]\right)\mdU_{r}^{(l)}\right\|_2\notag\\
    \leq& C_{\phi}\l2{\left(\mtH^{(l-1)}-\mH^{(l-1)}\right)[v^{**},:]}\l2{\mtW_{0}^{(l)}}+C_{\phi}\l2{\mH^{(l-1)}[v^{**},:]}\l2{\mdW_{0}^{(l)}}\qquad(\text{sub-additivity, sub-multiplicativity})\notag\\
    &+C_{\phi}C_{\rho}\sum_{r\in\sR}\sum_{v\in\sV}\bbsize{\mS_r^{(l)}[v^{**},v]}\left(\l2{\psi\left(\mtH^{(l-1)}[v,:]\right)-\psi\left(\mH^{(l-1)}[v,:]\right)}\l2{\mtW_{r}^{(l)}}\right.\notag\\
    &\hspace{13em}+\l2{\psi\left(\mtR^{(l-1)}[r,:]\right)-\psi\left(\mR^{(l-1)}[r,:]\right)}\l2{\mtU_{r}^{(l)}}\notag\\
    &\hspace{13em}+\left.\l2{\psi\left(\mH^{(l-1)}[v,:]\right)}\l2{\mdW_{r}^{(l)}}+\l2{\psi\left(\mR^{(l-1)}[r,:]\right)}\l2{\mdU_{r}^{(l)}}\right)\notag\\
    \leq& C_{\phi}\l2{\left(\mtH^{(l-1)}-\mH^{(l-1)}\right)[v^{**},:]}\l2{\mtW_{0}^{(l)}}+C_{\phi}\l2{\mH^{(l-1)}[v^{**},:]}\l2{\mdW_{0}^{(l)}}\qquad(\text{Lipschitzness of }\psi,\psi(\vzero)=\vzero)\notag\\
    &+C_{\phi}C_{\rho}C_{\psi}\sum_{r\in\sR}\sum_{v\in\sV}\bbsize{\mS_r^{(l)}[v^{**},v]}\left(\l2{\mtH^{(l-1)}[v,:]-\mH^{(l-1)}[v,:]}\l2{\mtW_{r}^{(l)}}\right.\notag\\
    &\hspace{14em}+\l2{\mtR^{(l-1)}[r,:]-\mR^{(l-1)}[r,:]}\l2{\mtU_{r}^{(l)}}\notag\\
    &\left.\hspace{14em}+\l2{\mH^{(l-1)}[v,:]}\l2{\mdW_{r}^{(l)}}+\l2{\mR^{(l-1)}[r,:]}\l2{\mdU_{r}^{(l)}}\right)\notag\\
    \leq& C_{\phi}(s_l+\ds )\Psi_{l-1}+C_{\phi}\Phi_{l-1}\ds\qquad(\text{definitions of }\Phi_{l}, \Psi_{l}, \Lambda_{l}, \Gamma_{l}, s_l,\text{ and }\ds)\notag\\
    &+C_{\phi}C_{\rho}C_{\psi}\sum_{r\in\sR}\sum_{v\in\sV}\bbsize{\mS_r^{(l)}[v^{**},v]}\big(\Psi_{l-1}(s_l+\ds)+\Gamma_{l-1}(s_l+\ds)+\Phi_{l-1}\ds+\Lambda_{l-1}\ds\big)\notag\\
    \leq& C_{\phi}(s_l+\ds )\Psi_{l-1}+C_{\phi}\Phi_{l-1}\ds\qquad(\text{definition of the infinity norm of a matrix})\notag\\
    &+C_{\phi}C_{\rho}C_{\psi}\big(\Psi_{l-1}(s_l+\ds)+\Gamma_{l-1}(s_l+\ds)+\Phi_{l-1}\ds+\Lambda_{l-1}\ds\big)\sum_{r\in\sR}\linf{\mS_r^{(l)}}\notag\\
    =& C_{\phi}(s_l+\ds )\Psi_{l-1}+C_{\phi}\Phi_{l-1}\ds\notag\\
    &+C_{\phi}C_{\rho}C_{\psi}(\Psi_{l-1}+\Gamma_{l-1})(s_l+\ds )\sum_{r\in\sR}\linf{\mS_r}+C_{\phi}C_{\rho}C_{\psi}(\Phi_{l-1}+\Lambda_{l-1})\ds\sum_{r\in\sR}\linf{\mS_r^{(l)}}\notag\\
    \leq& C_{\phi}(s_l+\ds )\Psi_{l-1}+C_{\phi}\Phi_{l-1}\ds\qquad(\text{definition of }k_r)\notag\\
    &+C_{\phi}C_{\rho}C_{\psi}(\Psi_{l-1}+\Gamma_{l-1})(s_l+\ds )\sum_{r\in\sR}k_r+C_{\phi}C_{\rho}C_{\psi}(\Phi_{l-1}+\Lambda_{l-1})\ds\sum_{r\in\sR}k_r\notag\\
    \leq& \tau(s_l+\ds )\Psi_{l-1}+\tau\Phi_{l-1}\ds
    +\kappa (s_l+\ds )\Gamma_{l-1}+\kappa \ds\Lambda_{l-1}\qquad(\text{definitions of $\tau$ and $\kappa$})\label{eq:psimiddle}\\
    \leq& \tau(s_l+\ds )\Psi_{l-1}+\tau \ds\eta_{l-1}\prod_{i=1}^{l-1}s_i+\kappa (s_l+\ds )\l2{\xr}\left(\prod_{i=1}^{l-1}(s_i+\ds ) - \prod_{i=1}^{l-1}s_i\right)+\kappa \ds\l2{\xr}\prod_{i=1}^{l-1}s_i\qquad(\text{Eq.~\eqref{relPhi},~\eqref{relPsi},~\eqref{rmpnnPhi}})\notag\\
    =&\tau(s_l+\ds )\Psi_{l-1}+\tau \ds\eta_{l-1}\prod_{i=1}^{l-1}s_i+\kappa (s_l+\ds )\l2{\xr}\prod_{i=1}^{l-1}(s_i+\ds ) - \kappa s_l\l2{\xr}\prod_{i=1}^{l-1}s_i\notag\\
    =& \tau(s_l+\ds )\Psi_{l-1}+\tau \ds\eta_{l-1}\prod_{i=1}^{l-1}s_i+\kappa \l2{\xr}\prod_{i=1}^{l}(s_i+\ds )-\kappa \l2{\xr}\prod_{i=1}^{l}s_i\notag\\
    =& \tau(s_l+\ds )\Psi_{l-1}+\tau \ds\eta_{l-1}\prod_{i=1}^{l-1}s_i+\kappa\l2{\xr}\left(\prod_{i=1}^{l}(s_i+\ds )-\prod_{i=1}^{l}s_i\right)\notag
  \end{align}
  }
  }
  From the above inequality, we can induce
  \begin{align*}
    \Psi_{l} + \tau\eta_{l-1}\prod_{i=1}^{l}s_i+\kappa\l2{\xr}\prod_{i=1}^{l}s_i\leq&\tau(s_l+\ds)\Psi_{l-1}+\tau(s_l+\ds)\eta_{l-1}\prod_{i=1}^{l-1}s_i+\kappa\l2{\xr}\prod_{i=1}^{l}(s_i+\ds)&
  \end{align*}
  where 
  \begin{align*}
  \Psi_{l} + \tau\eta_{l-1}\prod_{i=1}^{l}s_i+\kappa\l2{\xr}\prod_{i=1}^{l}s_i = \Psi_{l} + (\tau\eta_{l-1}+\kappa\l2{\xr})\prod_{i=1}^{l}s_i=\Psi_{l} + \eta_l\prod_{i=1}^{l}s_i
  \end{align*}
  
  Putting all this together, we have
  {\footnotesize
  \begin{align*}
  &\Psi_{l} + \eta_l\prod_{i=1}^{l}s_i-\kappa\l2{\xr}\prod_{i=1}^{l}(s_i+\ds )-\kappa\l2{\xr}\left(\sum_{i=1}^{l-1}\tau^i\right)\left(\prod_{i=1}^{l}(s_i+\ds)\right)\\
  &\leq\tau(s_l+\ds )\Psi_{l-1}+\tau(s_l+\ds)\eta_{l-1}\prod_{i=1}^{l-1}s_i-\kappa\l2{\xr}\left(\sum_{i=1}^{l-1}\tau^i\right)\left(\prod_{i=1}^{l}(s_i+\ds)\right)\\
  \end{align*}}
   Note that $\Psi_{1}\leq \tau(s_1+\ds )\Psi_{0}+\tau\ds\Phi_{0}
    +\kappa (s_1+\ds )\Gamma_{0}+\kappa \ds\Lambda_{0}\leq\tau\ds\l2{\xv}+\kappa\ds\l2{\xr}$ by Eq.~\eqref{eq:psimiddle} and the definitions of $\Psi_0, \Phi_0, \Lambda_0$ and $\Gamma_0$.
    
  Gathering up,
  {\footnotesize
  \begin{align*}
  &\Psi_{l} + \eta_{l}\prod_{i=1}^{l}s_i-\kappa\l2{\xr}\left(\sum_{i=0}^{l-1}\tau^i\right)\left(\prod_{i=1}^{l}(s_i+\ds)\right)\\
  &\leq\tau(s_l+\ds)\left(\Psi_{l-1}+\eta_{l-1}\prod_{i=1}^{l-1}s_i-\kappa\l2{\xr}\left(\sum_{i=0}^{l-2}\tau^i\right)\left(\prod_{i=1}^{l-1}(s_i+\ds)\right)\right)\\
  &\leq \tau^{l-1}\left(\prod_{i=2}^l(s_i+\ds)\right)\left(\Psi_{1}+\eta_{1}\prod_{i=1}^{1}s_i-\kappa\l2{\xr}\left(\sum_{i=0}^{0}\tau^i\right)\left(\prod_{i=1}^{1}(s_i+\ds)\right)\right)\qquad \text{(recursion)}\\
  &\leq \tau^{l-1}\left(\prod_{i=2}^l(s_i+\ds)\right)\left(\ds(\tau\l2{\xv}+\kappa\l2{\xr})+(\tau\l2{\xv}+\kappa\l2{\xr})s_1-\kappa\l2{\xr}(s_1+\ds)\right)\\
  &=\tau^l\left(\prod_{i=1}^l(s_i+\ds)\right)\l2{\xv}
  \end{align*}}
  
  We end up with
  \begin{align}
    \Psi_{l}\leq& \tau^l\left(\prod_{i=1}^l(s_i+\ds)\right)\l2{\xv} + \kappa\l2{\xr}\left(\sum_{i=0}^{l-1}\tau^i\right)\left(\prod_{i=1}^{l}(s_i+\ds)\right) - \eta_{l}\prod_{i=1}^{l}s_i\notag\\
    =& \eta_l\left(\prod_{i=1}^{l}(s_i+\ds )-\prod_{i=1}^{l}s_i\right)
    \qquad (\text{definition of }\eta_l)\label{rmpnnPsi}
  \end{align}

  Then,
  \begin{align}
  \max_{(h,r,t)\in\sE}\linf{f_{\vw+\vdw}(h,r,t)-f_{\vw}(h,r,t)}
  \leq& (2\Psi_{L}+\Gamma_{L})(s_{L+1}+\ds ) + (2\Phi_{L}+\Lambda_{L})\ds \notag\qquad(\text{Eq.~\eqref{tdmax}})\\
  \leq& \left(\prod_{i=1}^{L}(s_i+\ds )-\prod_{i=1}^{L}s_i\right)(2\eta_{L}+\l2{\xr})(s_{L+1}+\ds ) \notag \quad(\text{Eq.~\eqref{relPhi},~\eqref{relPsi},~\eqref{rmpnnPhi},~\eqref{rmpnnPsi}})\\
  &+ \left(\prod_{i=1}^{L}s_i\right)(2\eta_{L}+\l2{\xr})\ds\notag\\
  =& \left(\prod_{i=1}^{L+1}(s_i+\ds )-\prod_{i=1}^{L+1}s_i\right)(2\eta_{L}+\l2{\xr}) \notag\\
  \leq& \left((s+\ds )^{L+1}-s^{L+1}\right)\zeta_L \qquad(\text{definition of }s)\notag\\
  \leq& \ds(L+1){(s+\ds )}^{L}\zeta_L \qquad(0\leq s \leq s + \ds)\label{eq:perturbTD}
  \end{align}
  where \(\zeta_l = 2\eta_l+\l2{\xr}\).
  
  \item \paragraph{Generalization bound of ReED with translational distance decoder}
  Recall that we set the prior distribution $\sP$ as $\sN\left(\vzero_{n_\vw},\sigma^2\mI_{n_\vw\times n_\vw}\right)$. Since the distribution of the perturbed parameters $\sQ_{\vw+\vdw}$ is $\sN\left(\vw,\sigma^2\mI_{n_\vw \times n_\vw}\right)$, the distribution of the perturbation is $\sN\left(\vzero_{n_\vw},\sigma^2\mI_{n_\vw \times n_\vw}\right)$. For the perturbation matrices that follow the normal distribution, we can derive the following by replacing the standard normal variable in Corollary 4.2. in~\citeauthor{rmat}~\cite{rmat} with a random variable following the normal distribution having the mean of $0$ and the variance of $\sigma^2$.
  \begin{align}
  &\prob\left(\bbl2{\mdW_{0}^{(l)}}\geq \ds\right) \leq \left(d_{l-1}+d_l\right)e^{-\ds^2/\left(2\max(d_{l-1},d_l\right)\sigma^2)}\leq 2de^{-\ds^2/(2d\sigma^2)}\notag\\
  &\prob\left(\bbl2{\mdU_{0}^{(l)}}\geq \ds\right) \leq (d_{l-1}'+d_l')e^{-\ds^2/(2\max(d'_{l-1},d'_l)\sigma^2)}\leq 2de^{-\ds^2/(2d\sigma^2)}\notag\\
  &\prob\left(\bbl2{\mdW_{r}^{(l)}}\geq \ds\right) \leq (d_{l-1}+d_l)e^{-\ds^2/(2\max(d_{l-1},d_l)\sigma^2)}\leq 2de^{-\ds^2/(2d\sigma^2)}\notag\\
  &\prob\left(\bbl2{\mdU_{r}^{(l)}}\geq \ds\right) \leq (d_{l-1}'+d_l)e^{-\ds^2/(2\max(d'_{l-1},d_{l})\sigma^2)}\leq 2de^{-\ds^2/(2d\sigma^2)}\notag\\
  &\prob\left(\bbl2{\mbdW_{r}^{\langle j \rangle}} \geq \ds\right) \leq (d_{L}+d_{L+1})e^{-\ds^2/(2\max(d_{L},d_{L+1})\sigma^2)} \leq 2de^{-\ds^2/(2d\sigma^2)}\notag\\
  &\prob\left(\bbl2{\mbdU_{r}^{\langle j \rangle}} \geq \ds\right) \leq (d_{L}'+d_{L+1})e^{-\ds^2/(2\max(d_{L}',d_{L+1})\sigma^2)} \leq 2de^{-\ds^2/(2d\sigma^2)}\notag\\
  &\prob\left(\bbl2{\mdV_{r}^{\langle j \rangle}} \geq \ds\right) \leq (d_{L}+d_{L+1})e^{-\ds^2/(2\max(d_L, d_{L+1})\sigma^2)} \leq 2de^{-\ds^2/(2d\sigma^2)}\notag
  \end{align}
  Using Bernoulli's inequality, we can derive that the probability of all perturbation matrices having the spectral norm less than $\ds$ is greater than or equal to $1-2N_{\vw} de^{-\ds^2/(2d\sigma^2)}$, where \(N_{\vw} = 2\cdot L+2\cdot\nr L+3\cdot2\nr=2\nr L+6\nr+2L\) is the number of perturbation matrices.
  
  To satisfy the condition of Theorem~\ref{thm:transbound}, we set $2N_{\vw} de^{-\ds^2/(2d\sigma^2)}=1/2$. Then, we get $\ds=\sigma\sqrt{2d\ln{(4N_{\vw} d)}}$. Since the prior is independent of the learned parameters $\vw$, we cannot directly use $s$ to formulate $\sigma$. Therefore, we approximate $s$ with $\mathring{s}$ in the following range.
  \begin{align}
  \size{s-\mathring{s}}\leq \frac{1}{L+2}s \implies \frac{L+1}{L+2}s \leq \mathring{s} \leq \frac{L+3}{L+2}s\notag
  \end{align}
  
  Additionally, we assume that $\ds \leq \frac{1}{L+2}s$.
  Then, if
  \begin{align}
  \max_{(h,r,t)\in\sE}\linf{f_{\vw+\vdw}(h,r,t)-f_{\vw}(h,r,t)}\leq& \ds(L+1)(s+\ds )^{L}\zeta_L\qquad(\text{Eq.~\eqref{eq:perturbTD}}) \notag\\
  \leq& \ds(L+1)s^{L}\left(1+\frac{1}{L+2}\right)^{L}\zeta_L \qquad\left(\ds \leq \frac{1}{L+2}s\right)\notag\\
  \leq& \ds(L+1){\left(\frac{L+2}{L+1}\right)}^{L}\mathring{s}^{L}\left(1+\frac{1}{L+2}\right)^{L}\zeta_L \hspace{0.4em} (\text{range of }\mathring{s})\notag\\
  =& \ds(L+1)\mathring{s}^{L}\left(1+\frac{2}{L+1}\right)^{L}\zeta_L \notag\\
  \leq& \ds(L+1)\mathring{s}^{L}e^2\zeta_L \leq \frac{\gamma}{4}\qquad\left(\left(1+\frac{1}{x}\right)^x \leq e, \forall x \geq 0\right)\notag
  \end{align}
  is satisfied, we meet the condition of Theorem~\ref{thm:transbound}. 
  With $\ds=\sigma\sqrt{2d\ln{(4N_{\vw} d)}}$, we have
  \begin{align}
  &\ds=\sigma\sqrt{2d\ln{(4N_{\vw} d)}}\leq \frac{\gamma}{4e^2(L+1)\mathring{s}^{L}\zeta_L} \notag\\
  \rightarrow&\sigma\leq \frac{1}{\sqrt{2d\ln{(4N_{\vw} d)}}}\left(\frac{\gamma}{4e^2(L+1)\mathring{s}^{L}\zeta_L}\right)\notag
  \end{align}
  
  By setting $\sigma = \frac{1}{\sqrt{2d\ln{(4N_{\vw} d)}}}\left(\frac{\gamma}{4e^2(L+1)\mathring{s}^{L}\zeta_L}\right)$, we can calculate $D_{KL}(\sQ_{\vw+\vdw}\|\sP)$.
  \begin{align}
  D_{KL}(\sQ_{\vw+\vdw}\|\sP) & = \frac{\l2{\vw}^2}{2\sigma^2}\qquad (\text{KL divergence between two normal distributions})\notag
  \\ &= \frac{2d\ln{(4N_{\vw} d)}}{2\left(\frac{\gamma}{4e^2(L+1)\mathring{s}^{L}\zeta_L}\right)^2}\left(\sum \lfr{\mW}^2\right)\notag
  \\ &\leq \frac{{(4e^2(L+1)\mathring{s}^{L}\zeta_L)}^2d\ln{(4N_{\vw} d)}}{\gamma^2}\left(N_{\vw} s^2\right)\notag
  \\ &= \frac{16e^4N_{\vw} {(L+1)}^2\mathring{s}^{2L} s^2\zeta_L^2d\ln{(4N_{\vw} d)}}{\gamma^2}\notag
  \\ &\leq \frac{16e^4N_{\vw} {(L+1)}^2 s^2{\left(\frac{L+3}{L+2}s\right)}^{2L}\zeta_L^2d\ln{(4N_{\vw} d)}}{\gamma^2}\qquad (\text{range of }\mathring{s})\notag\\
  &\leq \frac{16e^6N_{\vw} {(L+1)}^2 s^{2L+2}\zeta_L^2d\ln{(4N_{\vw} d)}}{\gamma^2}\qquad \left(\left(1+\frac{1}{x}\right)^x \leq e,\forall x \geq 0\right)\notag
  \end{align}
  From Theorem~\ref{thm:transbound}, we get
  \begin{align}
  \sL_{0,\sE}(f_{\vw})\leq& \sL_{\gamma,\shE}(f_{\vw}) + \sqrt{\frac{1-\frac{\size{\shE}}{\nE}}{2|\shE|}\left[2D_{KL}(\sQ_{\vw+\vdw}\|\sP)+\ln\frac{4\theta(|\shE|,\nE)}{\delta}\right]}\notag\\
  \sL_{0,\sE}(f_{\vw})\leq& \sL_{\gamma,\shE}(f_{\vw}) + \sqrt{\frac{1-\frac{|\shE|}{\nE}}{|\shE|}\left[\frac{16e^6N_{\vw} {(L+1)}^2 s^{2L+2}\zeta_L^2d\ln{(4N_{\vw} d)}}{\gamma^2}+\frac{1}{2}\ln\frac{4\theta(|\shE|,\nE)}{\delta}\right]}\label{eq:singleDist}
  \end{align}
  
  Now, let us find some range of $s$ such that Theorem~\ref{thm:boundtd} trivially holds.
  
  First, if
  \begin{small}
    \begin{align}
    \linf{f_{\vw}(h,r,t)} =& \max_{(h,r,t)\in\sE}\max_{j\in \{0,1\}}\Big|-\l2{\mH^{(L)}[h,:]\mbW_{r}^{\langle j \rangle}+\mR^{(L)}[r,:]\mbU_{r}^{\langle j \rangle}-\mH^{(L)}[t,:]\mV_{r}^{\langle j \rangle}}\Big|\notag\\ 
    \leq& \max_{(h,r,t)\in\sE}\max_{j\in \{0,1\}}\left(\l2{\mH^{(L)}[h,:]}\l2{\mbW_{r}^{\langle j \rangle}}\right.\quad(\text{sub-additivity, sub-multiplicativity})\notag\\
    &\hspace{7em}\left.+\l2{\mR^{(L)}[r,:]}\l2{\mbU_{r}^{\langle j \rangle}}+\l2{\mH^{(L)}[t,:]}\l2{\mV_{r}^{\langle j \rangle}}\right)\notag\\ 
    \leq& \left(2\Phi_{L}+\Lambda_{L}\right)s_{L+1}\notag \qquad (\text{definitions of }\Phi_{l}, \Lambda_{L},\text{ and }s_{L+1})\\
    \leq& \left(2\eta_{L}+\l2{\xr}\right)s_{L+1}\prod_{i=1}^{L}s_i
    \leq \zeta_L s^{L+1} < \frac{\gamma}{2}\notag\\
      \to& s < {\left(\frac{\gamma}{2\zeta_L}\right)}^{\frac{1}{L+1}}\notag
    \end{align}
  \end{small}then Theorem~\ref{thm:boundtd} trivially holds since \(\sL_{0,\sE}(f_{\vw})=\sL_{\gamma,\shE}(f_{\vw})=1\) when \(\linf{f_{\vw}(h,r,t)}<\frac{\gamma}{2}\) holds for all $(h,r,t)\in\sE$.

  Also, if 
  \begin{align}
  &\sqrt{\frac{1-\frac{|\shE|}{\nE}}{|\shE|}\left[\frac{16e^6 N_{\vw} {(L+1)}^2 s^{2L+2}\zeta_L^2d\ln{(4N_{\vw} d)}}{\gamma^2}+\frac{1}{2}\ln\frac{4\theta(|\shE|,\nE)}{\delta}\right]}
  \geq \sqrt{\left(\frac{1-\frac{|\shE|}{\nE}}{|\shE|}\right)\left(\frac{4s^{2L+2}\zeta_L^2}{\gamma^2}\right)} > 1\notag
  \\ &\rightarrow s > {\left(\frac{\gamma}{2\zeta_L}\sqrt{\frac{|\shE|}{1-\frac{|\shE|}{\nE}}}\right)}^{\frac{1}{L+1}} \notag
  \end{align}
  then Theorem~\ref{thm:boundtd} holds regardless of the value of $\sL_{0,\sE}(f_{\vw})$ and $\sL_{\gamma,\shE}(f_{\vw})$ because the value of the loss function cannot exceed 1.

  Therefore, we only need to consider $s$ in range
  \begin{align}\label{eq:srangeDist}
    {\left(\frac{\gamma}{2\zeta_L}\right)}^{\frac{1}{L+1}} \leq s \leq {\left(\frac{\gamma}{2\zeta_L}\sqrt{\frac{|\shE|}{1-\frac{|\shE|}{\nE}}}\right)}^{\frac{1}{L+1}}
  \end{align}
  
  We also need to check whether the assumption $\ds\leq \frac{1}{L+2}s$ holds in this range. Note that if $\frac{\gamma}{4e^2(L+1)\zeta_L\left(\frac{L+1}{L+2}s\right)^{L}} \leq \frac{1}{L+2}s\notag$ holds, then the assumption also holds since $\ds=\frac{\gamma}{4e^2(L+1)\mathring{s}^{L}\zeta_L} \leq \frac{\gamma}{4e^2(L+1)\zeta_L\left(\frac{L+1}{L+2}s\right)^{L}}$. With a simple calculation, we get
  \begin{align}
    s^{L+1} \geq \frac{(L+2)\gamma}{4e^2(L+1)\zeta_L{\left(\frac{L+1}{L+2}\right)}^{L}}\notag
  \end{align}
  The above inequality holds if $s$ is in the range of Eq.~\eqref{eq:srangeDist}, since
  \begin{align}
    \frac{(L+2)\gamma}{4e^2(L+1)\zeta_L{\left(\frac{L+1}{L+2}\right)}^{L}}
    =&\frac{\gamma}{4e^2\zeta_L}{\left(1+\frac{1}{L+1}\right)}^{L+1}
    \leq\frac{\gamma}{4e\zeta_L} \leq \frac{\gamma}{2\zeta_L} \leq s^{L+1}\qquad \left(\left(1+\frac{1}{x}\right)^x \leq e, \forall x \geq 0\right)\notag
  \end{align}
  
  Therefore, we only need to consider Eq.~\eqref{eq:srangeDist} because otherwise Theorem~\ref{thm:boundtd} holds regardless of the choice of $\sigma$. While Eq.~\eqref{eq:singleDist} holds with probability $1-\delta$, it only holds for $s$ such that $\frac{L+1}{L+2}s\leq\mathring{s}\leq\frac{L+3}{L+2}s$. To make Eq.\eqref{eq:singleDist} hold for all $s$ in range Eq.~\eqref{eq:srangeDist}, we need to select multiple $\mathring{s}$ so that any $s$ in range Eq.~\eqref{eq:srangeDist} can be covered. By assuming that $\size{s-\mathring{s}}\leq \frac{1}{L+2}{\left(\frac{\gamma}{2\zeta_L}\right)}^{\frac{1}{L+1}} \leq \frac{1}{L+2}s$, we can calculate the number of $\mathring{s}$ we need to consider, i.e., the size of covering $C$, by dividing the length of the range of $s$ in Eq.~\eqref{eq:srangeDist} by the length of each cover, i.e., $\frac{2}{L+2}{\left(\frac{\gamma}{2\zeta_L}\right)}^{\frac{1}{L+1}}$. Let $\size{C}$ denote the size of covering $C$. By simple division, we get $\size{C}=\frac{(L+2)}{2}\left({\left(\sqrt{\frac{1}{|\shE|}-\frac{1}{\nE}}\right)}^{-\frac{1}{L+1}}-1\right)$. Using Bernoulli's inequality, we can conclude that the probability of Eq.~\eqref{eq:singleDist} holding simultaneously for $\size{C}$ choices of $\mathring{s}$ is $1-\size{C}\delta$. Therefore,
  \begin{align}
  \sL_{0,\sE}(f_{\vw})\leq& \sL_{\gamma,\shE}(f_{\vw})
  +\sqrt{\frac{1-\frac{|\shE|}{\nE}}{|\shE|}\left[\frac{16e^6N_{\vw} (L+1)^2 s^{2L+2}\zeta_L^2d\ln{(4N_{\vw} d)}}{\gamma^2}+\frac{1}{2}\ln\frac{4\theta(|\shE|,\nE)\size{C}}{\delta}\right]}\notag\\
  \leq& \sL_{\gamma,\shE}(f_{\vw})+ \sO\left(\sqrt{\frac{1-\frac{\nhE}{\nE}}{\nhE}\left[\frac{N_{\vw} L^2\zeta_L^2 s^{2L}d\ln{(N_{\vw} d)}}{\gamma^2}+\ln\frac{\theta(\nhE,\nE)}{\delta}\right]}\right)\notag
  \end{align}
  holds with probability of $1-\size{C}\cdot\frac{\delta}{\size{C}}=1-\delta$ regardless of $s$.
\end{proof}

\begin{thm:boundsm}[Generalization Bound for ReED with Semantic Matching Decoder]
  For any \(L\geq 0\), let \(f_{\vw} : \sV\times\sR\times\sV\rightarrow\RR^2\) be a triplet classifier designed by the combination of the RAMP encoder with $L$-layers in Definition~\ref{def:ramp} and the SM decoder in Definition~\ref{def:sm}. Let $k_r$ be the maximum of the infinity norms for all possible $\mS_r^{(l)}$ in the RAMP encoder. Then, for any \(\delta, \gamma>0\), with probability at least $1-\delta$ over a training triplet set \(\shE\) (such that $20\leq \nhE \leq \nE -20$) sampled without replacement from the full triplet set \(\sE\), for any \(\vw\), we have
\begin{align*}
       \sL_{0,\sE}(f_{\vw}) \leq \sL_{\gamma,\shE}(f_{\vw})
      +\sO\!\left(\!\!\!\sqrt{\frac{1-\frac{\nhE}{\nE}}{\nhE}\left[\frac{N_{\vw} L^2\eta_{L}^4{s}^{4L}d\ln{(N_{\vw} d)}}{\gamma^2}+\ln\frac{\theta(\nhE,\nE)}{\delta}\right]}\right)\notag
    \end{align*}
    where \(\theta(\nhE,\nE) = 3\sqrt{\nhE(1-\frac{\nhE}{\nE})}\ln{\nhE}, N_{\vw} = 2\nr L + \) 
    $2\nr+2L$, $\eta_L = \tau^L\l2{\xv}+\kappa\l2{\xr}\sum_{i=0}^{L-1}\tau^i, d=$
    $\max(\max_{0\leq l\leq L}(d_l),\max_{0\leq l\leq L}(d_l')), \tau =C_\phi+\kappa,$
    $\kappa=C_\phi C_\rho C_\psi\sum_{r\in\sR}k_r$, $s_{L+1}=\max_{r,j}\lfr{\mbU_{r}^{\langle j \rangle}}, s_l = $
    $\max(\lfr{\mW_{0}^{(l)}}, \lfr{\mU_{0}^{(l)}},\max_r\lfr{\mW_{r}^{(l)}},\max_r\lfr{\mU_{r}^{(l)}})$
    for $l \in \{1,2,\ldots,L\}$, and $s=\max_{1\leq l\leq L+1}(s_l)$.
\end{thm:boundsm}

\begin{proof}
We derive Theorem~\ref{thm:boundsm} from Theorem~\ref{thm:transbound} where we construct a posterior distribution $\sQ_{\vw+\vdw}$ by adding random perturbations $\vdw$ to $\vw$. Following~\cite{gcpac, nnpac}, we set the prior distribution $\sP$ as $\sN\left(\vzero_{n_\vw}, \sigma^2\mI_{n_\vw \times n_\vw}\right)$ and the posterior distribution $\sQ_{\vw+\vdw}$ as $ \sN\left(\vw, \sigma^2\mI_{n_\vw\times n_\vw}\right)$ where $n_\vw$ is the size of $\vw$. We first compute $\max_{(h,r,t)\in\sE}\linf{f_{\vw+\ddot{\vw}}(h,r,t)-f_{\vw}(h,r,t)}$, which we call the perturbation bound, so that we can calculate the standard deviation $\sigma$ of the prior distribution that satisfies $\prob\left(\max_{(h,r,t)\in\sE}\linf{f_{\vw+\vdw}(h,r,t)-f_{\vw}(h,r,t)} < \frac{\gamma}{4}\right)>\frac{1}{2}$. Afterwards, we calculate the KL divergence of $\sQ_{\vw+\vdw}$ from $\sP$ using $\sigma$ and substitute the KL divergence term in Theorem~\ref{thm:transbound} with our data and model-related terms, which finishes the proof.

  \item \paragraph{Perturbation bound of ReED with semantic matching decoder}
First, we compute the perturbation bound, $\max_{(h,r,t)\in\sE}\linf{f_{\vw+\ddot{\vw}}(h,r,t)-f_{\vw}(h,r,t)}$, and find $\sigma$ that makes $\prob(\max_{(h,r,t)\in\sE}\linf{f_{\vw+\vdw}(h,r,t)$-$f_{\vw}(h,r,t)} < \frac{\gamma}{4})>\frac{1}{2}$ true. Let $\mdW$ denote a perturbation (also called noise) matrix added to the original weight matrix $\mW$. As a result, we have $\mtW = \mW + \mdW$ where $\mtW$ is a perturbed weight matrix. Also, let $\mtH^{(l)}$ and $\mtR^{(l)}$ denote the outputs of the perturbed model at the $l$-th layer. Each element of $\mdW$ is an i.i.d. element drawn from $\sN\left(0,\sigma^2\right)$. Assume that the maximum of the Frobenius norms of the noise matrices is $\ds$. That is,  
  \[\ds = \max_l\left(\bblfr{\mdW_{0}^{(l)}}, \bblfr{\mdU_{0}^{(l)}},\max_r\bblfr{\mdW_{r}^{(l)}},\max_r\bblfr{\mdU_{r}^{(l)}}, \max_{r,j}\bblfr{\mbdU_{r}^{\langle j \rangle}}\right)\]
  Now, let us calculate the perturbation bound.
  \begin{align}
  &\max_{(h,r,t)\in\sE}\linf{f_{\vw+\vdw}(h,r,t)-f_{\vw}(h,r,t)} \notag
  \\&= \max_{(h,r,t)\in\sE}\max_{j\in\{0,1\}}\size{\mtH^{(L)}[h,:]\mbtU_{r}^{\langle j \rangle}\left(\mtH^{(L)}[t,:]\right)^\top-\mH^{(L)}[h,:]\mbU_{r}^{\langle j \rangle}\left(\mH^{(L)}[t,:]\right)^\top}\notag
  \end{align}
  Recall that $\Phi_{l} = \max_v\l2{\mH^{(l)}[v,:]}$ and $\Psi_{l} = \max_v\l2{\mtH^{(l)}[v,:]-\mH^{(l)}[v,:]}$.
  
  For any $(h,r,t)\in\sE$,
  \begin{small}
    \begin{align}
      &\left|{\mtH^{(L)}[h,:]\mbtU_{r}^{\langle j \rangle}\left(\mtH^{(L)}[t,:]\right)^\top-\mH^{(L)}[h,:]\mbU_{r}^{\langle j \rangle}\left(\mH^{(L)}[t,:]\right)^\top}\right|\notag\\
      =&\left|\left(\mtH^{(L)}[h,:]-\mH^{(L)}[h,:]+\mH^{(L)}[h,:]\right)\left(\mbU_{r}^{\langle j \rangle}+\mbdU_{r}^{\langle j \rangle}\right)\left(\mtH^{(L)}[t,:]-\mH^{(L)}[t,:]+\mH^{(L)}[t,:]\right)^\top\right.\notag\\
      &\hspace{0.5em}\left.-\mH^{(L)}[h,:]\mbU_{r}^{\langle j \rangle}\left(\mH^{(L)}[t,:]\right)^\top\right|\quad(\text{definition of }\mbtU_{r}^{\langle j \rangle})\notag \\
      =&\left|\left(\mtH^{(L)}[h,:]-\mH^{(L)}[h,:]\right)\mbU_{r}^{\langle j \rangle}\left(\mtH^{(L)}[t,:]-\mH^{(L)}[t,:]+\mH^{(L)}[t,:]\right)^\top\right.\notag\\
      &\hspace{0.5em}+\mH^{(L)}[h,:]\mbU_{r}^{\langle j \rangle}\left(\mtH^{(L)}[t,:]-\mH^{(L)}[t,:]\right)^\top\notag\\
      &\hspace{0.5em}\left.+\left(\mtH^{(L)}[h,:]-\mH^{(L)}[h,:]+\mH^{(L)}[h,:]\right)\mbdU_{r}^{\langle j \rangle}\left(\mtH^{(L)}[t,:]-\mH^{(L)}[t,:]+\mH^{(L)}[t,:]\right)^\top\right|\notag\\
      \leq&\l2{\mtH^{(L)}[h,:]-\mH^{(L)}[h,:]}\l2{\mbU_{r}^{\langle j \rangle}}\left(\l2{\mtH^{(L)}[t,:]-\mH^{(L)}[t,:]}+\l2{\mH^{(L)}[t,:]}\right)\notag\\
      &+\l2{\mH^{(L)}[h,:]}\l2{\mbU_{r}^{\langle j \rangle}}\l2{\mtH^{(L)}[t,:]-\mH^{(L)}[t,:]}\quad(\text{sub-additivity, sub-multiplicativity})\notag \\
      &+\left(\l2{\mtH^{(L)}[h,:]-\mH^{(L)}[h,:]}\hspace{-0.3em}+\l2{\mH^{(L)}[h,:]}\right)\bbl2{\mbdU_{r}^{\langle j \rangle}}\hspace{-0.1em}\left(\l2{\mtH^{(L)}[t,:]-\mH^{(L)}[t,:]}\hspace{-0.3em}+\l2{\mH^{(L)}[t,:]}\right)\notag\\
      \leq&\Psi_{L}\left(\Psi_{L}+2\Phi_{L}\right)s_{L+1}+\left(\Psi_{L}+\Phi_{L}\right)^2\ds\notag\qquad (\text{definitions of }\Psi_{L}, \Phi_{L},\text{ and }s_{L+1})
    \end{align}
  \end{small}
  Therefore,
  \begin{equation}
    \max_{(h,r,t)\in\sE}\linf{f_{\vw+\vdw}(h,r,t)-f_{\vw}(h,r,t)} \leq \Psi_{L}(\Psi_{L}+2\Phi_{L})s_{L+1}+(\Psi_{L}+\Phi_{L})^2\ds\notag
  \end{equation}
  
  Note that we can calculate the upper bounds of $\Phi_{L}$ and $\Psi_{L}$ using Eq.~\eqref{rmpnnPhi} and Eq.~\eqref{rmpnnPsi}, respectively. Then,
    \begin{align}
    \max_{(h,r,t)\in\sE}&\linf{f_{\vw+\vdw}(h,r,t)-f_{\vw}(h,r,t)} \notag\\
    \leq&\Psi_{L}(\Psi_{L}+2\Phi_{L})s_{L+1}+(\Psi_{L}+\Phi_{L})^2\ds\notag\\
    \leq& \eta_{L}\left(\prod_{i=1}^{L}(s_i+\ds )-\prod_{i=1}^{L}s_i\right)\left(\eta_{L}\left(\prod_{i=1}^{L}(s_i+\ds )-\prod_{i=1}^{L}s_i\right)+2\eta_{L}\prod_{i=1}^{L}s_i\right)s_{L+1}\notag\\
    &+{\left(\eta_{L}\left(\prod_{i=1}^{L}(s_i+\ds)-\prod_{i=1}^{L}s_i\right)+\eta_{L}\prod_{i=1}^{L}s_i\right)}^2\ds \notag\\
    =& \eta_{L}^2\left({\left(\prod_{i=1}^{L}(s_i+\ds )\right)}^2-{\left(\prod_{i=1}^{L}s_i\right)}^2\right)s_{L+1} + \eta_{L}^2{\left(\prod_{i=1}^{L}(s_i+\ds )\right)}^2\ds \notag\\
    =& \eta_{L}^2\left({\left({\left(\prod_{i=1}^{L}(s_i+\ds )\right)}^2-{\left(\prod_{i=1}^{L}s_i\right)}^2\right)}s_{L+1}+\left(\prod_{i=1}^{L}(s_i+\ds )\right)^2\ds\right)\notag\\
    \leq& \eta_{L}^2\left(\left({(s+\ds )}^{2L}-s^{2L}\right)s +{(s+\ds )}^{2L}\ds\right) \qquad(\text{definition of }s)\notag\\
    =& \eta_{L}^2\left({(s+\ds )}^{2L+1}-s^{2L+1}\right)\label{eq:perturbSM}
  \end{align}
  \item \paragraph{Generalization bound of ReED with semantic matching decoder}
Recall that we set the prior distribution $\sP$ as $\sN\left(\vzero_{n_\vw},\sigma^2\mI_{n_\vw\times n_\vw}\right)$. Since the distribution of the perturbed parameters $\sQ_{\vw+\vdw}$ is $\sN\left(\vw,\sigma^2\mI_{n_\vw \times n_\vw}\right)$, the distribution of the perturbation is $\sN\left(\vzero_{n_\vw},\sigma^2\mI_{n_\vw \times n_\vw}\right)$. For the perturbation matrices that follow the normal distribution, we can derive the following by replacing the standard normal variable in Corollary 4.2. in~\citeauthor{rmat}~\cite{rmat} with a random variable following the normal distribution having the mean of $0$ and the variance of $\sigma^2$.
  \begin{align}
  &\prob\left(\l2{\mdW_{0}^{(l)}}\geq \ds\right) \leq (d_{l-1}+d_l)e^{-\ds^2/(2\max(d_{l-1},d_{l})\sigma^2)}\leq 2de^{-\ds^2/(2d\sigma^2)}\notag\\
  &\prob\left(\l2{\mdU_{0}^{(l)}}\geq \ds\right) \leq (d_{l-1}'+d_l')e^{-\ds^2/(2\max(d'_{l-1},d'_{l})\sigma^2)}\leq 2de^{-\ds^2/(2d\sigma^2)}\notag\\
  &\prob\left(\l2{\mdW_{r}^{(l)}}\geq \ds\right) \leq (d_{l-1}+d_l)e^{-\ds^2/(2\max(d_{l-1},d_{l})\sigma^2)}\leq 2de^{-\ds^2/(2d\sigma^2)}\notag\\
  &\prob\left(\l2{\mdU_{r}^{(l)}}\geq \ds\right) \leq (d_{l-1}'+d_l)e^{-\ds^2/(2\max(d'_{l-1},d_{l})\sigma^2)}\leq 2de^{-\ds^2/(2d\sigma^2)}\notag\\
  &\prob\left(\l2{\mbdU_{r}^{\langle j \rangle}} \geq \ds\right) \leq 2d_{L}e^{-\ds^2/(2d_L\sigma^2)} \leq 2de^{-\ds^2/(2d\sigma^2)}\notag
  \end{align}
Using Bernoulli's inequality, we can derive that the probability of all perturbation matrices having the spectral norm less than $\ds$ is greater than or equal to $1-2N_{\vw} de^{-\ds^2/(2d\sigma^2)}$, where \(N_{\vw} = 2\cdot L+2\cdot \nr L+2\nr=2\cdot\nr L+2\nr+2L\) is the number of perturbation matrices.
  
  To satisfy the condition of Theorem~\ref{thm:transbound}, we set $2N_{\vw} de^{-\ds^2/(2d\sigma^2)}=1/2$. Then, we get $\ds=\sigma\sqrt{2d\ln{(4N_{\vw} d)}}$. Since the prior is independent of the learned parameters $\vw$, we cannot directly use $s$ to formulate $\sigma$. Therefore, we approximate $s$ with $\mathring{s}$ in the following range.    
  \begin{align}
  \size{s-\mathring{s}}\leq \frac{1}{2L+2}s\implies  \frac{2L+1}{2L+2}s \leq \mathring{s} \leq \frac{2L+3}{2L+2}s\notag
  \end{align}
  Additionally, we assume $\ds \leq \frac{1}{2L+2}s$. Then, if
  \begin{align}
  \max_{(h,r,t)}\linf{f_{\vw+\vdw}(h,r,t)-f_{\vw}(h,r,t)}\leq& \> \eta_{L}^2\left((s+\ds )^{2L+1}-s^{2L+1}\right) \qquad \text{(Eq.\eqref{eq:perturbSM})}\notag
  \\\leq&\> \eta_{L}^2\ds(2L+1)(s+\ds )^{2L} \qquad (0 \leq s \leq s+\ds)\notag
  \\\leq&\> \eta_{L}^2\ds(2L+1)s^{2L}\left(1+\frac{1}{2L+2}\right)^{2L}\qquad \left(\ds \leq \frac{1}{2L+2}s\right)\notag
  \\\leq&\> \eta_{L}^2\ds(2L+1)\left(\frac{2L+2}{2L+1}\mathring{s}\right)^{2L}\hspace{-0.2em}\left(1+\frac{1}{2L+2}\right)^{2L}\hspace{-0.5em} (\text{range of }\mathring{s})\notag
  \\=&\> \eta_{L}^2\ds(2L+1)\mathring{s}^{2L}{\left(1+\frac{2}{2L+1}\right)}^{2L}\notag
  \\\leq&\> \eta_{L}^2\ds(2L+1)\mathring{s}^{2L}e^2 \leq \frac{\gamma}{4}\qquad \left(\left(1+\frac{1}{x}\right)^x \leq e, \forall x \geq 0 \right)\notag
  \end{align} 
  is satisfied, we meet the condition of Theorem~\ref{thm:transbound}. With $\ds=\sigma\sqrt{2d\ln{(4N_{\vw} d)}}$, we have
  \begin{align}
  &\ds=\sigma\sqrt{2d\ln{(4N_{\vw} d)}}\leq \frac{\gamma}{4e^2(2L+1)\eta_{L}^2{\mathring{s}}^{2L}}\notag\\
  \rightarrow&\sigma\leq \frac{1}{\sqrt{2d\ln{(4N_{\vw} d)}}}\left(\frac{\gamma}{4e^2(2L+1)\eta_{L}^2{\mathring{s}}^{2L}}\right)\notag
  \end{align}
  
  By setting $\sigma = \frac{1}{\sqrt{2d\ln{(4N_{\vw} d)}}}\left(\frac{\gamma}{4e^2(2L+1)\eta_{L}^2{\mathring{s}}^{2L}}\right)$, we can calculate $D_{KL}(\sQ_{\vw+\vdw}\|\sP)$.
  \begin{align}
  D_{KL}(\sQ_{\vw+\vdw}\|\sP) & = \frac{\l2{\vw}^2}{2\sigma^2}\qquad (\text{KL divergence between two normal distributions})\notag
  \\ &= \frac{2d\ln{(4N_{\vw} d)}}{2\left(\frac{\gamma}{4e^2(2L+1)\eta_{L}^2{\mathring{s}}^{2L}}\right)^2}\left(\sum\lfr{\mW}^2\right)\notag
  \\ &\leq \frac{(4e^2(2L+1)\eta_{L}^2{\mathring{s}}^{2L})^2d\ln{(4N_{\vw} d)}}{\gamma^2}N_{\vw} s^2\notag
  \\ &\leq \frac{N_{\vw} s^2\left(4e^2(2L+1)\eta_{L}^2{\left(\frac{2L+3}{2L+2}s\right)}^{2L}\right)^2d\ln{(4N_{\vw} d)}}{\gamma^2}\qquad (\text{range of }\mathring{s})\notag
  \\ &\leq \frac{N_{\vw} s^2(4e^3(2L+1)\eta_{L}^2{s}^{2L})^2d\ln{(4N_{\vw} d)}}{\gamma^2}\qquad \left(\left(1+\frac{1}{x}\right)^x\leq e, \forall x \geq 0\right)\notag
  \\ &= \frac{16e^6N_{\vw} {(2L+1)}^2\eta_{L}^4{s}^{4L+2}d\ln{(4N_{\vw} d)}}{\gamma^2}\notag
  \end{align}
  From Theorem~\ref{thm:transbound}, we get
  {\footnotesize
  \begin{align}
    &\sL_{0,\sE}(f_{\vw}) \leq \sL_{\gamma,\shE}(f_{\vw})+\sqrt{\frac{1-\frac{|\shE|}{\nE}}{2|\shE|}\left[2D_{KL}(\sQ_{\vw+\vdw}\|\sP)+\ln\frac{4\theta(|\shE|,\nE)}{\delta}\right]}\notag\\
    \leq& \sL_{\gamma,\shE}(f_{\vw})+\sqrt{\frac{1-\frac{|\shE|}{\nE}}{|\shE|}\left[\frac{16e^6N_{\vw} {(2L+1)}^2\eta_{L}^4{s}^{4L+2}d\ln{(4N_{\vw} d)}}{\gamma^2}+\frac{1}{2}\ln\frac{4\theta(|\shE|,\nE)}{\delta}\right]}\label{eq:singleSem}
  \end{align}}
  Now, let us find some range of $s$ such that Theorem~\ref{thm:boundsm} trivially holds. First, if
  \begin{align}
  &\linf{f_{\vw}(h,r,t)}\leq \max\left(\left|{\mH^{(L)}[h,:]\mbU_{r}^{\langle 1 \rangle}{\left(\mH^{(L)}[t,:]\right)}^{\top}}\right|, \left|{\mH^{(L)}[h,:]\mbU_{r}^{\langle 2 \rangle}{\left(\mH^{(L)}[t,:]\right)}^{\top}}\right|\right)\notag\\
  &\leq \Phi_{L}^2s_{L+1}\leq s\eta_{L}^2{\left(\prod_{i=1}^{L}s_i\right)}^2\leq \eta_{L}^2s^{2L+1} < \frac{\gamma}{2}\notag\\
  &\rightarrow s < {\left(\frac{\gamma}{2\eta_{L}^2}\right)}^{\frac{1}{2L+1}}\notag
  \end{align}
  then Theorem~\ref{thm:boundsm} trivially holds since $\sL_{0,\sE}(f_{\vw})=\sL_{\gamma,\shE}(f_{\vw})=1$ when \(\linf{f_{\vw}(h,r,t)}<\frac{\gamma}{2}\) holds for all $(h,r,t)\in\sE$.
  Also, if 
  \begin{align}
  &\sqrt{\frac{1-\frac{|\shE|}{\nE}}{|\shE|}\left[\frac{16e^6N_{\vw} {(2L+1)}^2\eta_{L}^4{s}^{4L+2}d\ln{(4N_{\vw} d)}}{\gamma^2}+\frac{1}{2}\ln\frac{4\theta(|\shE|,\nE)}{\delta}\right]}\geq \sqrt{\frac{1-\frac{|\shE|}{\nE}}{|\shE|}\frac{4s^{4L+2}\eta_{L}^4}{\gamma^2}} > 1\notag
  \\ &\rightarrow s > {\left(\frac{\gamma}{2\eta_{L}^2}\sqrt{\frac{|\shE|}{1-\frac{|\shE|}{\nE}}}\right)}^{\frac{1}{2L+1}}\notag
  \end{align}
  then Theorem~\ref{thm:boundsm} holds regardless of the value of $\sL_{0,\sE}(f_{\vw})$ and $\sL_{\gamma,\shE}(f_{\vw})$ because the value of the loss cannot exceed 1.
  Therefore, we only need to consider $s$ in range
  \begin{align}\label{eq:srangeSem}
    {\left(\frac{\gamma}{2\eta_{L}^2}\right)}^{\frac{1}{2L+1}} \leq s \leq {\left(\frac{\gamma}{2\eta_{L}^2}\sqrt{\frac{|\shE|}{1-\frac{|\shE|}{\nE}}}\right)}^{\frac{1}{2L+1}}
  \end{align}
  We also need to check whether the assumption $\ds\leq \frac{1}{2L+2}s$ holds in this range. Note that if $\frac{\gamma}{4e^2(2L+1)\eta_{L}^2\left(\frac{2L+1}{2L+2}s\right)^{2L}} \leq \frac{1}{2L+2}s$ holds, then the assumption also holds since $\ds=\frac{\gamma}{4e^2(2L+1)\eta_{L}^2{\mathring{s}}^{2L}} \leq\frac{\gamma}{4e^2(2L+1)\eta_{L}^2\left(\frac{2L+1}{2L+2}s\right)^{2L}} $. With a simple calculation, we get
  \begin{align}
    s^{2L+1} \geq \frac{(2L+2)\gamma}{4e^2(2L+1)\eta_{L}^2\left(\frac{2L+1}{2L+2}\right)^{2L}}\notag
  \end{align}
  The above inequality holds if $s$ is in the range of Eq.~\eqref{eq:srangeSem} since
  \begin{align}
    \frac{(2L+2)\gamma}{4e^2(2L+1)\eta_{L}^2\left(\frac{2L+1}{2L+2}\right)^{2L}}=&\frac{\gamma}{4e^2\eta_L^2}\left(1+\frac{1}{2L+1}\right)^{2L+1}\leq\frac{\gamma}{4e\eta_L^2}\leq\frac{\gamma}{2\eta_L^2}\leq s^{2L+1} \qquad \left(\left(1+\frac{1}{x}\right)^x \leq e, \forall x \geq 0\right) \notag
  \end{align} 
  Therefore, we only need to consider Eq.~\eqref{eq:srangeSem} because otherwise Theorem~\ref{thm:boundsm} holds regardless of the choice of $\sigma$. While Eq.~\eqref{eq:singleSem} holds with probability $1-\delta$, it only holds for $s$ such that $\frac{2L+1}{2L+2}s \leq \mathring{s} \leq \frac{2L+3}{2L+2}s$. To make Eq.~\eqref{eq:singleSem} hold for all $s$ in range Eq.~\eqref{eq:srangeSem}, we need to select multiple $\mathring{s}$ so that any $s$ in range Eq.~\eqref{eq:srangeSem} can be covered. By assuming that $\size{s-\mathring{s}}\leq\frac{1}{2L+2}{\left(\frac{\gamma}{2\eta_{L}^2}\right)}^{\frac{1}{2L+1}} \leq \frac{1}{2L+2}s$, we can calculate the number of $\mathring{s}$ we need to consider, i.e., the size of covering $C$, by dividing the length of the range of $s$ in Eq.~\eqref{eq:srangeSem} by the length of each cover, i.e.,$\frac{2}{2L+2}{\left(\frac{\gamma}{2\eta_{L}^2}\right)}^{\frac{1}{2L+1}}$. Let $\size{C}$ denote the size of covering $C$. By simple division, we get $\size{C}=\frac{(2L+2)}{2}\left({\left(\sqrt{\frac{1}{|\shE|}-\frac{1}{\nE}}\right)}^{-\frac{1}{2L+1}}-1\right)$. Using Bernoulli's inequality, we can conclude that the probability of Eq.~\eqref{eq:singleSem} holding simultaneously for $\size{C}$ choices of $\mathring{s}$ is $1-\size{C}\delta$. Therefore,
  \begin{align}
  \sL_{0,\sE}(f_{\vw})\leq& \sL_{\gamma,\shE}(f_{\vw})+\sqrt{\frac{1-\frac{|\shE|}{\nE}}{|\shE|}\left[\frac{16e^6N_{\vw} {(2L+1)}^2\eta_{L}^4{s}^{4L+2}d\ln{(4N_{\vw} d)}}{\gamma^2}+\frac{1}{2}\ln\frac{4\theta(|\shE|,\nE)\size{C}}{\delta}\right]}\notag\\ 
  \leq& \sL_{\gamma,\shE}(f_{\vw})+ \sO\left(\sqrt{\frac{1-\frac{\nhE}{\nE}}{\nhE}\left[\frac{N_{\vw} L^2\eta_{L}^4{s}^{4L}d\ln{(N_{\vw} d)}}{\gamma^2}+\ln\frac{\theta(\nhE,\nE)}{\delta}\right]}\right)\notag
  \end{align}
holds with probability of $1-\size{C}\cdot\frac{\delta}{\size{C}} = 1-\delta$.
\end{proof}

\section{Experimental Details}
\label{appn:exp}
\begin{table}[t]
\centering
\caption{Dataset Statistic}
\label{tb:data}
\begin{tabular}{ccccc}
\toprule
& $\size{\sV}$ & $\size{\sR}$ & $\size{\shE}$ & $\size{\sE}$ \\
\midrule
\fb & 1,496 & 179 & 41,873 & 52,318 \\
\cde & 2,684 & 42 & 17,951 & 22,224 \\
\umls & 133 & 43 & 10,174 & 12,732 \\
\bottomrule
\end{tabular}
\end{table}

%\begin{table}[t]
%\centering
%\caption{Full Statistics of Datasets}
%\label{tb:fulldata}
%%\setlength{\tabcolsep}{0.59em}
%\begin{tabular}{ccccccc}
%\toprule
%& & & \fb & \cde & \umls & \\
%\midrule
%& $\sV$ & $\size{\sV}$ & 1,496 & 2,684 & 133 \\
% \hdashline\>
%& $\sR$ & $\size{\sR}$ & 179 & 42 & 43 \\
% \hdashline\>
%& \multirow{3}{*}{$\shE$} & $\size{\shE^+}$ & 20,962 & 16,698 & 5,133\\
%& & $\size{\shE^-}$ & 20,911 & 1,253 & 5,041\\
%& & $\size{\shE}$ & 41,873 & 17,951 & 10,174\\
% \hdashline\>
%& \multirow{3}{*}{$\sE$} & $\size{\sE^+}$ & 26,159 & 20,664 & 6,366\\
%& & $\size{\sE^-}$ & 26,159 & 1,560 & 6,366\\
%& & $\size{\sE}$ & 52,318 & 22,224 & 12,732\\
%\bottomrule
%\end{tabular}
%\end{table}

We conduct experiments on three real-world knowledge graphs: \fb~\cite{fb}, \cde~\cite{codex}, and \umls~\cite{umls,umls43}, shown in Table~\ref{tb:data}. We generate smaller versions of \fb and \cde via graph sampling~\cite{grail} for ease of analysis. We create smaller versions of \fb~\cite{fb} and \cde~\cite{codex} using a standard graph sampling~\cite{grail} and consider them as fully observed knowledge graphs $G$ for easier analysis. Specifically, we randomly sample five seed entities for \fb and ten seed entities for \cde. From the seed entities, we randomly sample 30 neighboring entities per hop for two hops. Then, we take all sampled entities and the triplets between them. While \cde contains negative triplets (i.e., false triplets needed for training a triplet classifier), \fb and \umls do not include negative triplets. For these datasets, we create negative triplets by corrupting either a head or a tail entity of each positive triplet, following~\citeauthor{ntn}~\cite{ntn}. 

We set $d_1=d_2=\cdots=d_L=d_{L+1}$ for all datasets. We use $d_1=96$ for \fb, $d_1=64$ for \cde, and $d_1=48$ for \umls. For RAMP+TD, we set the learning rate to be 0.0003 on \fb, 0.0005 for \cde, and 0.0002 for \umls. For RAMP+SM, we set the learning rate to be 0.0005 for all datasets. We set the margin of the margin loss for \fb and \cde to be 0.5, and 0.75 for \umls and run all models for 2,000 epochs.

In our implementation of ReED, we use the Adam optimizer~\cite{adam}. When implementing ReED, we used python 3.8 and PyTorch 1.12.1 with cudatoolkit 11.3. We run all our experiments using NVIDIA GeForce RTX 2080 Ti. We repeat each experiment ten times with the random seeds: 0, 10, 20, 30, 40, 50, 60, 70, 80, and 90. Our code and data are available at \url{https://github.com/bdi-lab/ReED} where more details about the experiments are explained in the README file.

\end{document}